\documentclass[12pt, letterpaper]{article}
\usepackage[margin=1in]{geometry}

\usepackage[linesnumbered, lined, ruled]{algorithm2e}
\usepackage{lscape}
\usepackage{booktabs}   
\usepackage{tabularx} 
\usepackage[sort, compress]{natbib}
\usepackage{graphicx}
\usepackage[colorlinks = true,citecolor=blue]{hyperref}

\usepackage{amsmath}
\usepackage{amssymb}
\usepackage{amsfonts}
\usepackage{amsthm}
\usepackage{color}
\usepackage{comment}
\usepackage{url}
\usepackage{multirow}

\newcommand\norm[1]{\left\lVert#1\right\rVert}
\newcommand{\bs}{\boldsymbol}

\newtheorem{definition}{Definition}
\newtheorem{proposition}[definition]{Proposition}
\newtheorem{theorem}[definition]{Theorem}

\begin{document}

\title{SAFES: Sequential Privacy and Fairness Enhancing Data Synthesis for Responsible AI}
\author{\small Spencer Giddens$^*$, Xiaon Lang$^\dagger$, and Fang Liu$^{\dagger\ddagger}$ \\
\small $^*$Lucy Family Institute for Data \& Society\vspace{-6pt}\\
\small University of Notre Dame, Notre Dame, IN 46556 \\
\small $^\dagger$Department of Applied and Computational Mathematics and Statistics\vspace{-6pt}\\
\small University of Notre Dame, Notre Dame, IN 46556 \\
\small $^\ddagger$Fang.Liu.131@nd.edu}
\date{}

\maketitle

\begin{abstract}
As data-driven and AI-based decision making gains widespread adoption across disciplines, it is crucial that both data privacy and decision fairness are appropriately addressed. 
Although differential privacy (DP) provides a robust framework for guaranteeing privacy and methods are available to  improve fairness, most prior work  treats the two concerns separately. Even though there are existing approaches that consider privacy and fairness simultaneously, they typically focus on a single specific learning task, limiting their generalizability. 
In response, we introduce SAFES, a \underline{S}equential Priv\underline{A}cy and \underline{F}airness \underline{E}nhancing data \underline{S}ynthesis procedure that sequentially combines DP data synthesis with a fairness-aware data preprocessing step.
SAFES allows users flexibility in navigating the privacy-fairness-utility trade-offs.
We illustrate SAFES with different DP synthesizers and fairness-aware data preprocessing methods and run extensive experiments on multiple real datasets to examine the privacy-fairness-utility trade-offs of synthetic data generated by SAFES.
Empirical evaluations demonstrate that for reasonable privacy loss, SAFES-generated synthetic data can achieve significantly improved fairness metrics with relatively low utility loss.

\noindent \textbf{keywords}: Differential privacy, Machine learning fairness, Synthetic data
\end{abstract}

\section{Introduction}
\label{sec:intro}

\subsection{Background}
Data-driven and AI-based decision making is  increasingly adopted across disciplines. 
These systems are frequently used to make socially consequential decisions including, but not limited to, loan approvals, hiring, and criminal justice decisions.
Yet the data collected in this process often contain sensitive personal information.
While such AI applications can and do have legitimate benefits, it is of paramount importance to ensure sensitive data are used responsibly and in accordance with the highest possible ethical standards.

There are two important ethical concerns when working with sensitive personal data in training machine learning (ML) and AI algorithms: privacy and fairness.
Even anonymized datasets can be leveraged by attackers to infer masked or removed data \citep{Narayanan2008, Ahn2015, Sweeney2015} and blackbox access to a model is sufficient to infer membership in the training data \citep{Shokri2017}, which can itself be a privacy violation.
Knowing that an individual belongs to a dataset used for recidivism prediction, for example, is equivalent to knowing that person was convicted of a crime.
The perpetuation of social discrimination, bias, and other forms of unfairness in the decisions made by ML/AI models raises another major ethical concern.
One famous previous work, for example, demonstrated the presence of discrimination against darker skin colors in commercial gender classification systems \citep{Buolamwini2018}.
The na\"{i}ve approach of simply removing the group indicator when training these models has been shown insufficient to properly ensure fairness \citep{Calders2013}.
Fairness has increasingly been recognized in the ML community as a complicated notion and a lot of research has been devoted to defining and ensuring fairness \citep{Verma2018}.

Since datasets with privacy concerns are likely to also have fairness concerns and vice versa, it is critical to develop efficient privacy- and fairness-enhancing methods for releasing and analyzing data.
In this work, we propose a framework for synthesizing data that simultaneously addresses privacy and fairness concerns by strategically and sequentially combining privacy-preserving data synthesis and fairness-aware data preprocessing.
We aim to provide a generalized solution that safeguards sensitive personal information, upholds fairness, and keeps the utility of released data close to the original simultaneously.

There remains a scarcity of research on general-use synthetic data that both satisfies DP guarantees and reduces structural bias, representing a critical area for advancing responsible AI.
We address this gap by proposing \emph{SAFES -- a \underline{S}equential Priv\underline{A}cy and \underline{F}airness \underline{E}nhancing data \underline{S}ynthesis} procedure -- which combines DP data synthesis with a fairness-aware data preprocessing. 
The output of SAFES is a synthetic dataset with formal privacy guarantees and improved fairness for downstream classifiers, largely preserved utility for general and ML applications.  

While applied in focus, this work is—to the best of our knowledge—the first conceptual framework to integrate differential privacy and fairness sequentially for synthetic data generation, and the first to empirically assess the framework’s utility and feasibility. We believe SAFES has several \emph{benefits}.
First, it is flexible with respect to both the level of privacy protection and the degree of fairness enforced (with certain data preprocessing methods for fairness). 
Second, 
SAFES exhibits \emph{fairness robustness} measured by various metrics across a wide range of privacy guarantees per our empirical results, implying that one can adjust the balance between privacy and utility without a significant sacrifice in fairness.
Third, 
SAFES is a \emph{general} framework that accommodates diverse DP synthesizers (with various types of DP privacy guarantees) and fairness-aware preprocessing methods aligned with specified fairness metrics.

\vspace{-6pt}\subsection{Related works}
\paragraph{Data synthesis with formal privacy guarantees}
Among the concepts developed for privacy protection, differential privacy (DP) \citep{Dwork2006Calibrating, Dwork2006OurData} has emerged as the state-of-the-art framework for guaranteeing privacy when releasing information from sensitive data.
To ensure DP, calibrated random noise at given privacy loss parameters is added to the outputs via DP mechanisms before release.
Many mechanisms have been developed for executing queries \citep{Hardt2012, Kotsogiannis2019} and 
training ML models \citep{Chaudhuri2011, Abadi2016} that satisfy DP.
However, the outputs of these methods are limited to the specific chosen task, and the privacy loss accumulates with each task executed, which limits the number of analyses that can be performed under a fixed privacy budget.
This fact motivates the use of DP for synthesizing datasets to be privacy-preserving counterparts to an original, sensitive dataset.
Data users can analyze DP-synthesized data as if it were the original data without additional privacy loss due to the immunity to post-processing property of DP.
Popular DP data synthesis methods include marginal-based synthesizers for discrete data \citep{Zhang2017,McKenna2019,Eugenio2021,McKenna2021,McKenna2022}. 
They compute a set of marginals with DP noise and train a model based on the sanitized marginals, from which data are synthesized.
Statistical models and deep generative models are also used for DP data synthesis \citep{Zhang2018, Bowen2020, NationalAcademies2024}.

\vspace{-6pt}\paragraph{Fairness-aware methods}
We focus on fairness for binary classification tasks in this paper, unless otherwise stated. This is because most work -- in the literature and in practice -- on fairness in AI/ML has focused on binary classification, since fairness challenges often appear in such problems. As a result, this area is the most studied and practically relevant.
Methods for ensuring fairness include preprocessing, in-processing, and post-processing procedures.
processing methods \citep{reweight, Hajian2013, Calmon2017} transform the dataset to be more fair prior to analysis. 
In-processing \citep{Kamishima2011, Fish2016} and post-processing \citep{Kamiran2012, Hardt2016} methods, on the other hand, modify the training process and the ultimate model decisions, respectively.
While each type of method takes a different approach, the ultimate goal is to ensure decisions made by ML classifiers are fair. 
Preprocessing methods aims to remove structural bias from the dataset itself, which in turn improves the fairness of classifiers trained with the preprocessed data. 
Specifically, those methods may assign weights to different dataset records \citep{reweight}, flip the protected attribute and/or outcome label until the proportion of each group receiving each outcome is similar \citep{Hajian2013}, and learn a randomized pre-processsor balancing discrimination, distortion, and utility \citep{Calmon2017}. 
The methods are flexible because they are agnostic to downstream analysis and learning tasks.
In-processing methods may incorporate regularizers when training a model to penalize over-reliance on a protected attribute \citep{Kamishima2011} or shift the classifier decision boundary for unprivileged groups \citep{Fish2016}.
Post-processing approaches include adjusting decision rules for probabilistic classifiers to favor unprivileged groups when prediction confidence is low \citep{Kamiran2012}, and reformulating trained classifiers to remove discrimination through linear programming \citep{Hardt2016}.

\vspace{-6pt}
\paragraph{Interplay between privacy, fairness, and utility}
The mutual influence between privacy and fairness has been widely studied. Research has explored the connection between classification fairness and DP, the potential inequitable application of DP guarantees across demographic groups \citep{Ekstrand2018}, and the possibility that fair ML methods increase privacy risks for underprivileged groups \citep{Chang2021}.
Studies report that DP-generated synthetic data can exacerbate unfairness \citep{Ganev2022}, and that fairness adjustments may unevenly allocate privacy risks \citep{Chang2021}.
Incorporating DP and fairness in ML/AI models generally entail a reduction in utility that is well-documented in the literature. There also exists a three-way trade-off between privacy, fairness, and utility.
\citep{Cummings2019, Agarwal2021} claim that it is impossible for a single mechanism to achieve both DP and fairness with non-trivial classification accuracy.
Classifiers trained on synthesized images via DP generative adversarial networks (GANs) exhibit reduced utility without fairness improvements \citep{Cheng2021}. 

For specific ML tasks with both DP and fairness constraints, methods exist for empirical risk minimization \citep{Ding2020}, logistic regression \citep{Xu2019}, and stochastic gradient descent (SGD) \citep{Tran2021}, among others. 
In- and post-processing fairness methods may also be modified to achieve DP \citep{Jagielski2019}. 
However, the output of these methods is a specific classifier, unlike our work that produces a full privacy-preserving and fairness-aware individual-level synthetic dataset agnostic to potential downstream uses.
\citep{Zhao2025} finds that common post-processing in DP outputs such as non-negativity or preservation of total population counts incur unfairness and varying the privacy budget across different groups mitigates these fairness issues.
Under-sampling a training dataset prior to DP synthesis may also produce better fairness metrics on downstream classification tasks \citep{Bullwinkel2022}.
Data synthesis with DP and ``justifiable fairness'' \citep{Salimi2019} can be achieved for a marginal-based DP synthesizer \citep{McKenna2021} by ensuring all directed paths between the protected attribute(s) and the response variable in the graphical model representation pass through a non-protected attribute \citep{Pujol2023}. 
However, since the fairness modification is entangled with data synthesis in this approach, it does not apply to scenarios where DP synthetic data were already released without fairness considerations. 
In addition, it only achieves justifiable fairness -- a binary condition (yes or no) and only one of many possible fairness definitions -- and does not guarantee fairness by other metrics, especially those with a continuous fairness parameter for tuning the trade-off between utility and fairness constraints.

\vspace{-6pt}\section{Definitions and notations}\vspace{-3pt}
We introduce, in this section, the definitions and notations employed in SAFES and the experiments.

\vspace{-6pt}\subsection{Differential privacy (DP)}\vspace{-3pt}
DP provides a theoretical framework for privacy by bounding the influence of a single individual in a dataset on outputs from the dataset.
Let $d(D, D^\prime)=1$ denote two 
neighboring datasets $D, D^\prime$, which differ by one individual.
\begin{definition}[($\varepsilon, \delta)$-differential privacy \cite{Dwork2006OurData, Dwork2006Calibrating}]\label{defn:DP}
    A randomized mechanism $\mathcal{M}$ is said to be \emph{($\varepsilon, \delta)$-differentially private} if for all $S\subset\textnormal{Range}(\mathcal{M})$ and for any neighboring datasets $D, D^\prime$,
    \begin{equation}
        P(\mathcal{M}(D)\in S) \le e^\varepsilon P(\mathcal{M}(D^\prime)\in S) + \delta.
    \end{equation}
    $\varepsilon>0$ and $\delta\in [0, 1)$ are privacy loss or budget parameters.
    If $\delta=0$, $(\varepsilon, \delta)$-DP reduces to $\varepsilon$-DP.
\end{definition}
Essentially, DP ensures that a mechanism output cannot differ too much with vs.~ without any single individual.
Smaller $\varepsilon$ corresponds to more privacy.
$\delta$ is often interpreted as the probability that $\varepsilon$-DP fails and is usually of $o(1/\mbox{poly}(n))$, where $n$ is the data sample size.
There are various extensions \citep{Bun2016, Mironov2017, Dong2019} to the original DP definition in Definition \ref{defn:DP}.
We present one of these extensions -- zero-concentrated DP 
-- that is used in our experiments.
\begin{definition}[Zero-concentrated DP (zCDP) \citep{Bun2016}]
    Let $D_\alpha$ be the R\'enyi divergence of order $\alpha$.
    A randomized mechanism $\mathcal{M}$ satisfies \emph{$\rho$-zCDP} if for any neighboring datasets $D, D^\prime$ and any $\alpha\in(1,\infty)$,
    \begin{equation}
        D_\alpha(\mathcal{M}(D)||\mathcal{M}(D^\prime))\le \alpha\rho.
    \end{equation}
\end{definition}
\begin{theorem}[Conversion of $\rho$-zCDP to $(\varepsilon, \delta)$-DP \citep{Canonne2020}]
\label{thm:conversion}
    Let $\mathcal{M}$ be a mechanism satisfying $\rho$-zCDP.
    For any given $\varepsilon\ge0$, $\mathcal{M}$ satisfies $(\varepsilon, \delta)$-DP with 
    \begin{equation*}\textstyle
        \delta = \min_{\alpha>1} \frac{\exp\{(\alpha-1)(\alpha\rho-\varepsilon)\}}{\alpha-1}\left(1-\frac{1}{\alpha}\right)^\alpha.
    \end{equation*}
\end{theorem}
To release an output from a function on $D$ with DP guarantees, mechanism $\mathcal{M}$ is calibrated according 
to the global sensitivity (GS) of the function, defined as follows (though originally defined using the $\ell_1$-norm \citep{Dwork2006Calibrating}, we use a more general definition $\ell_p$-GS \citep{Liu2019}. 
Let $f$ be a (potentially vector-valued) function of a dataset $D$. 
The \emph{$\ell_p$-GS} of $f$ is 
\begin{equation}
    \Delta_{p,f} = \max_{d(D,D^\prime) = 1}\norm{f(D) - f(D^\prime)}_p.
\end{equation}

The GS of a function is the maximum difference in the function outputs on two neighboring datasets.
The larger the GS is, the more noise is needed to achieve a fixed level of privacy guarantee via $\mathcal{M}$.
The Gaussian mechanism (Definition \ref{defn:Gauss}) and the exponential mechanism (Definition \ref{defn:exp}) are two commonly used DP mechanisms.
\begin{definition}[Gaussian mechanism \citep{Dwork2006OurData, Bun2016}]\label{defn:Gauss}
    Let $D\in\mathcal{D}$ be a dataset.
    For a given function $f:\mathcal{D}\rightarrow\mathbb{R}^n$ with $\ell_2$-global sensitivity $\Delta_{2, f}$, the Gaussian mechanism is $\mathcal{M}(D) = f(D) + \mathbf{e}$,
    where $e_i$ is drawn independently from Gaussian distribution $\mathcal{N}(0, \sigma^2)$.
\end{definition}
\begin{definition}[Exponential mechanism \citep{McSherry2007}]\label{defn:exp}
    Let $\xi>0$ be a privacy loss parameter.
    For a given target function $f:\mathcal{D}\rightarrow\mathcal{R}$ and utility function $u: \mathcal{D}\times \mathcal{R} \rightarrow\mathbb{R}$ with $\ell_1$-global sensitivity $\Delta_{1,u}$, the exponential mechanism releases $r\in\mathcal{R}$ with probability $P(\mathcal{M}(D)=r) \propto \exp\left(\frac{\xi u(D, r)}{2\Delta_{1,u}}\right)$.
\end{definition}
The Gaussian mechanism in Definition \ref{defn:Gauss} achieves $(\varepsilon, \delta)$-DP for $\sigma^2 = 2\log(1.25/\delta)\Delta_{2,f}^2/\varepsilon^2$ and $\rho$-zCDP for $\sigma^2 = \Delta_{2,f}^2/2\rho$.
The exponential mechanism in Definition \ref{defn:exp} achieves $\varepsilon$-DP for $\xi=\varepsilon$ and $\rho$-zCDP for $\xi=2\sqrt{2\rho}$ \citep{Cesar2021}.

Both $(\varepsilon,\delta)$-DP and $\rho$-zCDP are composable under repeated applications of randomized mechanisms to the same data.
Let $\mathcal{D}$ be the space of all possible datasets $D$ and let $\mathcal{M}_1:\mathcal{D}\rightarrow \mathcal{R}_1$ and $\mathcal{M}_2:\mathcal{D}\times \mathcal{R}_1\rightarrow\mathcal{R}_2$ satisfy $(\varepsilon_1,\delta_1)$-DP ($\rho_1$-zCDP) and $(\varepsilon_2,\delta_2)$-DP ($\rho_2$-zCDP), respectively.
Then $\mathcal{M}_2(D, \mathcal{M}_1(D))$ satisfies $(\varepsilon_1+\varepsilon_2,\delta_1+\delta_2)$-DP  ($(\rho_1+\rho_2)$-zCDP).
For $(\varepsilon,\delta)$-DP, this is known as basic composition \citep{mcsherry2009privacy}. 
While advanced composition gives a tighter privacy loss bound for $(\varepsilon,\delta)$-DP \citep{Dwork2010}, zCDP still provides tighter privacy loss bounds \citep{Bun2016, Mironov2017}.
This motivates the practice of specifying the overall privacy loss in terms of $(\varepsilon,\delta)$-DP, applying the conversion theorems to convert it to $\rho$-zCDP, then using zCDP procedures to compose privacy loss. 

Another useful property of DP, in addition to the privacy loss composition described above, is its immunity to post-processing \citep{Dwork2006Calibrating, Dwork2006OurData, Bun2016}.
That is, a post-processing procedure on an output of a DP mechanism, whether satisfying $(\varepsilon,\delta)$-DP or $\rho$-zCDP, does not incur further privacy loss, as long as the procedure does not access the original data.
In the case of synthetic data generated from a DP synthesizer, any tasks performed on the synthetic data would satisfy the same DP guarantees as the synthetic data itself.

\subsection{Fairness}
This work primarily focuses on binary outcome variables, as this is the primary setting for most fairness literature and metrics \cite{Mehrabi2021} and lends itself nicely to many real-world scenarios where fairness is a concern such as hiring and loan decisions. We seek to improve group fairness, meaning we desire that the privileged group does not receive undue over-representation among those with favorable outcomes/predictions based on group status.

We separate a dataset $D$ into three disjoint sets of variables, $D = (X, G, Y)$.
$Y$ is a binary variable consisting of ``favorable'' (e.g., approved for a loan) and ``unfavorable'' outcomes.
The protected attributes $G$ distinguish different groups ($g_1, g_2,\ldots$) (e.g., race, gender).
We assume each group variable can be collapsed to a binary attribute representing  ``privileged'' vs ``unprivileged''.
WLOG, we assume that the favorable outcome and the privileged group are each encoded as $1$.
$X$ contains the non-protected predictors.
To ensure that downstream classifiers trained on a dataset are fair with respect to the protected attributes, it is important to also assess the bias present in the dataset itself.
We introduce several metrics for measuring the dataset bias and downstream classifier fairness.

We consider a dataset to have \emph{structural bias} if the probability of an observation receiving the favorable outcome is different for observations in the privileged and unprivileged groups.  In statistical terms, structural bias is equivalent to non-zero correlation between group and the outcome variable.
Of course, not all such correlations indicate bias; some reflect genuine relationships.
Here, we focus on cases where this correlation, if it exists, is assumed to be non-causal and would disappear if we conditioned on other relevant covariates that are not related to privilege (e.g. years of schooling). We introduce conditional outcome difference in Definition \ref{def:COD} to measure the structural bias.
\begin{definition}[Conditional outcome difference (COD)]\label{def:COD}
Let $g$ be a protected attribute in $G$ in dataset $D\!=\!(X,G,Y)$, then
    \begin{equation}
        \textnormal{COD}(D) = P(Y=1|g=0) - P(Y=1|g=1).
    \end{equation}
\end{definition}
By definition, COD is a property of the dataset rather than of an  AI or ML algorithm.
When $Y$ is independent of $G$, COD = 0, which represents the least bias possible based on this definition.  

For datasets without structural bias, imbalances in data representation between privileged and unprivileged groups can increase the likelihood of prediction bias in downstream ML algorithms, which can be assessed using algorithmic fairness metrics.
Two such commonly used metrics for classifiers are statistical parity \citep{Dwork2012}, which ensures privileged and unprivileged groups are equally likely to get the favorable decision, and equalized odds \citep{Hardt2016}, which ensures privileged and unprivileged groups have identical true and false positive rates.
\begin{definition}[Statistical parity difference (SPD) and average odds difference (AOD)\citep{Dwork2012, Hardt2016}]
    Given a dataset $D=(X,G,Y)$, let $g$ be a protected attribute from $G$. 
    Let $\hat{Y}$ be the decision of a classifier learned from $D$. Then
    \begin{align}
        \textnormal{SPD}(D, \hat{Y}) = P(\hat{Y}& =1|g=0) - P(\hat{Y}=1|g=1),\\
        \textnormal{AOD}(D, \hat{Y}) = 0.5\big[&\big(P(\hat{Y}=1|Y=0,g=0) - P(\hat{Y}=1|Y=0,g=1)\big) + \notag\\ 
        &\big(P(\hat{Y}=1|Y=1,g=0) - P(\hat{Y}=1|Y=1,g=1)\big)\big].
    \end{align}
\end{definition}

We additionally define the conditional utility difference to measure the balance in the utility of a classifier between the unprivileged and privileged groups.
The utility function $u$ in CUD can be defined in various ways and is context-based.
For example, $u(Y, \hat{Y}|g)$ can be the classification accuracy in group $g$ or false negative rate  $P(\hat{Y}=0|Y=1, g)$.
\begin{definition}[conditional utility difference (CUD)]
    Given a dataset $D=(X,G,Y)$, let $g$ be a protected attribute from $G$ and $u(Y,\hat{Y}|g)$ be an arbitrary conditional utility function.
    Then 
    \begin{equation}
        \textnormal{CUD}(D, \hat{Y}) = u(Y,\hat{Y}|g=0) - u(Y,\hat{Y}|g=1).
    \end{equation}
\end{definition}
With the flexibility and generality of $u$, CUD encompasses many existing fairness metrics, such as overall accuracy equality \citep{Berk2021}, predictive parity \citep{Chouldechova2017}, and false positive/negative rate balance \citep{Chouldechova2017}. 

For all fairness metrics presented in this section, a value closer to $0$ is more fair.
Negative values typically indicate unfairness in favor of the privileged group (e.g., a greater proportion of the privileged group is approved for a loan), while positive values favor the unprivileged group.
We estimate each of these metrics empirically in our experiments.


\section{Approach}\label{sec:method}

\subsection{The SAFES procedure}
\label{sec:SAFES}

\begin{figure}[!htb]
    \vspace{-6pt}\centering
    \includegraphics[width=0.6\textwidth]{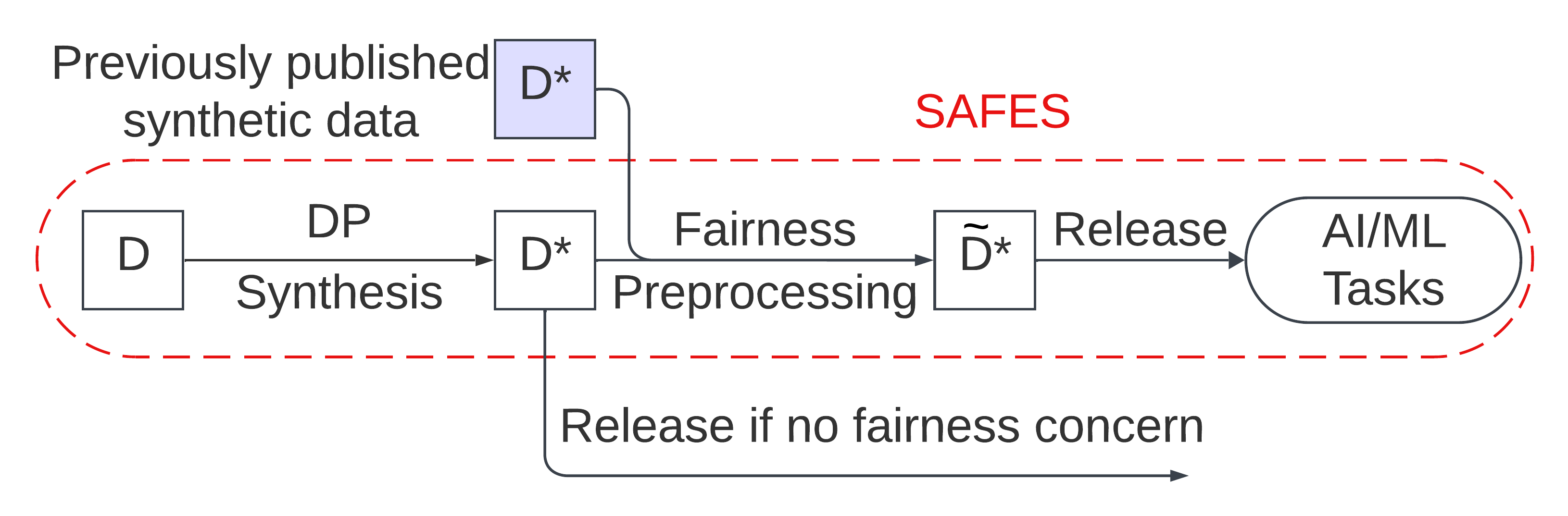} \vspace{-6pt}
    \caption{The SAFES procedure and its applications}
    \label{fig:safes}\vspace{-6pt}
\end{figure}

SAFES stands for \underline{S}equential Priv\underline{A}cy and \underline{F}airness \underline{E}nhancing data \underline{S}ynthesis. 
It is comprised of a DP data synthesizer and a fairness-aware data processor.
Fig.~\ref{fig:safes} provides a visualization of SAFES.
The figure suggests that SAFES is flexible and modular, allowing users to selectively apply its steps without requiring a strict start-to-finish approach.
Users can also directly release the DP synthetic data if fairness is not a concern or skip the DP synthesis step to work with previously released DP synthetic data.

Accordingly, the SAFES algorithm, as presented in  Algorithm \ref{alg:SAFES}, consists of two sequential steps: DP data synthesis followed by fairness-aware data preprocessing. The formal claim for its DP guarantees is given in Proposition \ref{pro:SAFES_DP}.

\begin{algorithm}[H]
    \SetAlgoLined
    \DontPrintSemicolon
    \SetKwInOut{Input}{Input}
    \SetKwInOut{Output}{Output}
    \Input{Dataset $D$, DP privacy loss parameters $\Theta_1$, fairness parameters $\Theta_2$ if applicable.}
    \Output{Privacy-preserving and fairness-aware synthetic dataset $\tilde{D}^*$.}
    \BlankLine
    Generate a privacy-preserving synthetic dataset $D^*$ at privacy loss $\Theta_1$ via a DP synthesizer $S$.\;
    Transform $D^*$  (with fairness parameters $\Theta_2$ if applicable) to obtain $\tilde{D}^*$ via a fairness data processor $T$.\;
    \caption{The SAFES Procedure}
    \label{alg:SAFES}
    \Return{$\tilde{D}^*$}
\end{algorithm}
\begin{proposition}
\label{pro:SAFES_DP}
    SAFES satisfies DP at privacy loss $\Theta_1$.
\end{proposition}
The proof is straightforward.
Since $S$ satisfies DP with parameters $\Theta_1$, $D^*$ achieves $\Theta_1$-DP guarantees.
Since the fairness data preprocessing $T$ only operates on $D^*$ and never accesses the original data $D$, the post-processing theorem for DP ensures the same DP guarantees on $\tilde{D}^*$.

Algorithm \ref{alg:SAFES} accommodates any DP types adopted by a DP synthesizer.
For example, if $\rho$-zCDP is used, then $\Theta_1=\{\rho\}$; 
if $(\varepsilon,\delta)$-DP or $\varepsilon$-DP is used, then $\Theta_1=\{\varepsilon,\delta\}$ or $\Theta_1=\{\varepsilon\}$. 
The immunity to post-processing property of DP allows examining and comparing different fairness parameters $\Theta_2$, if a fairness preprocessing approach involves such a parameter, to achieve a desired fairness-utility trade-off at no additional privacy cost, a a key advantage of SAFES’s sequential treatment of privacy and fairness.

\subsection{DP data synthesizer}
Algorithm \ref{alg:SAFES} is sufficiently general to permit any DP data synthesizer. 
We list two such procedures that are used in our experiments below.
The first is the Adaptive and Iterative Mechanism (AIM) procedure \citep{McKenna2022} and the second is Differentially Private Conditional Tabular Generative Adversarial Network (DP-CTGAN) \citep{Dp-ctgan}.  
AIM can provide strong utility for generated DP synthetic data but it only works for data with categorical or ordinal variables; if a dataset contains  continuous variables, discretization will be needed before applying AIM, resulting in information loss. 
DP-CTGAN does not have such restriction and can work with both continuous and discrete variables.

\paragraph{The AIM procedure}
AIM has been shown to consistently outperform other synthesis methods in various utility metrics in previous works \citep{McKenna2022}.
In addition, previous studies suggest that marginal-based methods, to which AIM belong, for large privacy loss do not exacerbate group unfairness as much as deep generative model-based DP synthesis methods \citep{Pereira2024}. 
In general, for marginal-based synthesizers, a workload 
$W=\{\bs\mu'_1, \bs\mu'_2, \ldots, \bs\mu'_k\}$ is first specified, where each $\bs\mu'_i$ for $i=1,\ldots,k$ represents a marginal of soem order on data $D$. 
The initial data distribution  estimate $\hat{p}$ (referred to as graphical model in \citep{McKenna2022}) is formed that minimizes the difference between its one-way marginals and the DP one-way marginals calculated from  data $D$. $\hat{p}$ is then adaptively updated with higher-order marginals given a set of marginals selected  from $W$ (possibly with replacement) with DP guarantees.  
When all the pre-specified privacy budget is fully used, the iteration stops and the most updated
$\hat{p}$ is sampled to generate synthetic data $D^*$ to release. 

The pseudo-code for AIM is illustrated in Algorithm \ref{alg:AIM} with $(\epsilon,\delta)$-DP and $\rho$-zCDP guarantees, but other DP types with the associated mechanisms can be used. 
The privacy budget spending scheme and update rule ($\sigma_{t+1}$ and $\xi_{t+1}$ on line 13 of Algorithm \ref{alg:AIM}) ensure that marginals measured in later iterations receive a greater portion of the privacy budget when necessary to ensure useful marginals are learned in each iteration $t$.
In particular, if the difference between the marginals produced by $\hat{p}_t$ and $\hat{p}_{t-1}$ is small, meaning little information is gained in iteration $t$, $\xi_{t+1}=2\xi_t$ and $\sigma_{t+1}=\sigma_t/2$.
See Algorithm 3 in the arXiv version of the AIM paper \citep{mckenna2022aim} for more details.
Considerations are also taken to ensure computational tractability for the graphical representation.
These iterations proceed until the privacy budget is exhausted, at which point the current graphical model is sampled to obtain the synthetic dataset.
\paragraph{The DP-CTGAN synthesizer}
DP-CTGAN builds on CTGAN’s conditional sampling and mode-specific normalization for continuous columns with DP guarantees through DP-SGD. Algorithm \ref{alg:DPCTGAN} summarizes the training steps. There are $m$ conditional draws given a particular level of a categorical variable within each iteration. For each $j\in\{1,\dots,m\}$, a categorical mask $M_j$ and conditional vector $\texttt{cond}_j$ are sampled to balance the discrete levels;  the GAN generator produces a fake sample $\hat r_j=G_{\Phi_{\mathcal{G}}}(z_j,\texttt{cond}_j)$ given a sample $z_j\!\sim\!\mathcal{N}(0,I)$, and the GAN critic is refined with $s$ condition-matched real–vs–synthetic comparisons via a conditional loss.
The generator is then updated with a loss function that combines a loss term aligned with $M_j$ for $j=1,\dots,m$ and an adversarial term against the current critic, and an optional gradient penalty term. 
Iterations continue until the pre-specified privacy budget is exhausted, after which the trained generator generates the synthetic dataset $D^*$.

\begin{algorithm}[H]
    \SetAlgoLined
    \DontPrintSemicolon
    \SetKwInOut{Input}{Input}
    \SetKwInOut{Output}{Output}
    \Input{Tabular data $D\!=\!(X, G, Y)$ with $d$ variables, marginal workload $W\!=\!\{\bs\mu'_1,\ldots,\bs\mu'_k\}$, privacy budget $(\varepsilon_0, \delta_0)\!\!$}
    \Output{Privacy-preserving synthetic dataset $D^*$}
    \BlankLine
    Initialize $\sigma_0$, $\xi_0$ \tcp{\small{conservative initialization is recommended}}
    Convert $(\varepsilon_0, \delta_0)$ to target $\rho_0$ in zCDP using the conversion theorem \ref{thm:conversion}\;
    $\hat{\boldsymbol{\mu}}_0\gets \boldsymbol{\mu}_0 + \mathcal{N}(\mathbf{0}, \sigma_0^2I)$, where $\boldsymbol{\mu}_0$ contains all 1-way marginals on $D$\;
    $\hat{p}_0\gets \arg \min_p \sum_{i=1}^d
    \norm{\boldsymbol{\mu}_{0,p}[i]-\hat{\boldsymbol{\mu}}_0[i]}_2^2/\sigma_0$, where $p$ denotes the unknown underlying distribution for data $D$ and $\boldsymbol{\mu}_{0,p}$ contains all 1-way marginals given $p$\;
    $\rho_\textnormal{used}\gets d/(2\sigma^2_0)$\;
     $w_j\gets\sum_{i=1}^k|\bs{\mu}'_j\cap \bs{\mu}'_i|$ for $j=1$ to $k$ \tcp{\small{$|\cdot|$ is the cardinality}}
    $t \gets 0$; $\sigma_{t+1}\gets\sigma_0$; $\xi_{t+1}\gets\xi_0$\;
    \While{$\rho_{\textnormal{used}}<\rho_0$}{
        $t\gets t+1$\;
        $\rho_\textnormal{used}\gets \rho_\textnormal{used} + \xi_t^2/8 + 1/(2\sigma^2_t)$ \tcp{\small{privacy loss accounting}}
        Determine a computationally feasible set of marginals $W^\prime\subseteq W$ and select $\bs\mu^\prime_j\in W^\prime$ via the exponential mechanism with 
        $u(D, \bs\mu'_j)\!=\!w_j(\!\|\bs\mu'_j\!-\! \bs\mu'_{j,\hat{p}_{t-1}}\!\|_1\!\!-\!(2/\pi)^{1/2}\sigma_t n_{\bs\mu'_j}\!)$ at privacy loss $\xi_t$, where $\bs\mu'_{j,\hat{p}_{t-1}}$ is the selected marginal but measured by $\hat{p}_{t-1}$
        \tcp{\small{$u$ favors marginals with larger improvement in expected error under $\hat{p}_{t-1}$ and higher-order marginals; $n_{\bs\mu'_j}$ is the number of cells in marginal $\bs\mu'_j$}}
       $\hat{\bs{\mu}}_t \gets\bs\mu'_j + \mathcal{N}(\mathbf{0}, \sigma_t^2I)$\;
        $\hat{p}_t\gets \arg \min_p \sum_{i=0}^t\norm{\bs\mu_{i,p}-\hat{\boldsymbol{\mu}}_i}_2^2/\sigma_i$ \tcp{\small{graphical model learning}}
        Update $\sigma_{t+1}$ and $\xi_{t+1}$ \tcp{\small{according to Algorithm 3 of AIM arXiv paper \citep{mckenna2022aim}}}
    }
    Sample $D^* = (X^*, G^*, Y^*)$ from $\hat{p}_t$\;
    \Return $D^*$\;
    \caption{The AIM procedure \citep{McKenna2022}}
    \label{alg:AIM}
\end{algorithm}

\begin{algorithm}[H]
    \SetAlgoLined
    \DontPrintSemicolon
    \SetKwInOut{Input}{Input}
    \SetKwInOut{Output}{Output}
    \Input{Tabular dataset $D$ with discrete columns $\mathbf{d}_c$,  batch size $m$, critic steps $s$,  learning rate $\alpha$,  clipping bound $b$, noise scale $\sigma$, gradient-penalty weight $\lambda$,  target privacy budget $(\varepsilon_0,\delta_0)$, privacy accountant $\mathcal{A}$}
    \Output{Privacy-preserving synthetic dataset $D^*$} 
    \BlankLine
    Preprocess $D$ (one-hot for discrete variables, mode-specific normalization for continuous variables)\;
    Initialize parameters $\Phi_{\mathcal{G}}$ and $\Phi_{\mathcal{C}}$ for generator $\mathcal{G}$ and critic $\mathcal{C}$ \;
    $\varepsilon_{\textnormal{used}}\gets 0$\;
    \While{$\varepsilon_{\textnormal{used}} < \varepsilon_0$}{
        \For{$j=1$ \KwTo $m$}{
        Construct conditional categorical mask $M_j$ over $\mathbf{d}$ and sample conditional vector $\texttt{cond}_j$\;
        Sample $z_j \sim \mathcal{N}(\mathbf{0}, I)$ and generate fake row $\hat{r}_j \gets \mathcal{G}_{\Phi_{\mathcal{G}}}(z_j,\texttt{cond}_j)$\;
        Sample real row $r_j \sim D \,|\, \texttt{cond}_j$\;

       
        Sample $\texttt{cond}_{k}^{(j)}$, $r_{k}^{(j)}$, and $\hat{r}_{k}^{(j)}$ at $k=1,\ldots, s$ \;
        Compute critic loss $L_{\mathcal{C}}^{(j)} \;\gets\; s^{-1}\sum_{k=1}^{s}\!\Big(\mathcal{C}_{\Phi_{\mathcal{C}}}(\hat r_{k}^{(j)}, \texttt{cond}_{k}^{(j)}) - \mathcal{C}_{\Phi_{\mathcal{C}}}(r_{k}^{(j)}, \texttt{cond}_{k}^{(j)})\Big) \,$ .
        \;
        Draw DP noise $\,\mathbf{e}_j \sim \mathcal{N}(\mathbf{0}, (\sigma b)^2 I)$;\;
        Update $\,\Phi_{\mathcal{C}} \gets \Phi_{\mathcal{C}} - \alpha\,\texttt{Adam}\Big(\nabla_{\Phi_{\mathcal{C}}} \big(L_{\mathcal{C}}^{(j)}\;+\;  \lambda\,L_{\mathrm{GP}}\big) \;+\mathbf{e}_j\Big)$ \tcp*{\small{$L_{\mathrm{GP}}$ is the optional gradient penalty term}}
        }                    
    Compute generator loss $L_{\mathcal{G}} \;\gets\; m^{-1}\sum_{j=1}^{m}\mathrm{CrossEntropy}(\hat{\mathbf{d}}_c[j], M_j)
        \;-\; s^{-1}\sum_{k=1}^sC_{\Phi_{\mathcal{C}}}\!\big(\hat r^{(s)}_{k}, \texttt{cond}^{(s)}_{k}\big)$
     \;
    Update $\Phi_{\mathcal{G}} \gets \Phi_{\mathcal{G}} - \alpha\,\texttt{Adam}(\nabla_{\Phi_{\mathcal{G}}} L_{\mathcal{G}})$\;
     Update $\varepsilon_{\textnormal{used}} \gets \mathcal{A}\!\left(\varepsilon_{\textnormal{used}},\, \delta_0;\, m, b, \sigma\right)$ \tcp*{\small{cumulative $(\varepsilon,\delta)$ via the accountant}}
    }
    Sample $D^*$ from $G_{\Phi_{\mathcal{G}}}$ \;
    \Return $D^*$\;
    \caption{The DP-CTGAN procedure with for $(\varepsilon,\delta)$-DP guarantees \citep{Dp-ctgan}}
    \label{alg:DPCTGAN}
\end{algorithm}

\subsection{Fairness-aware data preprocessing}
Similar to DP synthesizers, SAFES is general to permit any fairness data preprocessing procedures.
There are limited fairness-aware data preprocessing methods in the literature. 
We use two such procedures in our experiments.
The first transforms $(X,Y)\subset D$ but leaves $G$ unchanged \citep{Calmon2017}. 
The loss function used in procedure has to satisfy three constrains, as detailed later; for that reason and for convenience, we refer to the procedure as Triple-constrained Transformation (TOT).
The second is a reweighting (RW) procedure \citep{reweight} that keeps $(X,Y,G)$ unchanged but assigns instance weights so the weighted empirical distribution makes $G$ and $Y$ independent for downstream training or learning with $D$.

\paragraph{The TOT procedure}
TOT permits explicit control over the balance between group and individual fairness and also connects well with the broader statistical learning framework due to its probabilistic transformation.
It transforms $(X,Y)$ to $(\tilde{X},\tilde{Y})$, but leaves $G$ unchanged.
We summarize the main idea of TOT below and refer readers to the original paper \citep{Calmon2017} for more details and theoretical results.
TOT learns a randomized mapping $T$ in the form of a conditional distribution $q_{\tilde{X}, \tilde{Y}|X,Y,G}$ by solving an optimization problem that balances controlling discrimination between groups (group fairness), limiting distortion of individual observations (individual fairness), and maintaining a data distribution similar to the one before the transformation (utility preservation).
For group discrimination control, it requires $q_{\tilde{Y}|G}$ be similar for any two groups.
Specifically, 
\begin{equation}
\label{eq:discrimination_constraint}
    J(q_{\tilde{Y}|G}(y|g_1), q_{\tilde{Y}|G}(y|g_2))=|q_{\tilde{Y}|G}(y|g_1)/q_{\tilde{Y}|G}(y|g_2)-1| \le\eta_{y,g_1,g_2}
\end{equation}
for all $y$ and groups $g_1,g_2\in G$.
$\eta$ controls the trade-off between enforcing group fairness and maintaining statistical relationships in the input dataset; 
different $\eta$ can be used for different groups or response values.
We denote $\eta_{y, g_1, g_2}$ for different $\{y, g_1, g_2\}$  collectively by $\boldsymbol{\eta}$.

Enforcing group fairness via equation \eqref{eq:discrimination_constraint} risks changing certain individuals unrealistically.
To avoid impossible or unrealistic individuals being represented in $(\tilde{X},\tilde{Y})$, TOT limits the conditional expected distortion.
Let $\phi:(\mathcal{X}\times\mathcal{Y})^2\rightarrow\mathbb{R}$ be a nonnegative distortion function, where $\mathcal{X}$ and $\mathcal{Y}$ are the domains of $X$ and $Y$.
TOT requires
\begin{equation}
\label{eq:E_distortion_constraint}
    \mathbb{E}\left[\phi(\mathbf{x}, y, \tilde{\mathbf{x}}, \tilde{y}) | G=\mathbf{g}, X=\mathbf{x}, Y=y\right] \le c_{\mathbf{g},\mathbf{x}, y}
\end{equation}
for all $(\mathbf{g},\mathbf{x}, y)$.
If $\phi$ is binary-valued, equation \eqref{eq:E_distortion_constraint} reduces to the probability of an undesirable mapping
\begin{equation}
\label{eq:P_distortion_constraint}
    P\left(\phi(\mathbf{x}, y, \tilde{\mathbf{x}}, \tilde{y})=1 | G=\mathbf{g}, X=\mathbf{x}, Y=y\right) \le c_{\mathbf{g},\mathbf{x}, y},
\end{equation}
where 1 indicates the largest possible distortion -- an individual jumps from one group in the original to the other one after the transformation. A small $c_{\mathbf{g},\mathbf{x}, y}$ value would limit the probability of that  happening.
Similar to $\eta$, different values of $c_{\mathbf{g},\mathbf{x}, y}\in[0,1]$ may be specified for different groups or response values.
We denote the bounds $c_{\mathbf{g},\mathbf{x}, y}$ collectively as $\mathbf{c}$.
The distortion function $\phi$ is also customizable based on the context of the problem, and there is not necessarily a single best function for a given case.
For example,  $\phi$ that modifies an individual's age by several decades should output a high distortion value $\phi$, while  $\phi$ that does not change an individual's age would output a $0$ distortion value. 

Finally, to ensure the transformation does not drastically modify the distribution of the input dataset, TOT minimizes the total variation distance (TVD) as the loss function, which, for discrete attributes, is
\begin{equation}
\label{eq:utility_objective}
    \textnormal{TVD}(q_{\tilde{X}, \tilde{Y}}, q_{X,Y}) = \textstyle\frac{1}{2}\sum_{\mathbf{x},y}\left|q_{\tilde{X}, \tilde{Y}}(\mathbf{x},y) - q_{X,Y}(\mathbf{x}, y)\right|.
\end{equation}
Equations \eqref{eq:discrimination_constraint}, \eqref{eq:E_distortion_constraint}, and \eqref{eq:utility_objective}, plus the constraint that $q_{\tilde{X}, \tilde{Y}|D}$ needs to a proper distribution, give the optimization
\begin{align} \label{eq:fairness_optimization}
    & \underset{q_{\tilde{X}, \tilde{Y}|D}}{\text{minimize}} \textstyle\frac{1}{2}\sum_{\mathbf{x},y}\left|q_{\tilde{X}, \tilde{Y}}(\mathbf{x},y) - q_{X,Y}(\mathbf{x}, y)\right|,\; \text{subject to}\\
    & J(q_{\tilde{Y}|G}(y|g_1), q_{\tilde{Y}|G}(y|g_2))\le\eta_{y,g_1,g_2}, \notag\\
    &\mathbb{E}\left[\phi(\mathbf{x}, y, \tilde{\mathbf{x}}, \tilde{y}) | G=\mathbf{g}, X=\mathbf{x}, Y=y\right] \le c_{\mathbf{g},\mathbf{x}, y}.\notag
\end{align}
Equation \eqref{eq:fairness_optimization} can be solved via a standard convex solver like the embedded conic solver (ECOS) \citep{Domahidi2013} available in the Python CVXPY library \citep{Diamond2016}.

\paragraph{The RW method}
RW \cite{Kamiran2012} is a fairness data preprocessing method that debias the data distribution for specific ML tasks.
Unlike TOT, which learns a probabilistic transformation of $(X,Y)$, RW leaves $D=(X,Y,G)$ unchanged but assigns a weight to each record in $D$ so that, under the weighted empirical distribution, the sensitive attribute $G$ and the label $Y$ are approximately independent.
Let $\hat{P}$ denote the empirical probability and define
\begin{equation}\label{eqn:joint}
P(G=g, Y=c)\;\triangleq\;\hat{P}(G=g)\,\hat{P}(Y=c),
\end{equation}
implying $G$ and $Y$ are independent under equation \eqref{eqn:joint}. RW designs instance-level reweighting schemes so that $G$ and $Y$ are independent (i.e.,  equation \eqref{eqn:joint} holds) after the procedures. One such scheme is 
\begin{equation}
\label{eq:rw_weight}
W(g,c)\;=\;\frac{P(G=g, Y=c)}{\hat{P}(G=g, Y=c)}\,.
\end{equation}
Define the reweighted joint distribution 
$\hat P_{\,W}(G=g, Y=c)\;=\;W(g,c)\,\hat{P}(G=g, Y=c),$
which is equal to $P(G=g, Y=c)$ equation \eqref{eqn:joint}, ensuring $G$ and $Y$ is independent after the reweighting.

Classification models (e.g., weighted logistic regression or support vector machine) built upon the reweighted samples enjoy improved fairness.
RW is parameter-free, fast to compute, and does not re-assign individuals to different groups or relabel individuals with different outcomes.

\section{Experiments}
We run experiments using the SAFES procedure on three real-world datasets:  Adult \citep{Becker1996},  
 the Correctional Offender Management Profiling for Alternative Sanctions (COMPAS) \citep{ProPublica2016}, and German Credit \cite{german_credit} data. 
Each  dataset is publicly available for download.
We evaluate the three-way privacy-fairness-utility trade-offs in SAFES against data synthesis or preprocessing methods that focus on either privacy preservation or fairness enhance, but not both.
We also compare the results to those obtained when the original dataset, rather than a synthetic counterpart, is used directly (i.e., no privacy or fairness constraints, but still undergoing the preprocessed feature sets described in Section \ref{sec:data})
so as to directly separate and understand the privacy-fairness-utility trade-off as a function of $\varepsilon$ and $\eta$.

To evaluate the \emph{general utility} of the synthetic data via SAFES, we 1) compare the synthetic data with the original data by measuring the TVD between all one-, two-, and three-way marginals for the datasets and 2) perform a Kolmogorov-Smirnov (KS) test to evaluate the distributional similarity between the original and the synthetic datasets. TO evaluate \emph{ML utility}  of the synthetic data, We train a classifier  on the synthetic data, and calculate the prediction accuracy, false positive (FP) and false negative (FN) rates, F1 score, and the ROC AUC on a test data. To evaluate \emph{fairness}, we measure the COD of the synthetic data, the SPD, the AOC, and the FP and FN rate balance for the fairness of the trained classifier.

\subsection{Data}\label{sec:data}
We use the Adult data in three types of experiments: evaluating SAFES(AIM + TOT) for synthetic data utility and fairness when the original data $D$ contains only categorical variables (Section \ref{sec:AIM+TOT}), evaluating  SAFES(DP-CTGAN + RW)for synthetic data utility and fairness when $D$ contains a mixture of categorical and continuous variable (Section \ref{sec:CTGAN+RW}), and sensitivity analysis of the synthetic data to hyper-parameter specification (Section \ref{sec:sensitivity}). 
The COMPAS dataset is used for the SAFES(AIM + TOT) experiment in Section \ref{sec:AIM+TOT}.
We use the German credit dataset to demonstrate SAFES(DP-CTGAN + RW) on a dataset with a larger number of mixed categorical and continuous variables.

\paragraph{Adult} The Adult dataset is a subset of US 1994 Census income data on 48,842 individuals.
It is frequently used in both privacy and fairness literature due to the inclusion of sensitive variables (e.g., income) and the encoding of real-world discrimination (e.g., pay disparities based on gender/race).
We consider a subset of 4 features (race, sex, education, age) and one response variable $Y$ (income).
The favorable outcome for $Y$ is  \$50,000.
The protected attributes are $G$ = \{race, sex\} with white and male being the privileged groups, respectively.
These variables were preprocessed as follows.
All non-white races were collapsed into one non-white category; for the experiment in Sections \ref{sec:AIM+TOT} and \ref{sec:sensitivity}, age was compressed into decades; for the experiment in Section \ref{sec:CTGAN+RW}, it was kept as is; education below 11th grade and above a Bachelor's degree were combined into one category each.
Table \ref{tab:Adult_features} summarizes the variables.
\begin{table}[!htb]
\caption{Variables in the experiments on the Adult dataset.}
\label{tab:Adult_features}
\centering
\begin{tabular}{ll}
    \toprule
    Feature & Levels \\
    \midrule
    Race & \{White, Non-white\} \\
    Sex & \{Male, Female\} \\
    Age (years) & Categorical: \{17-26, 27-36, \ldots, 87-96\}; Continuous: [17, 96] \\
    Education & \{$<$11th grade, 11th grade, High school, Some college,\\
    & Associate's, Vocational, Bachelor's, Graduate\} \\
    Income & \{$>$\$50k, $\le$\$50k\} \\
    \bottomrule
\end{tabular}
\end{table}

\paragraph{COMPAS} This dataset contains criminal history information for 6,172 defendants.
While it is more commonly used in fairness literature, it also carries privacy concerns -- even inferring membership in the dataset is equivalent to disclosing that an individual was accused of a crime.
We consider a subset of five feature variables (race, sex, age, number of priors, and degree of charge) and one response variable $Y$ (recidivism within two years) on a subset of 5,278 African-American and Caucasian individuals. 
The favorable outcome for $Y$ is that the individual did not recidivate.
The protected attributes are $G$ = \{race, sex\} with Caucasian and female being the privileged groups, respectively. We discretized age and number of priors into 3 categories. 
Table \ref{tab:COMPAS_features} summarizes the variables.
\begin{table}[!htb]
\caption{Variables in the experiment on the COMPAS dataset.}
\label{tab:COMPAS_features}
\centering
\begin{tabular}{ll}
    \hline
    Feature & Levels \\
    \hline
    Race & \{Caucasian, African-American\} \\
    Sex & \{Male, Female\} \\
    Age & \{$<$25, 25-45, $>$45\} \\
    Charge degree & \{Felony, Misdemeanor\} \\
    Number of priors & \{0, 1-3, $>$3\} \\
    Recidivism & \{Yes, No\} \\
    \bottomrule
\end{tabular}
\end{table}

\paragraph{German Credit} This dataset contains 1000 loan applications and is widely used in fairness literature. 
It has 20 categorical and numerical attributes.  The response variable $Y$ is binary, indicating creditworthiness (Good vs.\ Bad). We take the protected attribute as sex, derived from the personal status and sex field by mapping \{male: divorced/separated, married/widowed, single\} to Male and \{female: divorced/separated/married, single\} to Female and treat Male as the privileged group.  As a preprocessing step, we rescale the attribute credit amount by dividing it with $1000$. Table \ref{tab:german_credit_features} in the Appendix summarizes the variables.

\subsection{SAFES (AIM + TOT) for categorical data}
\label{sec:AIM+TOT}
We use the Adult (discretized age) and the COMPAS data as listed in Table \ref{tab:Adult_features} and \ref{tab:COMPAS_features} for this experiment.
\paragraph{Privacy settings}
We set the workload $W$ in the AIM Algorithm \ref{alg:AIM} to be all two-way marginals in both datasets.
We pre-specify the overall privacy loss in term of $(\varepsilon, \delta)$-DP with a range of $\varepsilon$ given in Table \ref{tab:settings} and $\delta=10^{-9}=o(1/n)$. 
\begin{table}[!htb]
\caption{Privacy and fairness algorithmic settings. SAFES is compared to the other three cells on utility and fairness metrics.}
\label{tab:settings}
\centering
\begin{tabular}{@{}c@{\hspace{2pt}}|@{\hspace{2pt}}c@{\hspace{-12pt}}c@{}}
    \toprule
     & original  & DP $\varepsilon\in\{10^{-2}, 10^{-1.5}, \ldots, 10^{1}\}; \delta=10^{-9}$ \\
    \midrule
    original & none & privacy-preserving only\\
    fairness $\eta\in\{0.025, 0.1\}$ for Adult;  & \multirow{2}{*}{fairness-aware only} &  \multirow{2}{*}{SAFES}\\
    \{0.08, 0.15\} for COMPAS & & \\
   \hline
\end{tabular}
\end{table}

\paragraph{Fairness settings}
We examine two  fairness parameter $\eta$ values in each of the two datasets as listed in Table \ref{tab:settings}. The values were selected to create separation in the fairness metrics, allowing us to visualize how the metrics changes as $\eta$ varies. Other $\eta$ values could have been used and we would expect a monotonic transition of the metric values beyond or between the results obtained at our examined $\boldsymbol{\eta}$ values.
In the Adult data, we set $\phi(\mathbf{x}, y, \tilde{\mathbf{x}}, \tilde{y}) =3$   if education changes by more than one stage or if age changes by more than a decade, 2 if age changes by a decade, 1 if income decreases, and 0 otherwise.
In the COMPAS data, we set  $\phi_\textnormal{Age}$ and $\phi_\textnormal{Priors}$ to be 1 if the value is changed to the adjacent category and 2 if changed to a non-adjacent category (e.g., < 25 to > 45 for age); 
$\phi_\textnormal{Charge}$ outputs 1 if the value is changed from felony to misdemeanor and 2 if changed from misdemeanor to felony;
$\phi_\textnormal{Recidivism}$ outputs 2 if the value is changed; $\phi$ for each variable output 0 if there is no change.
For coding and interpretability reasons, in both datasets, we convert the $\phi$ functions described above into a binary $\phi$ as defined in equation \eqref{eq:P_distortion_constraint} with threshold values $(0.99, 1.99, 2.99)$ and  $\mathbf{c}=(0.1, 0.05, 0)$ corresponding to each threshold. 
The output values from $\phi$ are selected in accordance with common sense and societal norms, though alternative values can be adopted when suitable. 
The small $\mathbf{c}$ values implies that only minor changes to each record are permissible with small probability and large changes or changes to too many attributes are strongly discouraged.
For example, in the Adult data, the fairness preprocessing decreases income with probability $0.1$ and changes age by at most a decade with probability $0.05$, but is not allowed to change education by more than one stage or age by more than one decade.
Any other change is freely permitted.
We evaluate the sensitivity of the fairness-utility trade-off to different choices of $\mathbf{c}$ in Section \ref{sec:sensitivity}.

\paragraph{Experiment settings}\label{sec:algoaim}
For both the Adult and COMPAS data, we split the original dataset (after the preprocessing described in Section \ref{sec:data}) into a training set and a test set in an 75/25 ratio. 
Using the training data as input, synthetic data of the same sample size as the training data were generated via SAFES in Algorithm \ref{alg:SAFES} using AIM as the DP data synthesizer and TOT for the fairness transformation.
As the privacy- and fairness- aware transformations in SAFES are probabilistic, we run 35 repeats to summarize the average performance and stability, for each of combination settings defined by the privacy loss parameters $\varepsilon$ and fairness parameters $\eta$ listed in Table \ref{tab:settings}. We use the SmartNoise SDK \citep{SmartNoiseSDK} for the AIM subroutine in the experiments, and the AIF360 library \citep{Bellamy2018} for the TOT subroutine.
The randomness associated with DP occasionally makes the fairness transformation infeasible for small $\varepsilon$ (e.g., $\approx30\%$ of the attempts failed for $\varepsilon\in\{10^{-2}, 10^{-1.5}\}$ and $\eta=0.08$ for COMPAS).
Therefore, even though 35 repeats were attempted for each combination of $\varepsilon$ and $\eta$, a small number of the results presented below are summarized based on $<35$ repeats. For the downstream classifier trained on the SAFES synthetic data, we use  logistic regression. The fairness metrics are evaluated with the protected attribute being race and sex individually, as well as jointly (e.g., white male as the privileged group).

\begin{figure}[!htb]
\centering
    \includegraphics[width=1\textwidth]{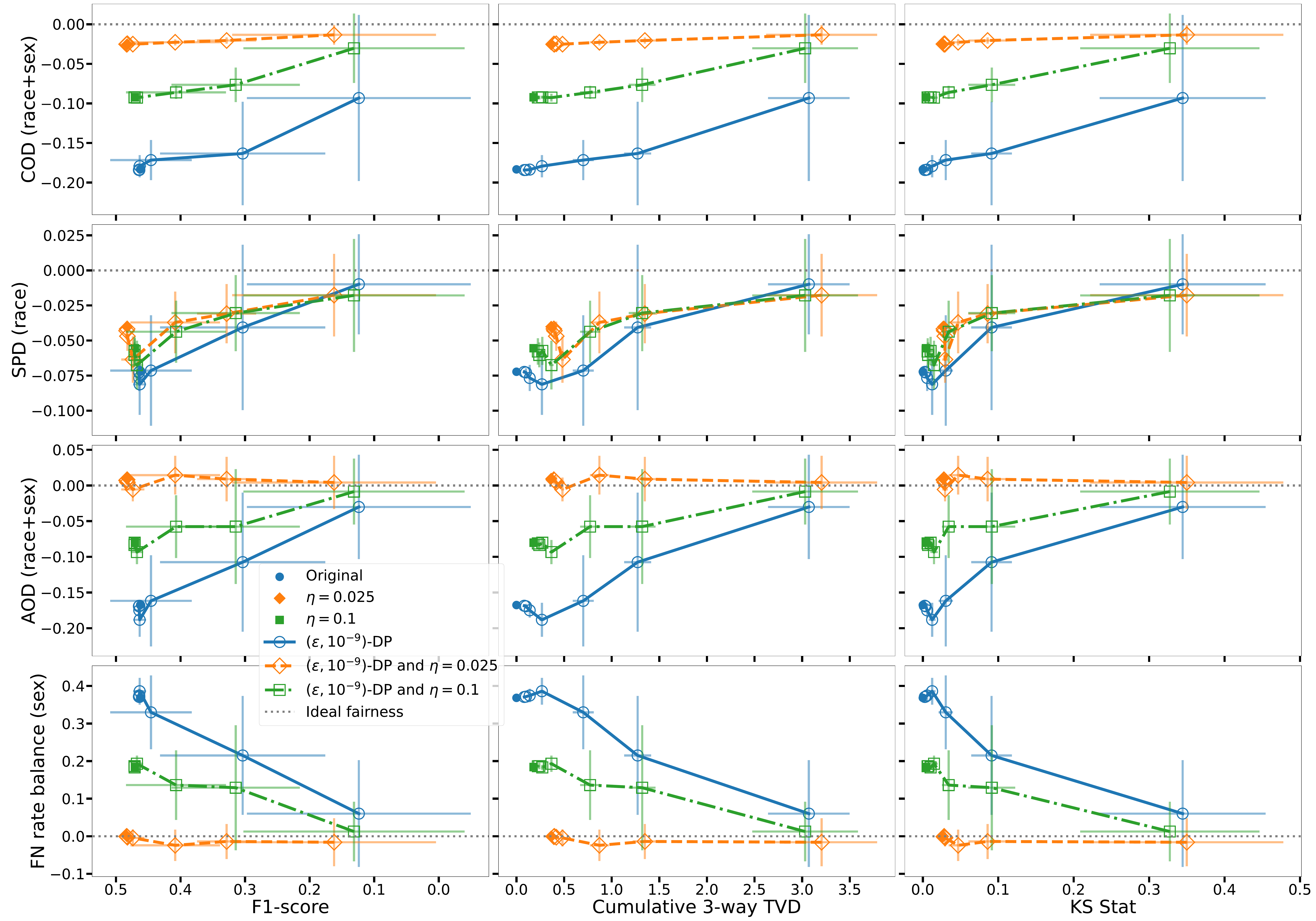}\vspace{-8pt}
    \caption{Examples of the privacy (points on each line) vs fairness (y-axis) vs utility (x-axis) trade-off in the Adult experiment.
    In each plot, each point on a line represents the mean and the error bar indicates $\pm1$ SD over 35 repeats at a different privacy loss $\varepsilon$ value $\in\{10^{-2}\; (\mbox{rightmost}), 10^{-1.5}, 10^{-1}, \ldots, 10\; (\mbox{leftmost})\}$; lines represent different fairness parameters $\eta$; $x$-axis values further left correspond to better utility.}
    \label{fig:Adult_utilityXfairness}\vspace{-9pt}
\end{figure}

\paragraph{Results}
Examination of the general utility, classification performance, and fairness metrics suggests similar insights across the board.
For this reason and to conserve space, we present the results for a representative subset of the utility and fairness metrics in Figs.~\ref{fig:Adult_utilityXfairness} and \ref{fig:COMPAS_utilityXfairness}.
Complete tabular numeric results and additional figures are found in the Appendix.
The observations from the figures on the privacy-fairness-utility trade-off are summarized below. The findings in (1) to (3) below can also be observed in Figs.~\ref{fig:Adult_TVD} - \ref{fig:compas_AOD} in the Appendix.
\vspace{-3pt}\begin{enumerate}
\item[(1)] As expected, stringent privacy (small $\varepsilon$) and fairness (small $\eta$) parameters result in higher general utility loss (larger marginal TVD and KS statistic) and worse classifier prediction performance (accuracy, F1 score, AUC, etc.).
Owing to its dual focus on privacy preservation and fairness improvement, SAFES-generated synthetic data tends to be inferior to privacy-only and fairness-only synthetic data in the utility metrics, following the expected three-way privacy-fairness-utility trade-off.
However, both experiments suggest that similar-to-baseline utility can be achieved with improved fairness metrics for $\varepsilon\le1$ (generally considered strong privacy), demonstrating the value of SAFES compared to privacy- or fairness-only approaches.
\item[(2)] The performance of the classifiers trained on SAFES synthetic data is similar to that obtained in other works on DP data synthesis using these datasets.
For example, the $\approx0.8$ accuracy in the Adult experiment (Figure  \ref{fig:Adult_pred} in the Appendix) matches the accuracy obtained by \citet{Pujol2023}; the $\approx0.7$  AUC in the COMPAS experiment (Figure  \ref{fig:compas_pred} in the Appendix) outperforms the AUC by \citet{Pereira2024}.
\item[(3)] DP guarantees at small $\varepsilon$ is more influential on general utility than the fairness enhancement quantified by $\eta$. For larger $\varepsilon$, $\eta$ has a more noticeable impact on general utility. The fairness transformation generally has little impact on the classifier prediction performance, with or without DP guarantees.
\item[(4)] Even for strong privacy guarantees and fairness constraints ($\varepsilon\in(0.05,1)$ and small $\eta$), synthetic data via SAFES have similar utility to the original by several metrics, including the TVD of three-way marginals. 
\item[(5)] In general, at very small $\varepsilon$, the SAFES synthetic data and the DP-only synthetic data display more fairness than the original.
Another way to interpret this is that the SAFES synthetic data do not outperform the DP-only synthetic data in the fairness metrics at small privacy budgets.
This is due to the large amount of noise injected for the DP guarantees, but comes at a cost of poor utility.
For $\varepsilon$ large enough to avoid this but still small enough to yield strong privacy guarantees, the synthetic data show significantly enhanced fairness for all three protected groups compared to the original that favors the privileged group, along with similar utility to the original.
\item[(6)] Fairness guaranteed by SAFES at small $\eta$ is robust to changes in privacy loss and stabilizes as $\eta$ decreases, allowing confident calibration of the privacy-utility balance while maintaining fairness. 
\end{enumerate}

\begin{figure}[!htb]
   \vspace{-6pt} \centering
    \includegraphics[width=1\textwidth]{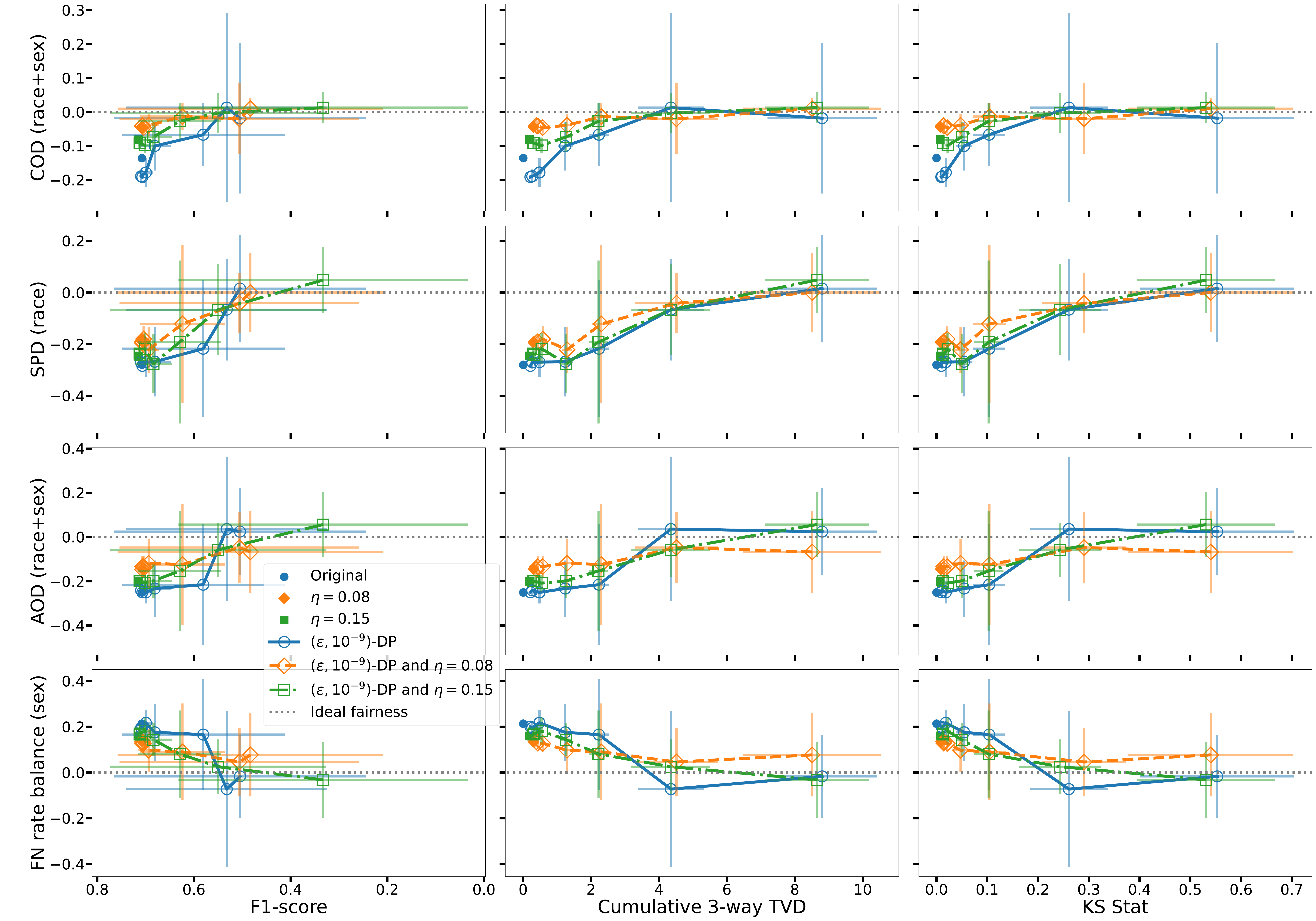}\vspace{-6pt}
    \caption{Examples of the privacy (points on each line) vs fairness (y-axis) vs utility (x-axis) trade-off in the COMPAS experiment.
    In each plot, each point on a line represents the mean and the error bar indicates $\pm1$ SD over 35 repeats at a different privacy loss parameter $\varepsilon$ value $\in\{10^{-2}\; (\mbox{rightmost}), 10^{-1.5}, 10^{-1}, \ldots, 10\; (\mbox{leftmost})\}$; lines represent different fairness parameters $\eta$; $x$-axis values further left correspond to better utility.}
    \label{fig:COMPAS_utilityXfairness}\vspace{-6pt}
\end{figure}

\begin{figure}[!htb]
    \vspace{-6pt}\centering
    \includegraphics[width=0.5\linewidth]{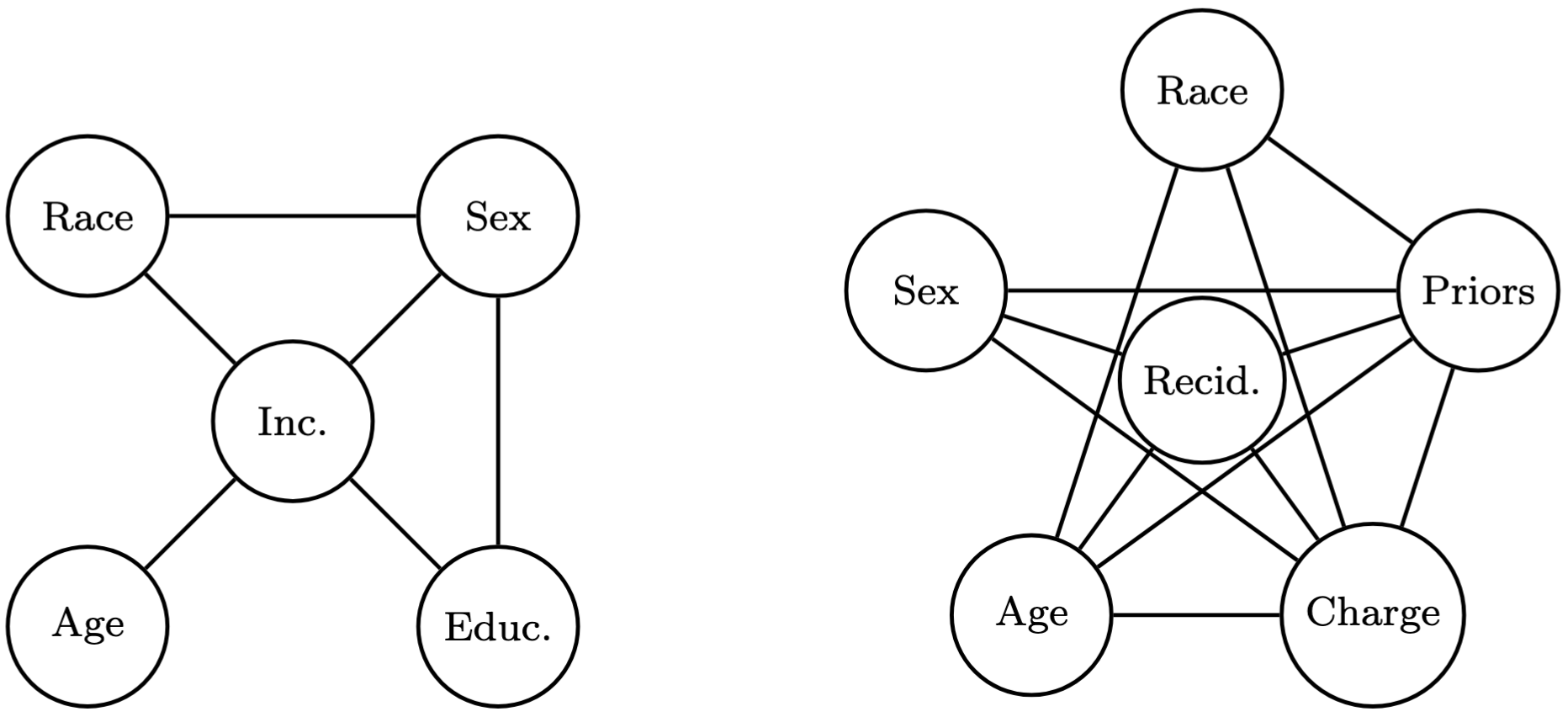}\vspace{-6pt}
    \caption{Examples of graphical model representations learned by the AIM algorithm for the Adult (left) and COMPAS (right) examples. 
    The Adult example was obtained with $\varepsilon=0.1$, while the COMPAS example was obtained with $\varepsilon=1$.}
    \label{fig:graphical_model_examples}\vspace{-9pt}
\end{figure}

Figs.~\ref{fig:Adult_utilityXfairness} and \ref{fig:COMPAS_utilityXfairness} suggest the FN rate balance in several cases is positive rather than negative.
At first glance, this seems to indicate that the  unprivileged sex group (female in Adult, male in COMPAS) is more likely to have the favorable outcome per this metric.
However, a positive value is rather an indication that more unprivileged than privileged individuals are incorrectly classified to the unfavorable outcome, which is still a disadvantageous result for the unprivileged group.
Additionally, when measuring COD jointly on race+sex in the COMPAS experiment, the COD of the DP-only synthetic data (solid blue line) does not appear to converge to the COD of the original dataset (solid blue dot).
This is an artifact of the graphical models constructed by the AIM subroutine in SAFES when selecting from only one- and two-way marginals, as we did in our evaluations.
Three-way relationships, if they exist, such as between race, sex, and recidivism, may not be fully captured.
Fig.~\ref{fig:graphical_model_examples} shows an example of this for the Adult and COMPAS experiments.

We note that the classification utility (accuracy, F1 score, and AUC) with the original data are lower than the state-of-the-art for these datasets, though they outperform a trivial classifier.
This is partly because the original data includes only a subset of attributes and the continuous attributes were discretized to provide a fair baseline for assessing the  utility loss from DP and fairness data preprocessings via AIM and TOT, both of which have known scalability limitations in terms of both number of attributes and the levels within an attribute and only work with categorical variables (thus the data preprocessing steps described in Section \ref{sec:data}, resulting in fewer and less refined predictors being available to fit classifiers than would be possible if AIM and TOT were not applied.

\subsection{SAFES (DP-CTGAN + RW) for data with mixed continuous and categorical data}\label{sec:CTGAN+RW}
We use the Adult data in Table \ref{tab:Adult_features} with continuous age and the German Credit data \ref{tab:COMPAS_features} for this experiment.

\paragraph{Privacy settings}
We use the standard GAN objective with binary cross-entropy for both the critic and the generator without the gradient penalty term in the DP-CTGAN algorithm.  The moment account method \cite{abadi2016deep} is used as the privacy accountant $\mathcal{A}$ in Algorithm \ref{alg:DPCTGAN}. 
We set the clipping bound $b=10$, batch size $m=500$, critic steps $s=1$, Adam learning rate $\alpha=2\times10^{-4}$. For the Adult data, we examine $\varepsilon\in\{10^{-0.5}, 10^{0}, \ldots, 10^{2}\}$ and $\delta=10^{-7}$; the noise multiplier $\sigma$ corresponding to different $\varepsilon$ values are as follows: $\sigma=25$ for $\varepsilon\le 0.5$; $\sigma=15$ for $0.5<\varepsilon\le 1.0$; $\sigma=3.5$ for $1.0<\varepsilon\le 3.5$; $\sigma=2.5$ for $3.5<\varepsilon\le 10$; $\sigma=0.9$ for $10<\varepsilon\le 50$; and $\sigma=0.5$ for $\varepsilon>50$. For the German Credit data, we examine one privacy loss at $\varepsilon=5$ and $\delta=5 \times 10^{-5}$, the corresponding  noise multiplier of which is $\sigma=3$.

\paragraph{Fairness settings}
Unlike TOT, RW has no hyperparameters. Instead, it reweighs the training data distribution through loss function formulation by attaching an instance weight $W(G,Y)$ in equation \eqref{eq:rw_weight} calculated from the data. 

\paragraph{Experiment settings}
In both the Adult (with continuous age) and German credit experiments, we split the original data (after the preprocessing steps described in Section \ref{sec:data}) into a training set and a test set in an 75/25 ratio. 
Synthetic data of the same sample size as the training set are generated via SAFES in Algorithm \ref{alg:SAFES} using DP-CTGAN as the DP data synthesizer; RW was applied for fairness when training the downstream classifier.
We run 35 repeats on the Adult data and 15 repeats on German credit data, and report the average performance and stability across the runs.
We use the SmartNoise SDK \citep{SmartNoiseSDK} for the DP-CTGAN subroutine in the experiments, and the AIF360 library \citep{Bellamy2018} for the RW subroutine. We train a logistic classifier on the DP synthetic data with RW and evaluate the prediction performance on the test set using the same set of metrics as in Section \ref{sec:AIM+TOT}. 
For fairness evaluation, no COD is calculated as RW does not change the data; we calculate SPD, AOC, and the FP/FN rate balance for the trained logistic classifier with RW, where the protected attributes are race and sex separately and jointly in the Adult data and the protected attribute is sex in the German Credit data.

\paragraph{Results}
A subset of the results for the Adult data are shown in Figs.~\ref{fig:Adult_continuous_utilityXfairness} and \ref{fig:Adult_cont_TVD_KS}; the rest can be found in the Appendix.  Since RW does not alter the data, thus does not influence data-level metrics like TVD, Fig.~\ref{fig:Adult_cont_TVD_KS} visualizes only the privacy-utility (i.e., ignoring fairness) trade-off for the general utility metrics.
As expected, more stringent privacy decreases utility. 
 Fig.~\ref{fig:Adult_continuous_utilityXfairness} also shows that RW consistently improves fairness of the downstream classifier.  
The results for the German Credit experiment are presented in Table \ref{tab:German_credit}. First, RW for fairness barely affects the ML utility (Accuracy and F1 score) of the data but notably improves the fairness metrics (SPD and AOD) (the 2nd compared with the 1st rows in Table \ref{tab:German_credit}). Second, synthetic data via CTGAN, with and without DP, further improve the fairness on top of RW (the 3rd, 4th rows compared with 2nd rows). Third, DP reduces the ML utility of the synthetic data by 5\% to 7\% (the 4th vs.~3rd rows), but the decrease in the ML utility in the synthetic data generated by CTGAN -- without DP (the 3rd vs.~2nd rows) is much more pronounced, representing a major challenge for GAN-based DP synthesizers in general. 
This is also reflected in Figs.~\ref{fig:Adult_continuous_utilityXfairness} and \ref{fig:Adult_cont_TVD_KS}, as well as noted in previous work \citep{Pereira2024}. 
That is, it is challenging to achieve near baseline utility with reasonable DP guarantees for GAN-based methods -- even the largest privacy budgets in our experiments leave a noticeable utility gap compared to the baseline results.
Lastly, in both the Adult and German Credit experiments, improved fairness due to DP noise (at the cost of low utility) is also present, as it was in the AIM + TOT experiments.
\begin{figure}[!htb]
   \vspace{-6pt} \centering
    \includegraphics[width=1\textwidth]{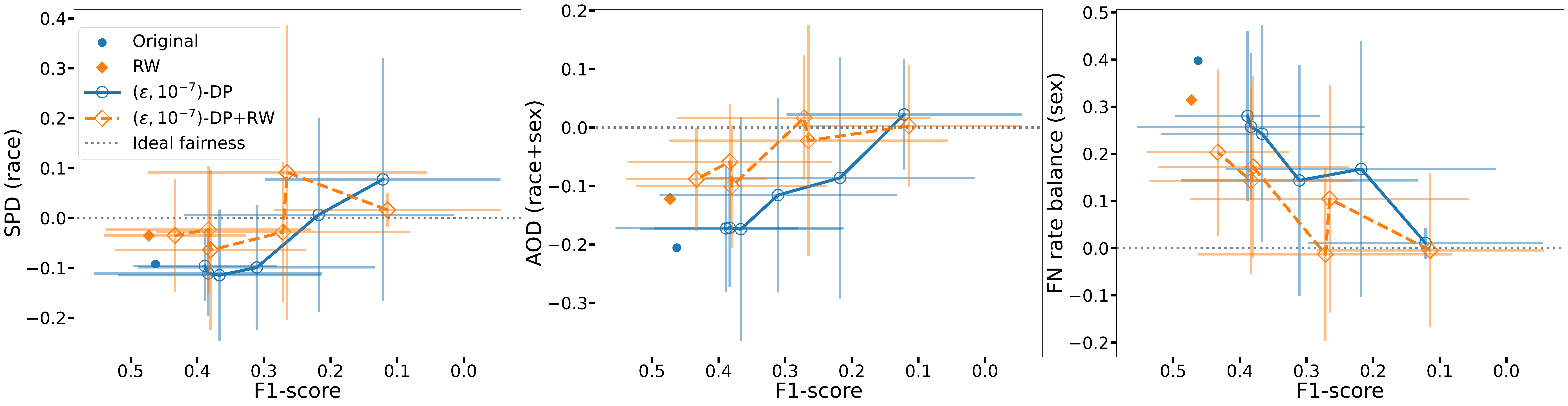}\vspace{-6pt}
    \caption{Examples of the privacy (points on each line) vs fairness (y-axis) vs utility (x-axis) trade-off in the DP-CTGAN + RW experiment on the Adult data (with continuous age).
    In each plot, each point on a line represents the mean and the error bar indicates $\pm1$ SD over 35 repeats at a different privacy loss parameter $\varepsilon$ value $\in\{10^{-0.5}\; (\mbox{rightmost}), 10^{0}, 10^{0.5}, \ldots, 10^{2}\; (\mbox{leftmost})\}$; the orange dashed and blue solid lines correspond to with and without RW, respectively; for consistency with Figs.~\ref{fig:Adult_utilityXfairness} and \ref{fig:COMPAS_utilityXfairness}, $x$-axis values further left correspond to better utility.}
    \label{fig:Adult_continuous_utilityXfairness}\vspace{-3pt}
\end{figure}

\begin{figure}[!htb]
    \centering
    \includegraphics[width=0.275\linewidth]{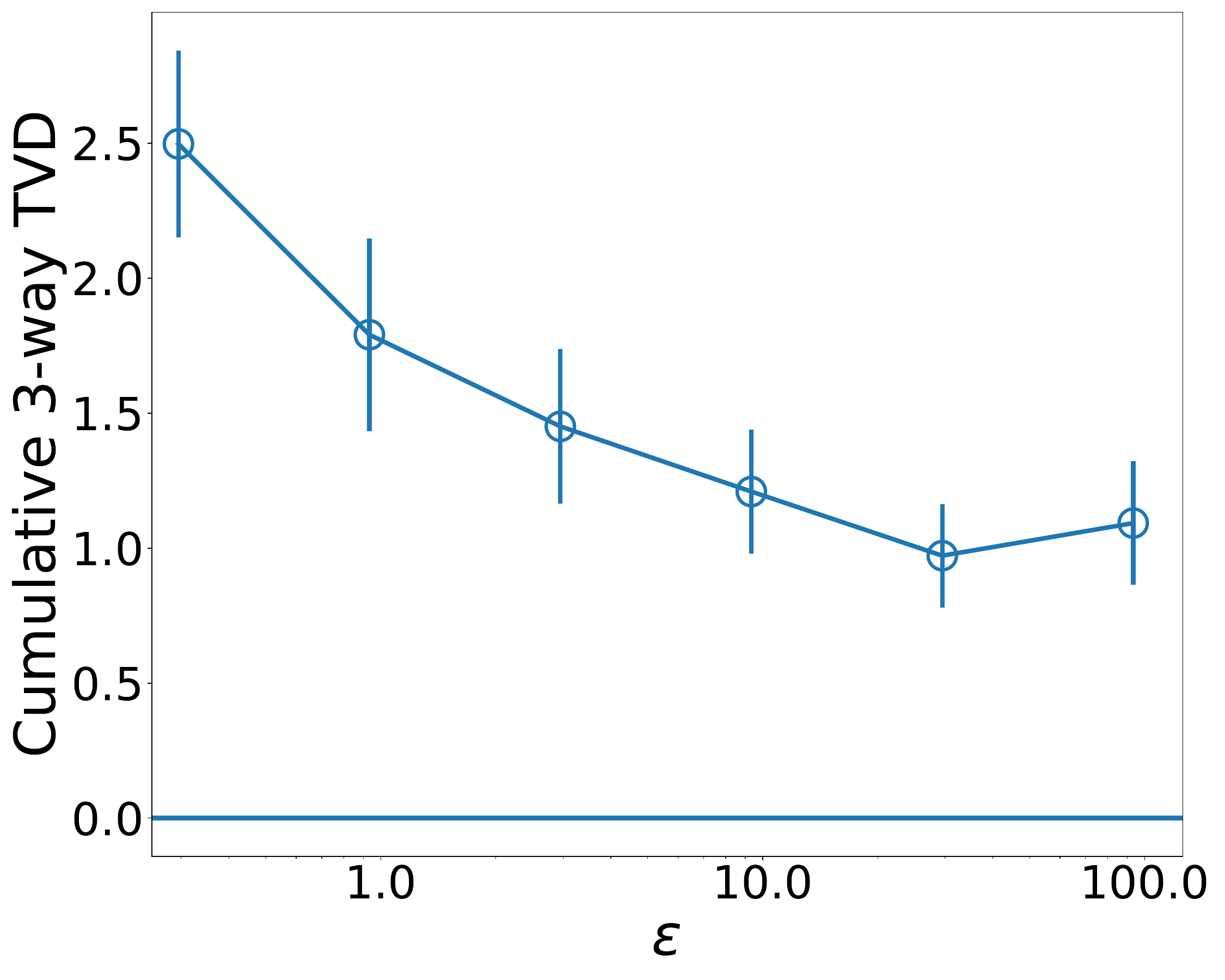}
    \includegraphics[width=0.275\linewidth]{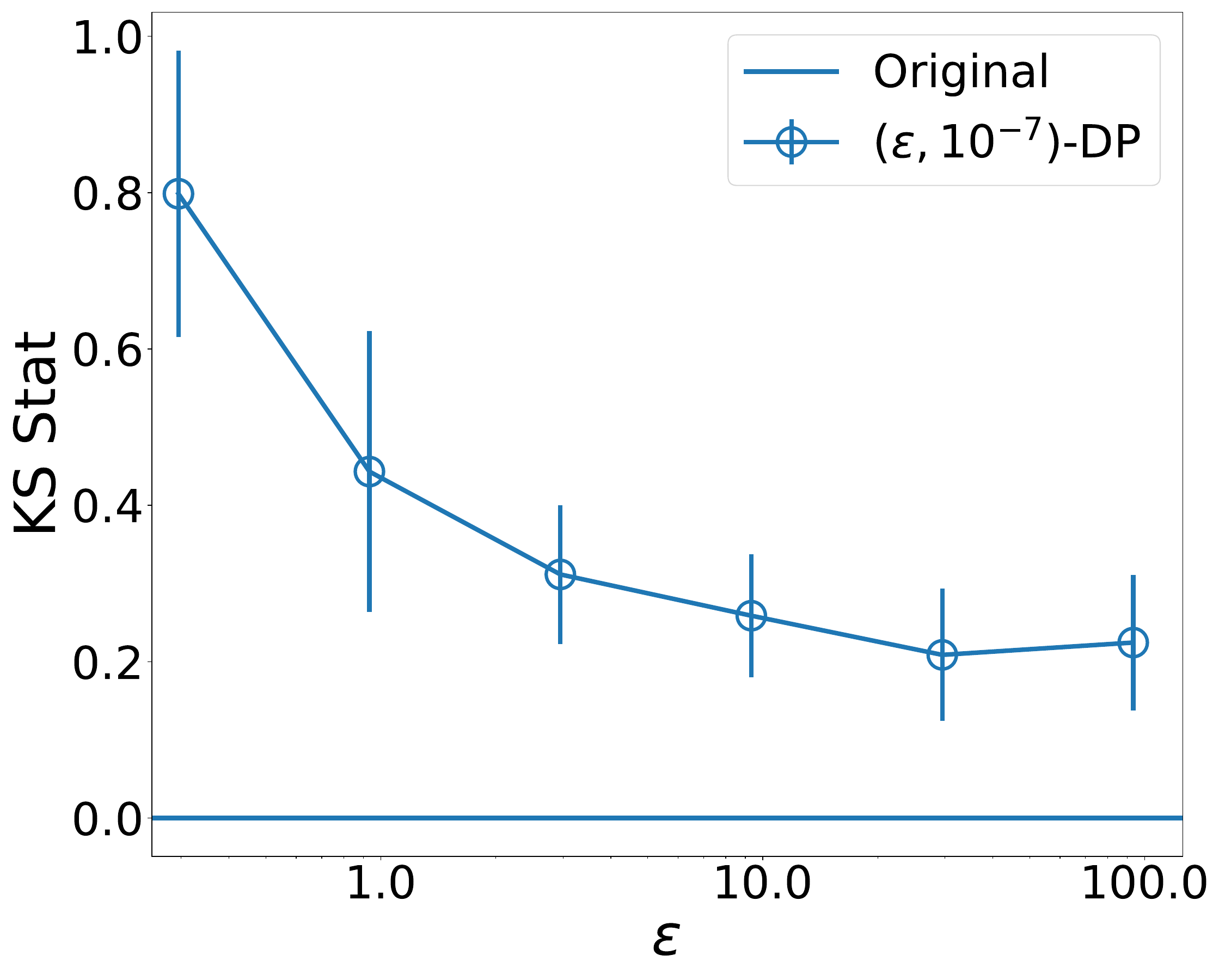}
    \vspace{-6pt}\caption{Mean $\pm$ 1 SD (error bars) 3-way TVD and KS test statistic comparing original and synthetic datasets generated by SAFE(DP-CTGAN + RW) in the Adult data with continuous age.}
    \label{fig:Adult_cont_TVD_KS}\vspace{-6pt}
\end{figure}

\begin{table}[!htb]
\caption{Mean (SD) ML utility and fairness metrics over 15 repeats for logistic regression classifier trained on synthetic German Credit data with and without DP and fairness RW.}
\label{tab:German_credit}
\centering
\begin{tabular}{r|rrrr}
    \toprule
    Synthesizer & Accuracy & F1 score & SPD & AOD \\
    \midrule
    Original & 0.764 (0) & 0.837 (0) & -0.187 (0) & -0.146 (0) \\
    RW only & 0.760 (0) & 0.833 (0) & -0.127 (0) & -0.087 (0) \\
    CTGAN (no DP)+RW & 0.613 (0.085) & 0.728 (0.104) & 0.033 (0.067) & 0.026 (0.064) \\
    DP-CTGAN+RW (\textbf{SAFES}) & 0.572 (0.117) & 0.651 (0.186) & 0.020 (0.068) & 0.018 (0.057) \\
    \bottomrule
\end{tabular}
\end{table}

Overall, although DP-CTGAN permits non-categorical variables and does not suffer from the same scaling issues as the AIM + TOT approach for larger numbers of features, the utility of the synthetic data suffer, consistent with previous findings on GAN-based DP synthesizers.
It is important to keep this in mind when using SAFES or a DP data synthesizer on data with continuous variables, as discretizing the variables may offer better privacy- utility tradeoff than adopting a GAN-based DP synthesizer.

\subsection{Sensitivity analysis}
\label{sec:sensitivity}
Each algorithm -- whether privacy-preserving or fairness-oriented -- with the exception of the fairness RW method, involves hyperparameters. In particular, the TOT procedure has several domain hyperparameters, and DP-CTGAN also includes procedural hyperparameters (e.g., learning rate, batch size) that are generic to optimization and not specific to the privacy–fairness–utility trade-off emphasized in this paper. The experimental results in Sections~\ref{sec:AIM+TOT} and \ref{sec:CTGAN+RW} illustrate how the privacy-loss parameter $\varepsilon$ and the TOT fairness parameter $\eta$ (Section~\ref{sec:AIM+TOT}) shape the privacy–fairness–utility trade-off. Additional TOT domain hyperparameters that may affect this trade-off include the distortion function $\phi$ and the bounds on the probabilities of allowed transformation. 
Given $\phi$ is a dimensionless function with infinitely many possible specifications, we focus on analyzing the sensitivity of the fairness-utility trade-off with respect to $\mathbf{c}$ below.

\begin{figure}[!htb]
\centering
\small (a)  F1 score \raisebox{-3cm}{\includegraphics[width=0.8\textwidth, trim=0pt 6pt 0pt 0pt, clip]{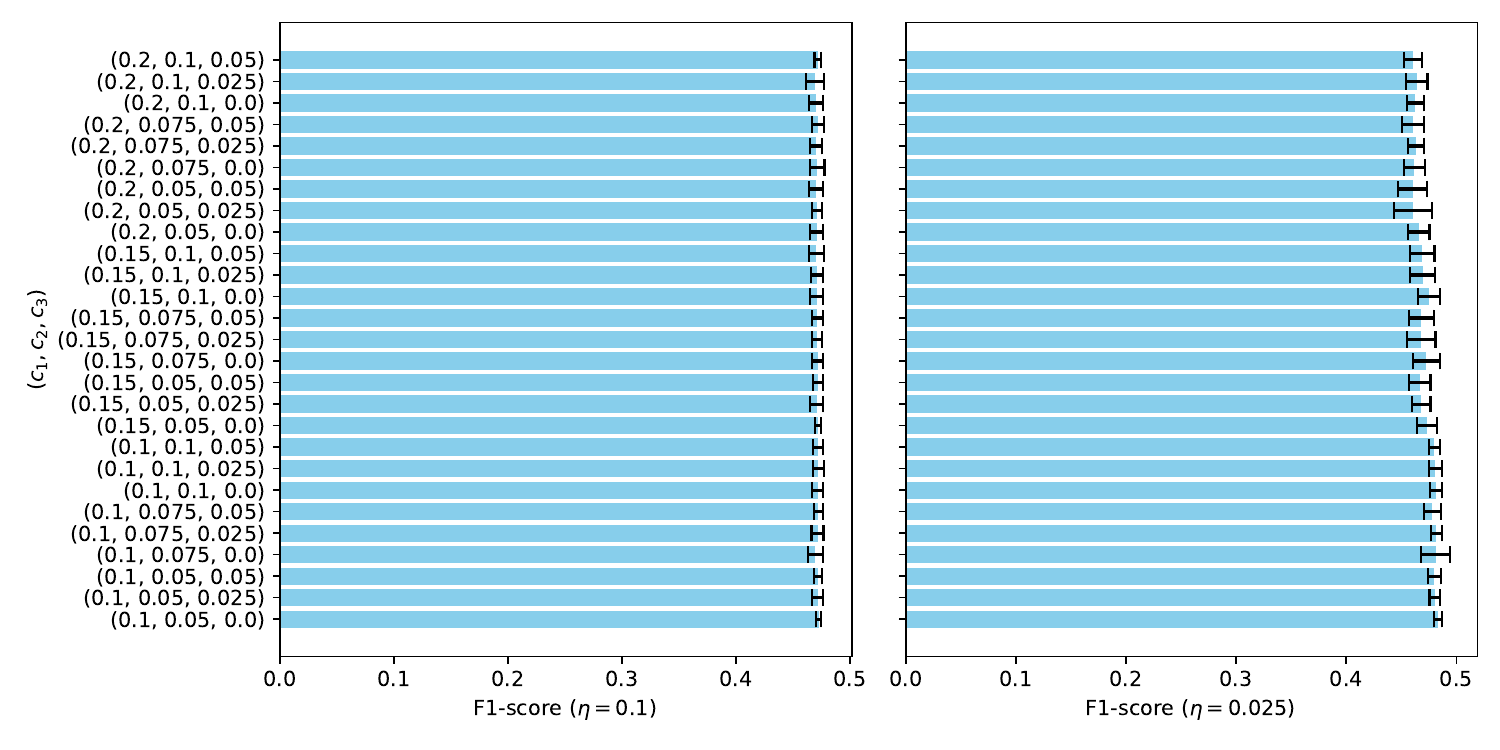}}\\
\small (b) 1-way TVD \raisebox{-3cm} {\includegraphics[width=0.8\textwidth, trim=0pt 6pt 0pt 0pt, clip]{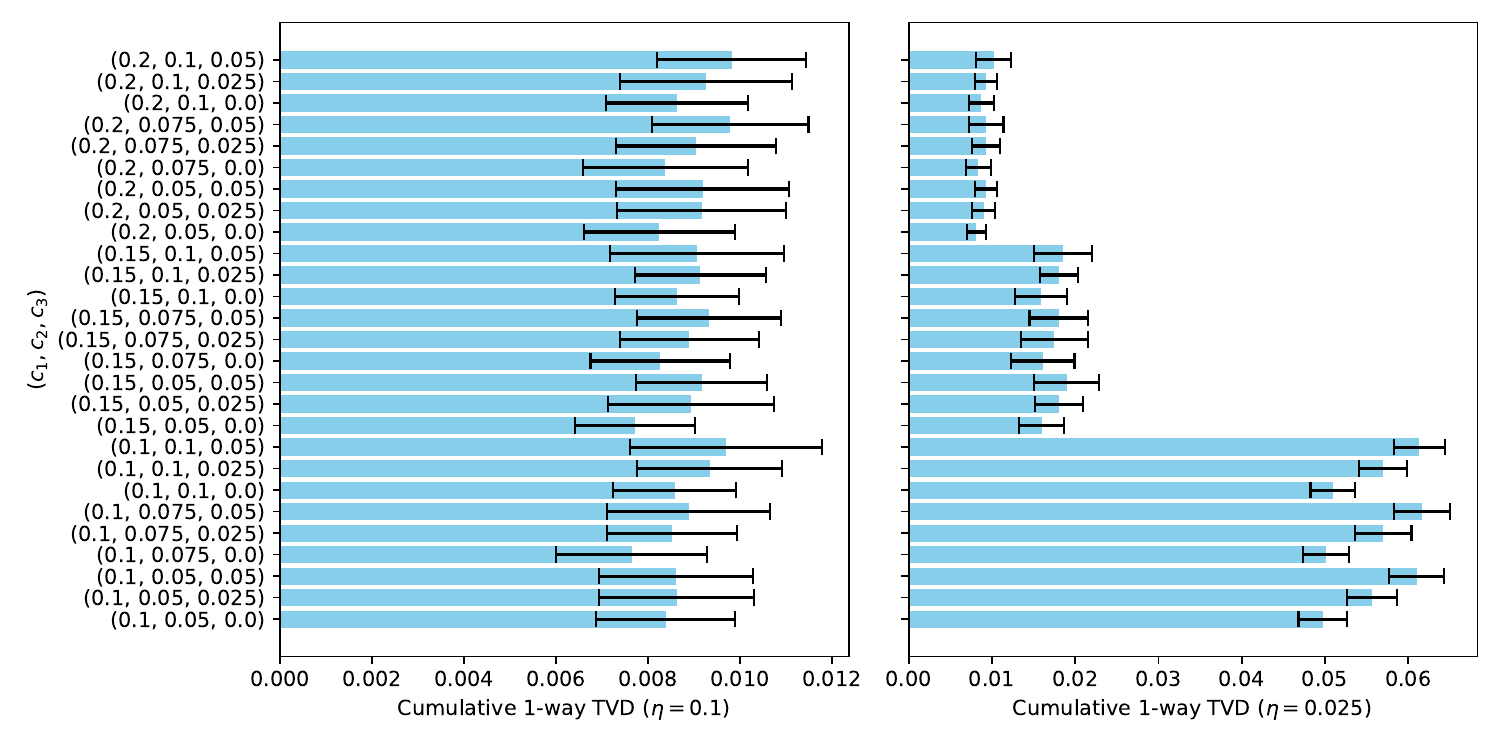}}\\
\small (c)  3-way TVD \raisebox{-3cm}{\includegraphics[width=0.8\textwidth, trim=0pt 6pt 0pt 0pt, clip]{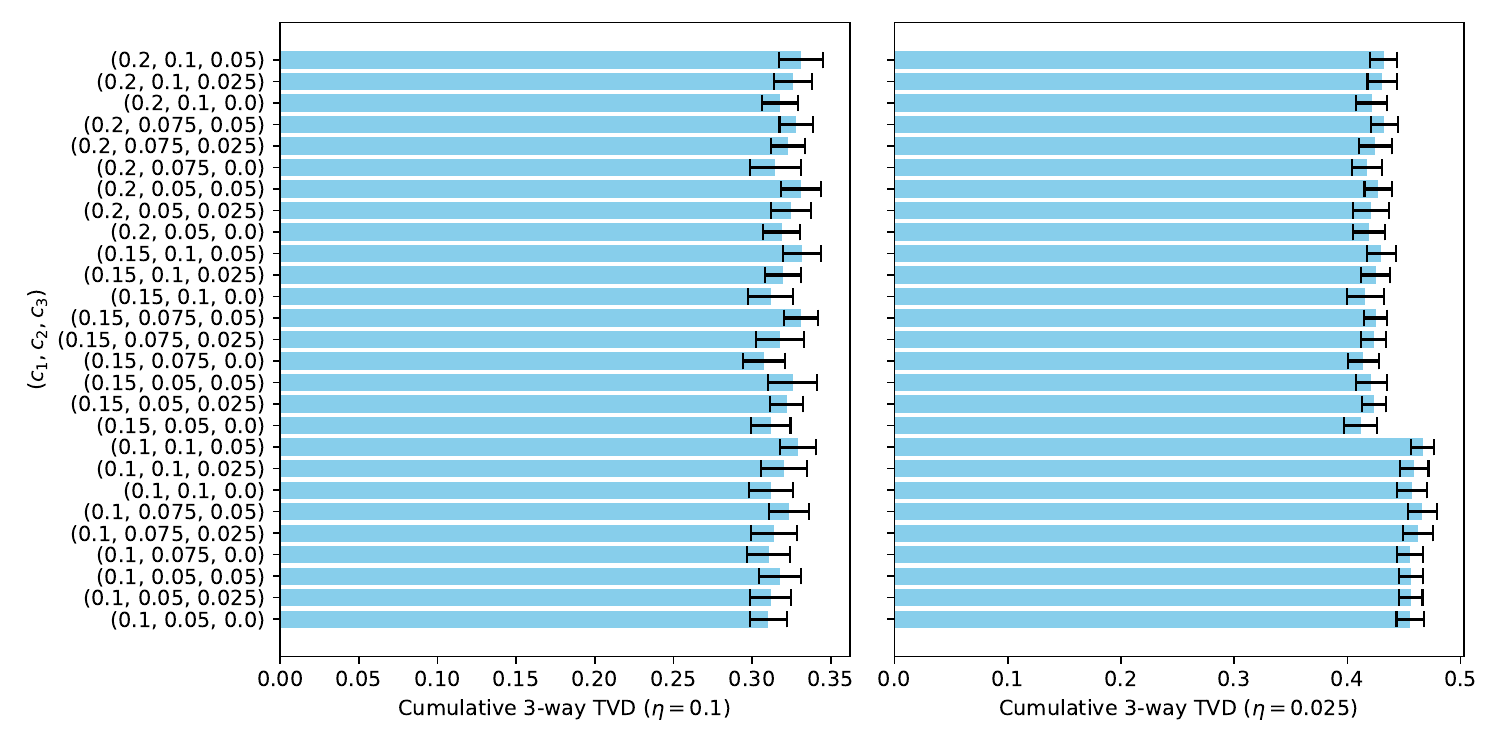}}\\  
\caption{Mean $\pm$ SD (error bars) of F1 score (ML utility) and TVD (general utility) over 35 synthetic datasets generated via SAFES (AIM + TOT) at various values of $\mathbf{c}=(c_1,c_2,c_3)$ in equation \eqref{eq:P_distortion_constraint} in the Adult dataset.}
\label{fig:TVD_sensitivity}
\end{figure}

Specifically, we fix privacy loss $\varepsilon$ at 1, keep the distortion function the same as in Section~\ref{sec:AIM+TOT}, and generate 35 datasets via SAFES (AIM + TOT) for each combination of $c_1\in\{0.2, 0.15, 0.1\}, c_2\in\{0.1, 0.075, 0.05\}, c_3\in\{0.05, 0.025, 0\}$, where $(c_1, c_2, c_3)=\mathbf{c}$, and $\eta \in \{0.1, 0.025\}$, leading to a total of $3\times3\times3\times 2=54$ scenarios.  The results of the sensitivity analysis for a subset of utility metrics are presented in Fig.~\ref{fig:TVD_sensitivity} (other general utility metrics, and performance metrics of a classifier trained on the synthetic data exhibited similar trends). 

\begin{figure}[!htb]
\centering
\includegraphics[width=0.8\textwidth, trim=0pt 6pt 0pt 0pt, clip]{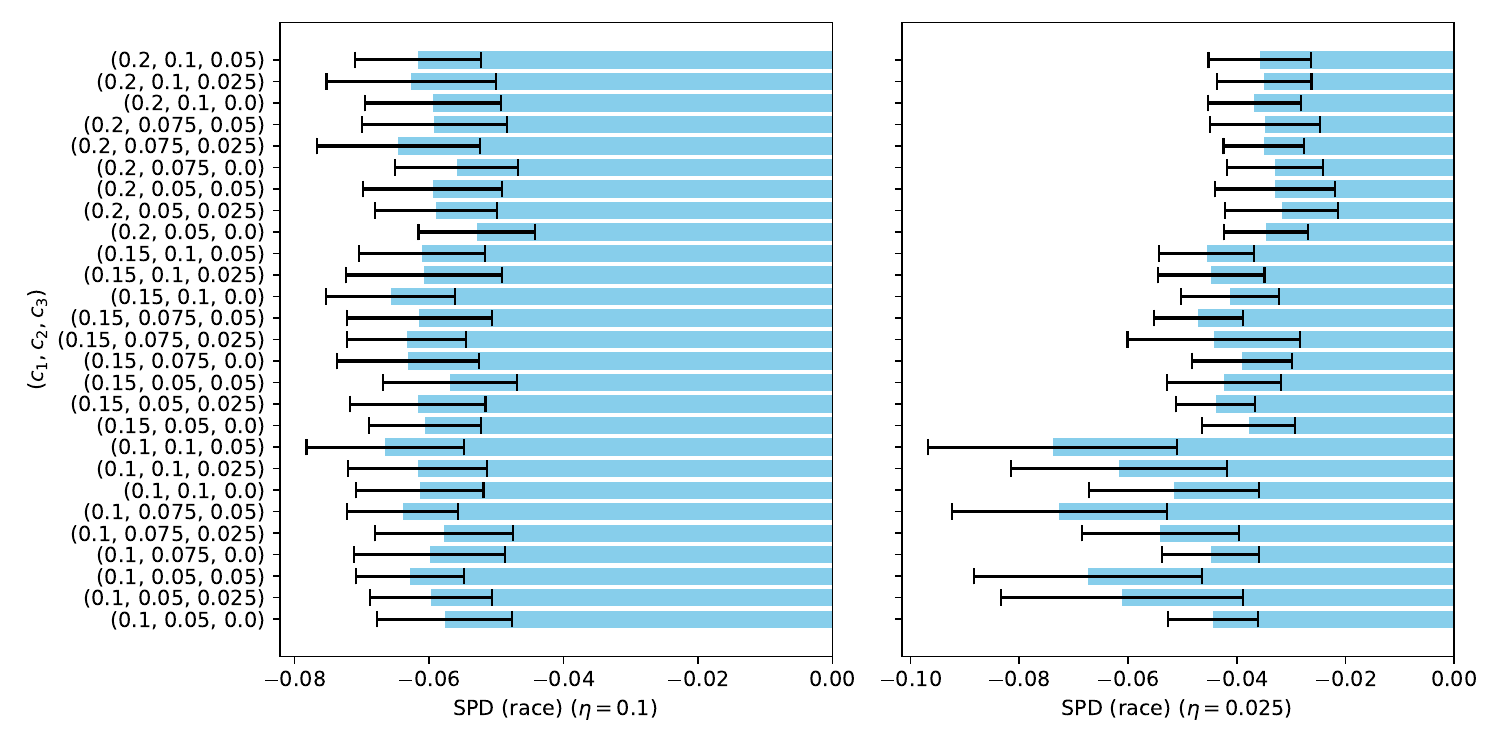}
\caption{Mean $\pm$ SD (error bars) SPD (fairness)  over 35 synthetic datasets generated via SAFES (AIM + TOT) at various values of $\mathbf{c}=(c_1,c_2,c_3)$ in equation \eqref{eq:P_distortion_constraint} in the Adult dataset.}
\label{fig:SPD_sensitivity}
\end{figure}

The utility at $\eta=0.1$ -- whether that's general utility measured by TVD or the ML utility measured by the F1 score of the downstream classification -- show robustness to varying $\mathbf{c}$ value.
For the more stringent fairness  setting of $\eta=0.025$, we observe a few trends.
First, increasing $c_1$ (the probability of adjusting income from $>\$50k$ to $\ge$\$50k) has negligible impact on the ML utility of the downstream classifier, as seen in Fig.~\ref{fig:TVD_sensitivity}(a).
For the general utility metrics measured by TVD, changing $\mathbf{c}$ affects the 1-way TVD in Fig.~\ref{fig:TVD_sensitivity}(a), but not the 3-way TVD presented in Fig.~\ref{fig:TVD_sensitivity}(b). In the case of 1-way TVD, increasing $\mathbf{c}$ (the probability of allowing small or large changes) in general decreases the utility,  as expected, except a surprising observation that increasing $c_1$ (a greater probability of small changes) improves the 1-way TVD.
We attribute this to TOT's ability to optimize fairness by favoring numerous small adjustments over fewer large ones, thereby reducing the need for substantial transformations that may more negatively affect the general utility of the data.
This conjecture is confirmed by the lower sensitivity observed in the 3-way TVD that jointly measuring across distributional differences over several variables in Fig.~\ref{fig:TVD_sensitivity}(b). This finding suggests a potential pathway to robust utility in TOT-transformed data, achieved alongside improved fairness (see Fig.~\ref{fig:SPD_sensitivity}, which shows that increasing $c$ in general improves fairness measured by SPD,  as expected, especially at the more stringent fairness requirement of $\eta=0.025$. 


\subsection{Runtime for SAFES (AIM + TOT)}
As stated in previous sections, the main reasons for us to work with a subset of attributes from the Adult and COMPAS datasets for the AIM + TOT experiments in Section \ref{sec:AIM+TOT} is that both AIM and TOT face computational bottlenecks as the number of attributes increases. 
To further explore and quantify the computational scalability of SAFES (AIM + TOT) as implemented in our experiments, we evaluate the average runtime required to generate a synthetic dataset of different sample sizes and numbers of features, as displayed in Table~\ref{tab:runtime}.
We observe that the runtime grows slowly with the sample size, but grows exponentially with the number of variables. 
This is a known limitation of both the AIM DP synthesizer and the TOT fairness data processor.
For AIM, the number of marginals and marginal cell counts grow exponentially with the number of variables and categories of each variable. 
For TOT, the number of constraints in the optimization increases with the data dimensionality.
\begin{table}[!htb]
\caption{Mean (SD) run time (seconds) over 30 runs of SAFES (AIM + TOT) on the Adult data of various sample sizes and attribute sets.}
\label{tab:runtime}
\centering
\begin{tabular}{r|rrr}
    \toprule
    & \multicolumn{3}{c}{Features} \\
    \cline{2-4}
    & \multicolumn{1}{c}{\{education, race,} & \multicolumn{1}{c}{\{education, race,} & \multicolumn{1}{c}{\{education, race,}\\
    Sample size & \multicolumn{1}{c}{income\}} & \multicolumn{1}{c}{sex, income\}} & \multicolumn{1}{c}{age, sex, income\}} \\
    \midrule
    10000 & 6.748 (0.821) & 11.493 (0.864) & 21.032 (1.869) \\
    20000 & 6.581 (0.398) & 12.512 (1.199) & 22.255 (1.488) \\
    30000 & 6.911 (0.487) & 12.618 (1.231) & 25.617 (1.809) \\
    40000 & 6.766 (0.272) & 12.427 (0.783) & 29.450 (3.919) \\
    48842 & 7.163 (0.524) & 12.459 (0.823) & 34.591 (4.307) \\
    \bottomrule
\end{tabular}
\end{table}

\color{black}
\section{Conclusions and discussion}
We have presented the SAFES procedure, which synthesizes data that simultaneously achieve DP guarantees and satisfy fairness constraints.
We have run extensive experiments on three real datasets to evaluate the privacy, fairness, and utility trade-off in SAFES.
The results demonstrate that in certain contexts, synthetic data generated by SAFES can enjoy strong DP guarantees and improved fairness for downstream classifier without significantly degrading the general utility of the dataset or constraining specific downstream learning tasks.


SAFES is a general framework. We demonstrated its usage with the AIM and DP-CTGAN synthesizers and the TOT and reweighting fairness data processor. Users of SAFES can employ any suitable DP synthesizer and fairness preprocessing method. 
Compared with the wide range of available DP synthesizers, fairness preprocessing methods are far less numerous. 
From a probabilistic ML perspective, we regard TOT as a principled approach despite its limitation in computational scalability; other data preprocessing methods for fairness tend to be more ad hoc and lack a strong probabilistic foundation. A comparison of different DP synthesizers and fairness data preprocessing methods is beyond the scope of this work but would be an interesting topic for future research.

For practical implementation of SAFES, we provide the following guidelines for choosing $\varepsilon$ for a DP synthesizer and $\eta$ in the TOT procedure. 
We recommend fixing the privacy parameter $\varepsilon$ first according to the desired privacy requirements, prevailing expectations, or common practice in the DP community, prior to preprocessing any sensitive dataset.
For example, although $\varepsilon\le1$ is generally considered strong privacy guarantees, larger values of $\varepsilon$ have been used by government agencies or tech company to balance privacy and utility (e.g., $\varepsilon=19.61$ by the US Census Bureau \citep{USCensus}, $\varepsilon=2,8$ by Apple \citep{Apple2017}, $\varepsilon=13.69$ by Google \citep{Xu2024}, etc.~\citep{Desfontain2024}).
For $\eta$ in the TOT procedure, our experiments indicate that the loss in utility as $\eta$ decreases is relatively small.
A practical strategy is therefore to choose choose the smallest $\eta$ that still permits a feasible optimization (i.e., the problem remains solvable).
That said, some applications may experience significant utility loss at small $\eta$, making the fairness–utility trade-off problem-specific. In such cases, the DP synthetic dataset produced by SAFES’s DP synthesizer can be used to test a range of $\eta$ values and identify an appropriate trade-off at no additional privacy loss.

In future work, we hope to develop methods to improve the scalability of SAFES to more complex data settings and diverse data types. 
Our experiment on the sensitivity analysis suggests that certain values of $\mathbf{c}$ in the TOT procedure seem to simultaneously improve fairness and utility.
A more systematic investigation of this trend would be a promising direction.
While our study focuses on binary classification -- the predominant setting in the fairness literature and metrics \cite{Mehrabi2021}, another interesting direction is to extend SAFES to satisfy fairness constraints in other settings, such as multi-class classification, regression, ranking, and clustering.
A central challenge for these extensions, like in many fairness settings, is a lack of a set of accepted meaningful and interpretable fairness criteria.

The SAFES procedure has the potential for positive societal impacts in a wide range of fields (e.g., healthcare, hiring, criminal justice), where privacy and fairness considerations are necessary in the deployment of responsible AI. 
Organizations might consider integrating SAFES into their data-driven decision making pipeline with confidence that they are gaining useful insights from synthetic data with guaranteed privacy and improved fairness.
On a cautionary note, SAFES should not be treated as a black box.
It is important to understand the privacy and fairness requirement of a given problem before implementing SAFES; a lack of understanding of these aspects can easily result in unsubstantiated claims about privacy and/or fairness, potentially exacerbating the very issues SAFES is designed to mitigate.

\section*{\normalsize{Code}}
The code for the experiments in this paper can be found at \url{https://github.com/sgiddens/SAFES}.

\section*{\normalsize{Acknowledgments}}

This work was partially supported by the University of Notre Dame Schmitt Fellowship and Lucy Graduate Scholarship to S.G. 
A portion of the work was carried out while S.G. was an intern at SandboxAQ, and S.G. acknowledges helpful discussion from supervisors Nicolas Gama and Sandra Guasch.

\bibliographystyle{apalike}
\bibliography{references}

\section*{Appendix}
\appendix

\setcounter{figure}{0}
\setcounter{table}{0}

\section{Additional figures in  the experiments}

\begin{figure}[!htb]
    \centering
    \includegraphics[width=0.325\linewidth]{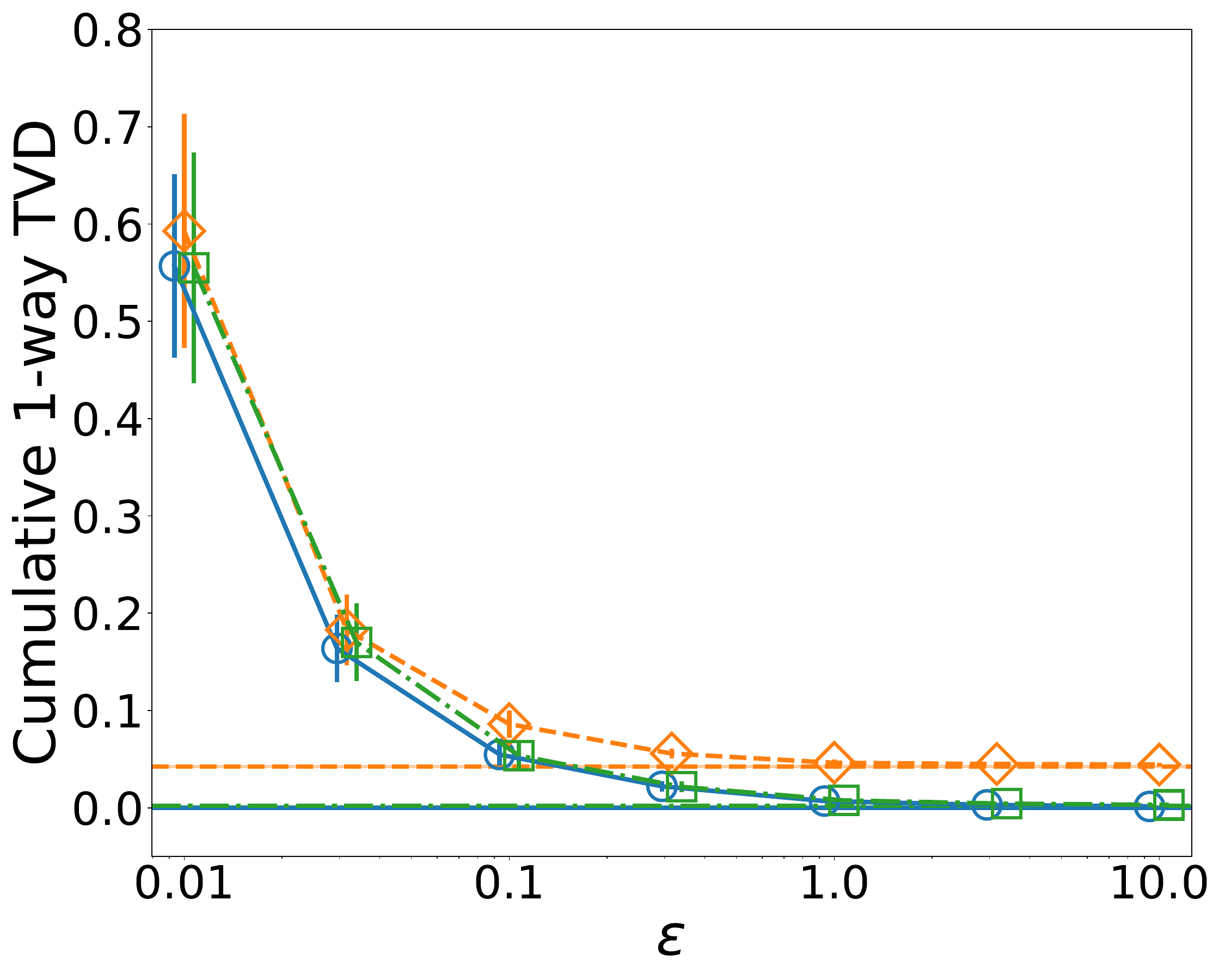}
    \includegraphics[width=0.325\linewidth]{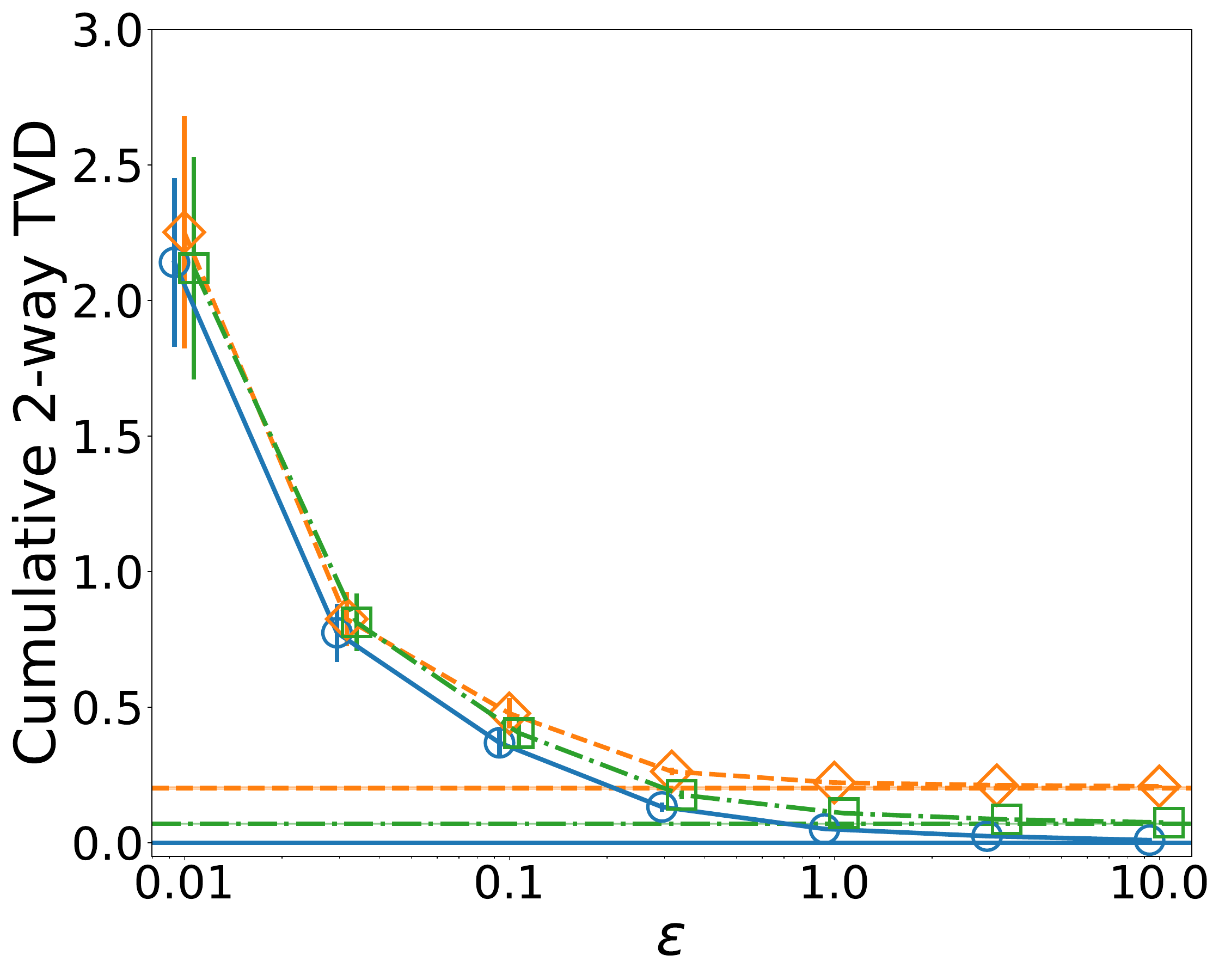}
    \includegraphics[width=0.325\linewidth]{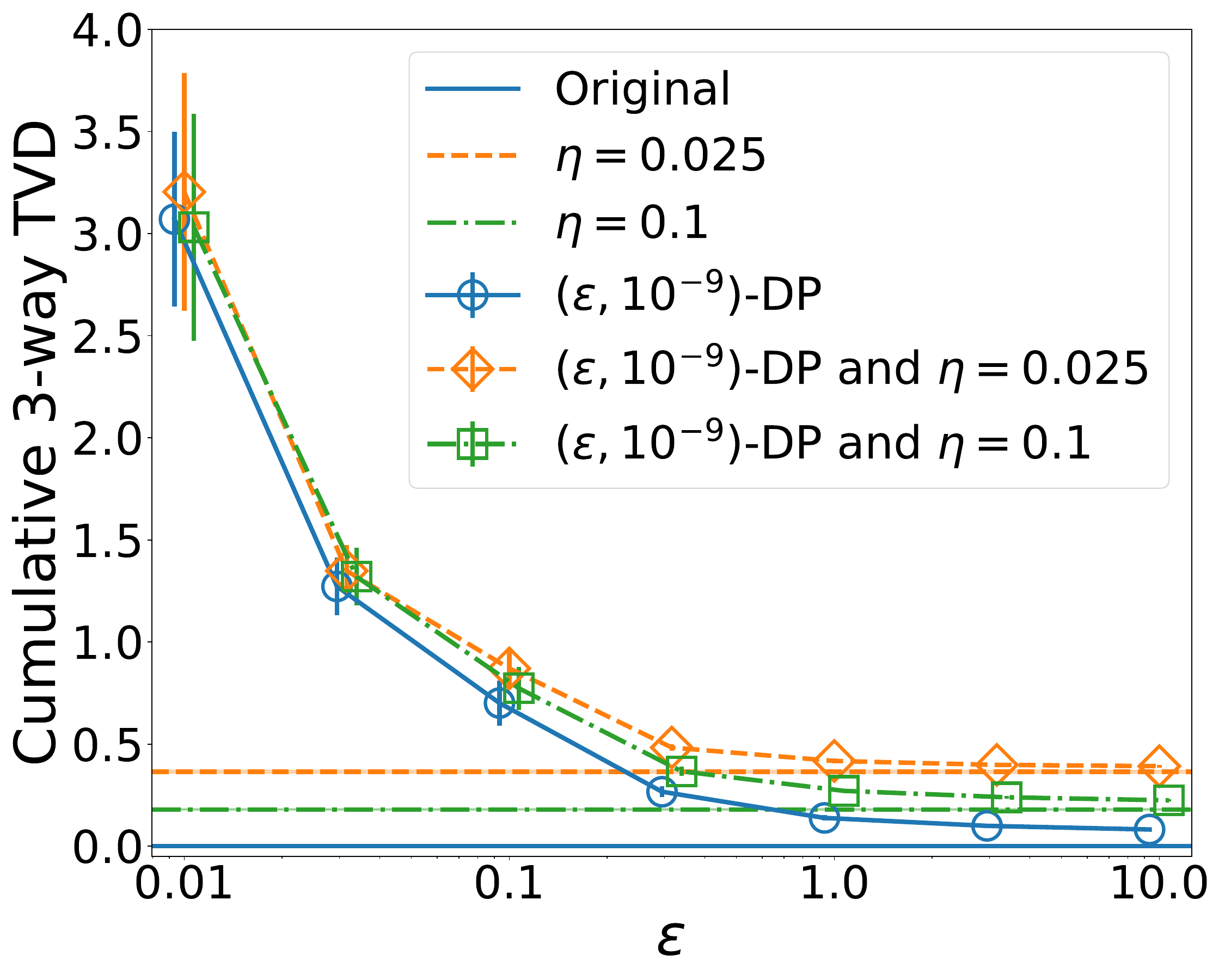}
    \caption{Mean $\pm$ 1 SD (error bars and shaded regions) summed TVD in each marginal set for 1-way, 2-way, and 3-way marginals between the synthetic data vs the original data for the Adult experiment.}
    \label{fig:Adult_TVD}
\end{figure}

\begin{figure}[!htb]
    \centering
    \includegraphics[width=0.35\linewidth]{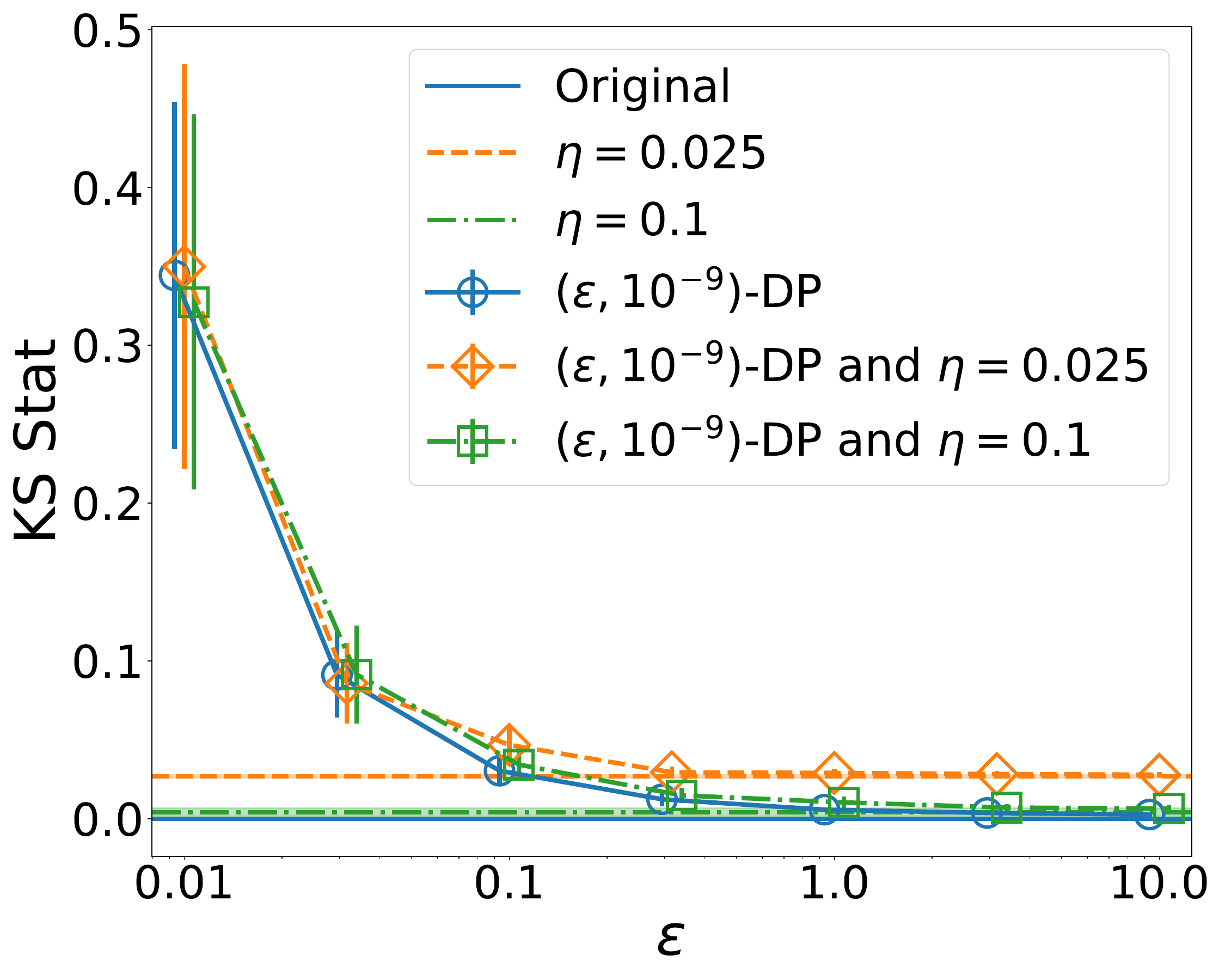}
    \includegraphics[width=0.35\linewidth]{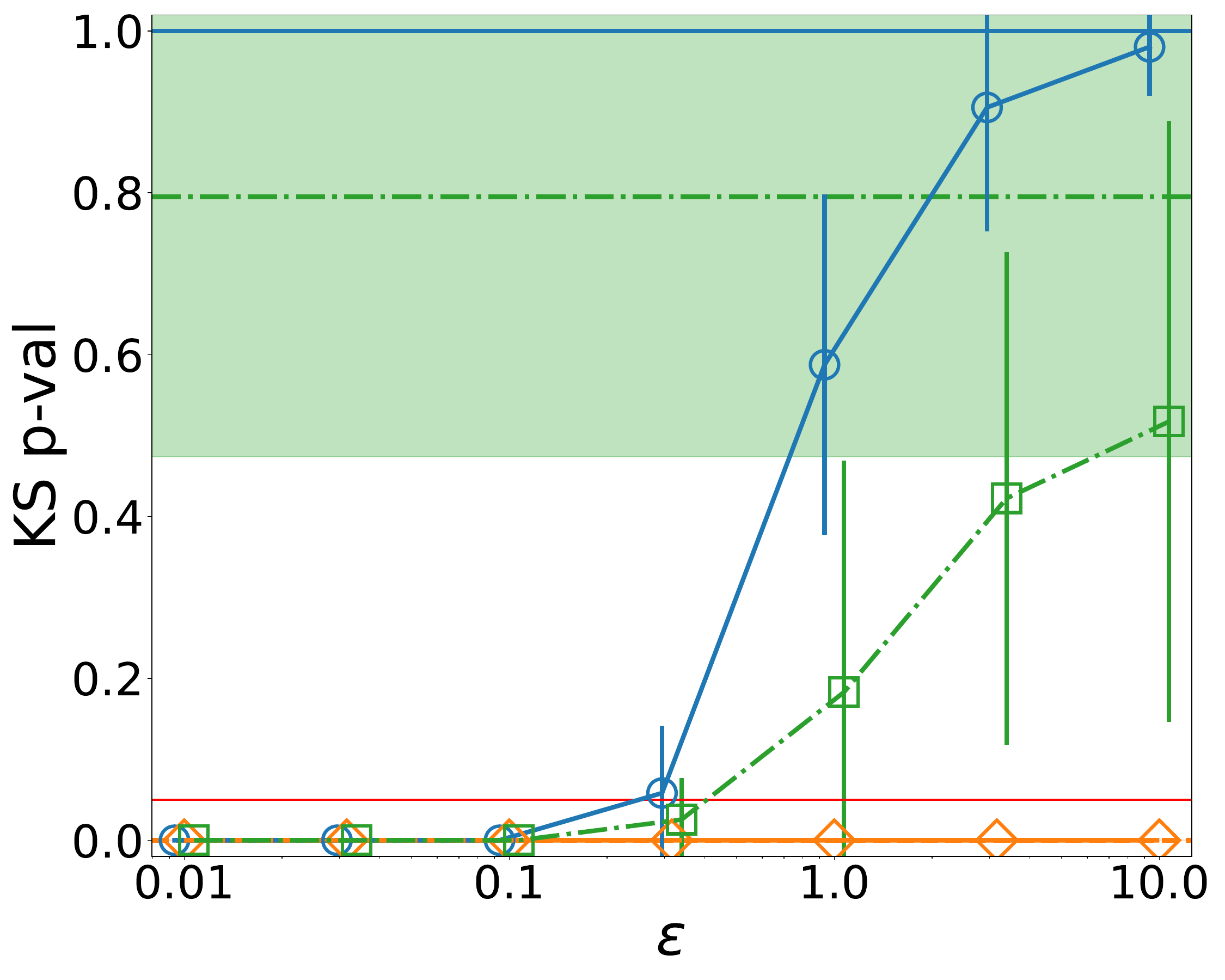}
    \caption{Mean $\pm$ 1 SD (error bars and shaded regions) test statistic and corresponding p-value for the KS test comparing original and synthetic datasets for the Adult experiment.
    Statistical significance threshold of $\alpha=0.05$ is marked in red in the plot on the right.}
    \label{fig:Adult_KS}
\end{figure}

\begin{figure}[!htb]
    \centering
    \includegraphics[width=0.325\linewidth]{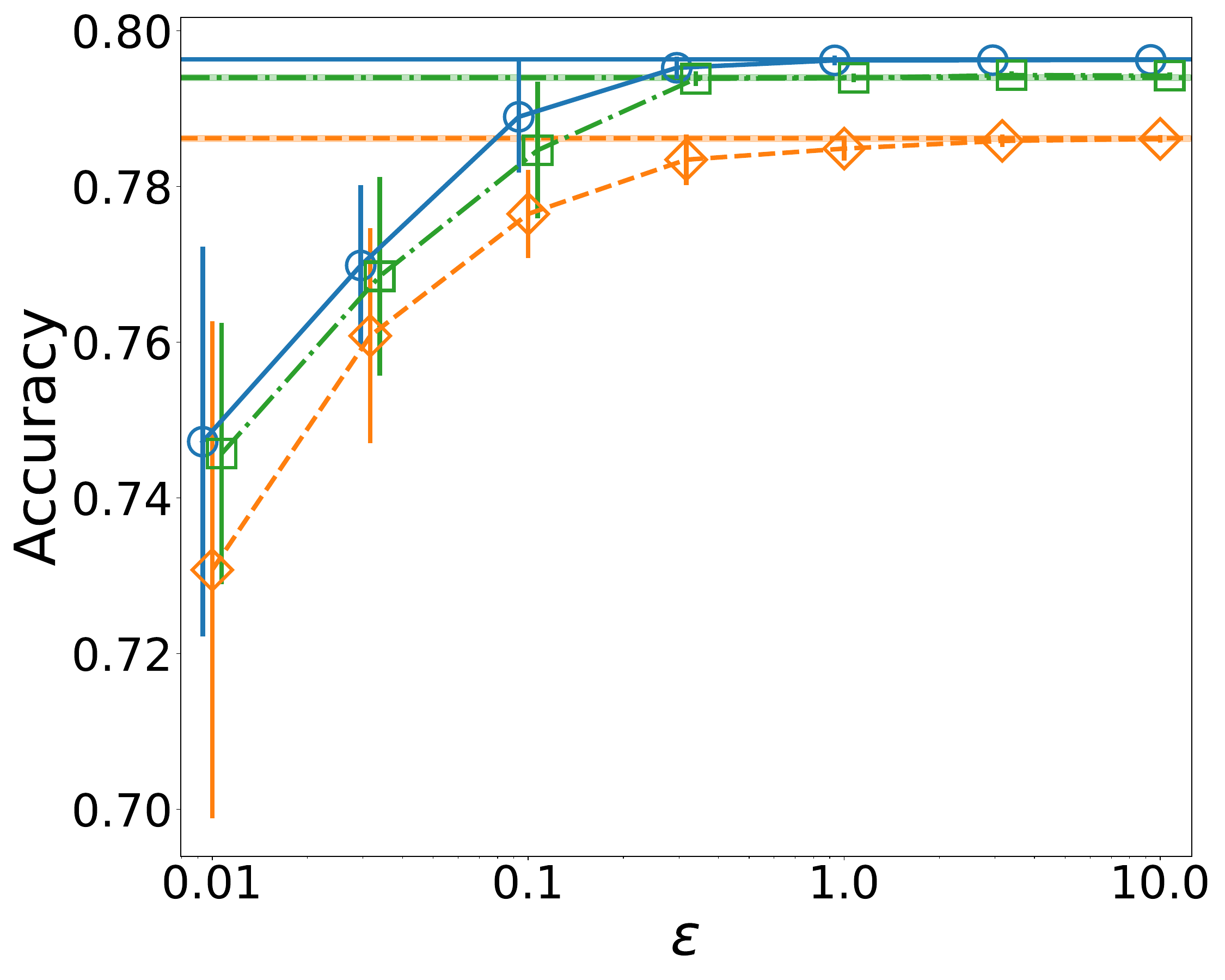}
    \includegraphics[width=0.325\linewidth]{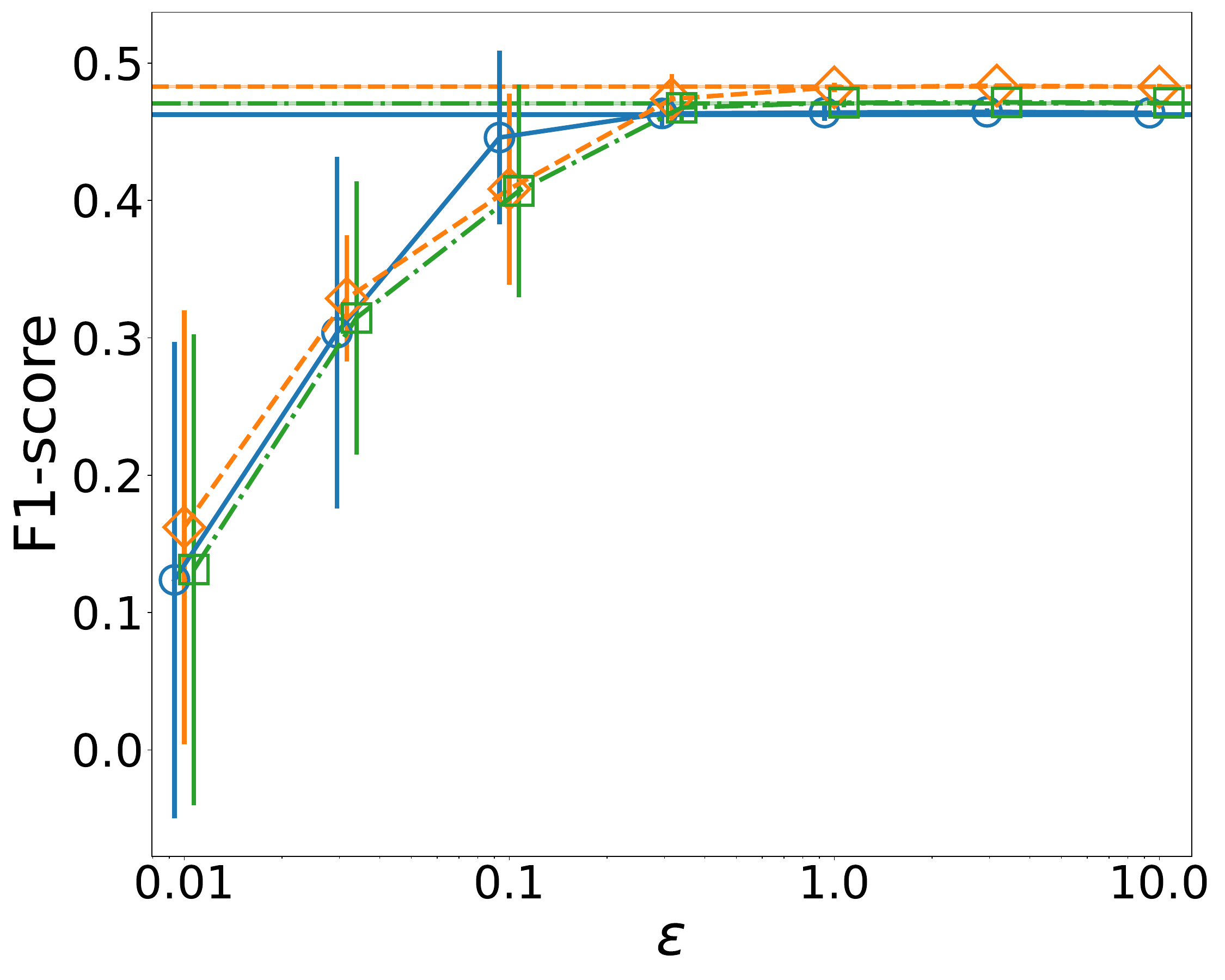}
    \includegraphics[width=0.325\linewidth]{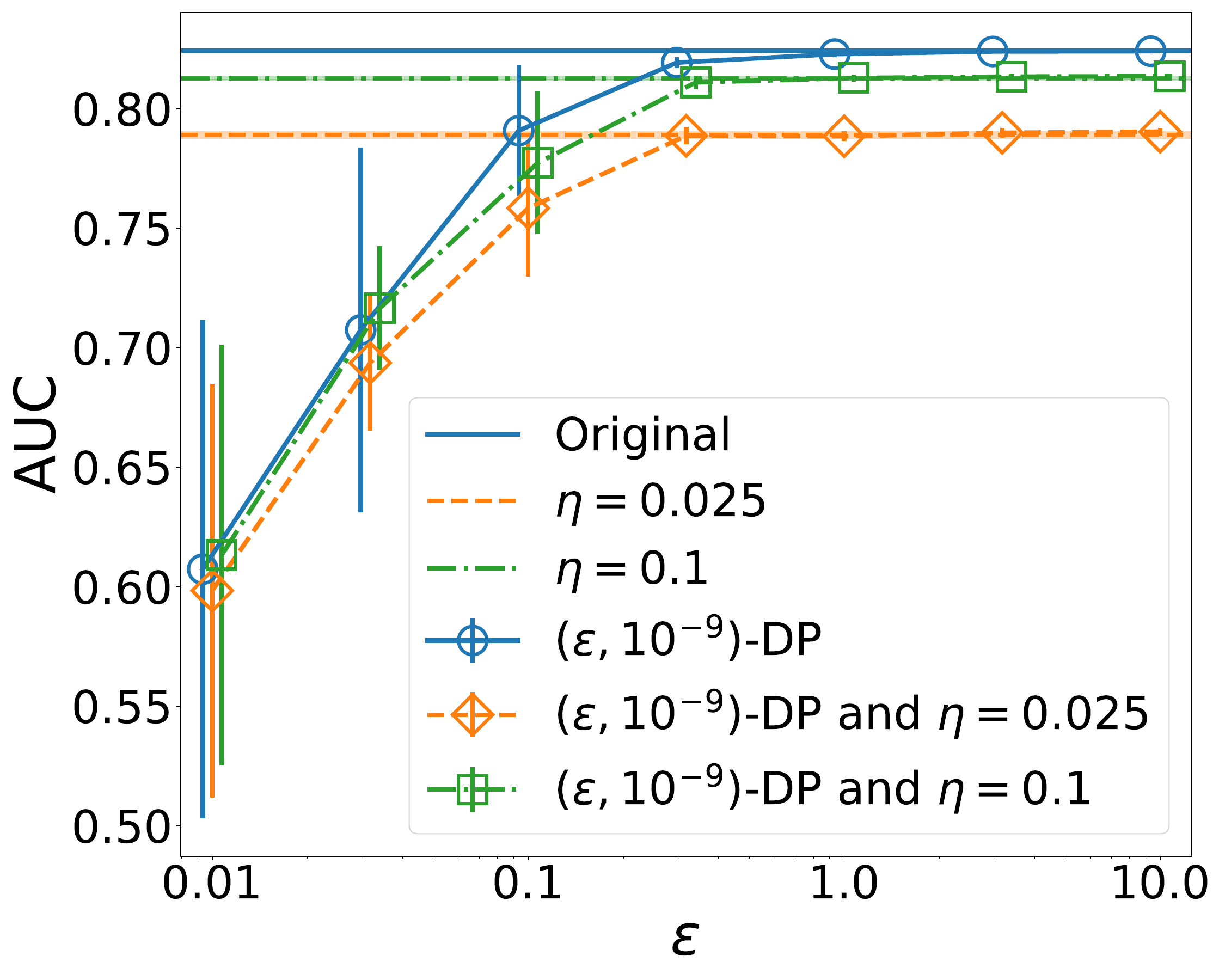}
    \caption{Mean $\pm$ 1 SD (error bars and shaded regions) prediction performance of the logistic regression model trained on SAFES synthetic data for the Adult experiment.}
    \label{fig:Adult_pred}
\end{figure}

\begin{figure}[!htb]
    \includegraphics[width=0.325\linewidth]{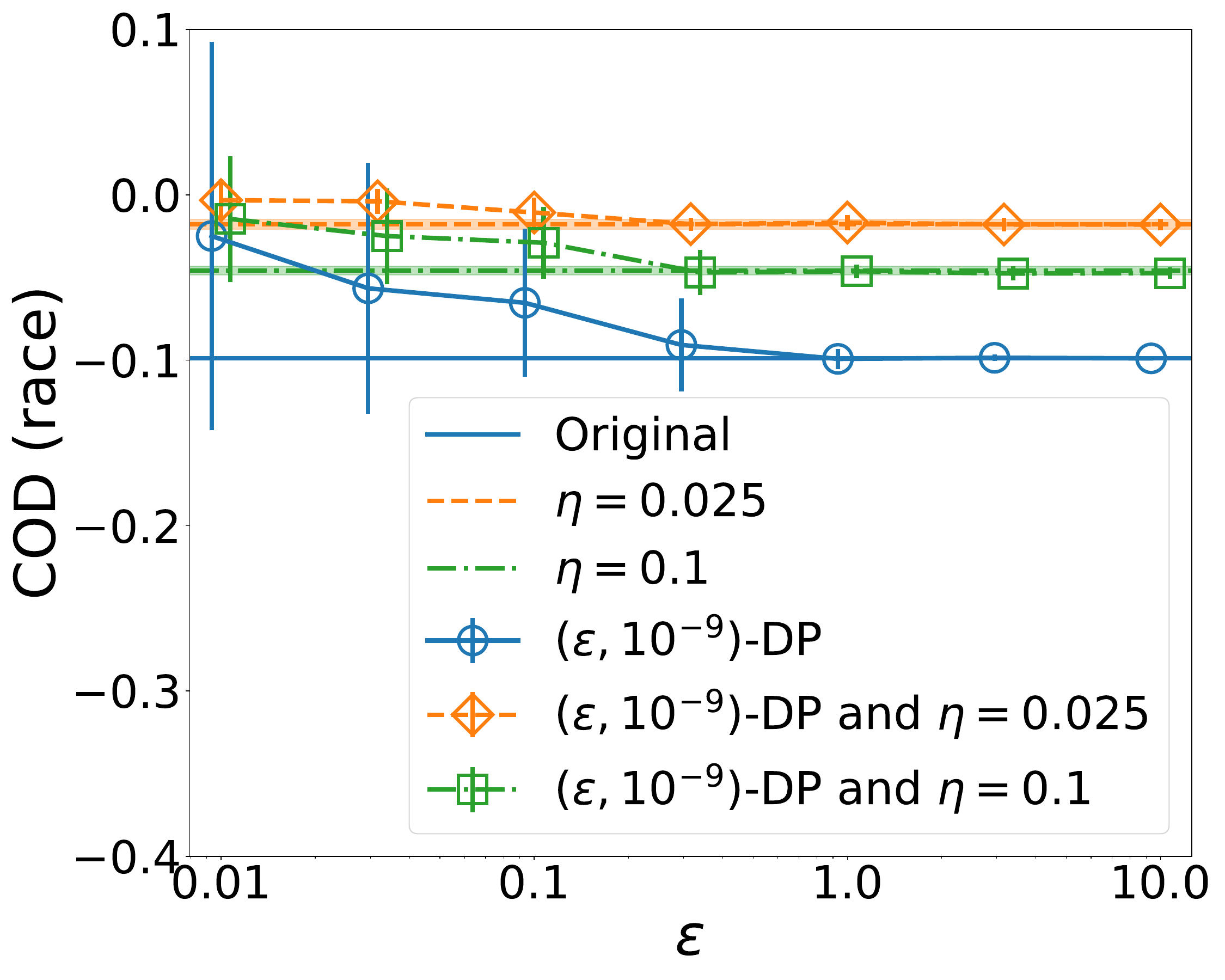}
    \includegraphics[width=0.325\linewidth]{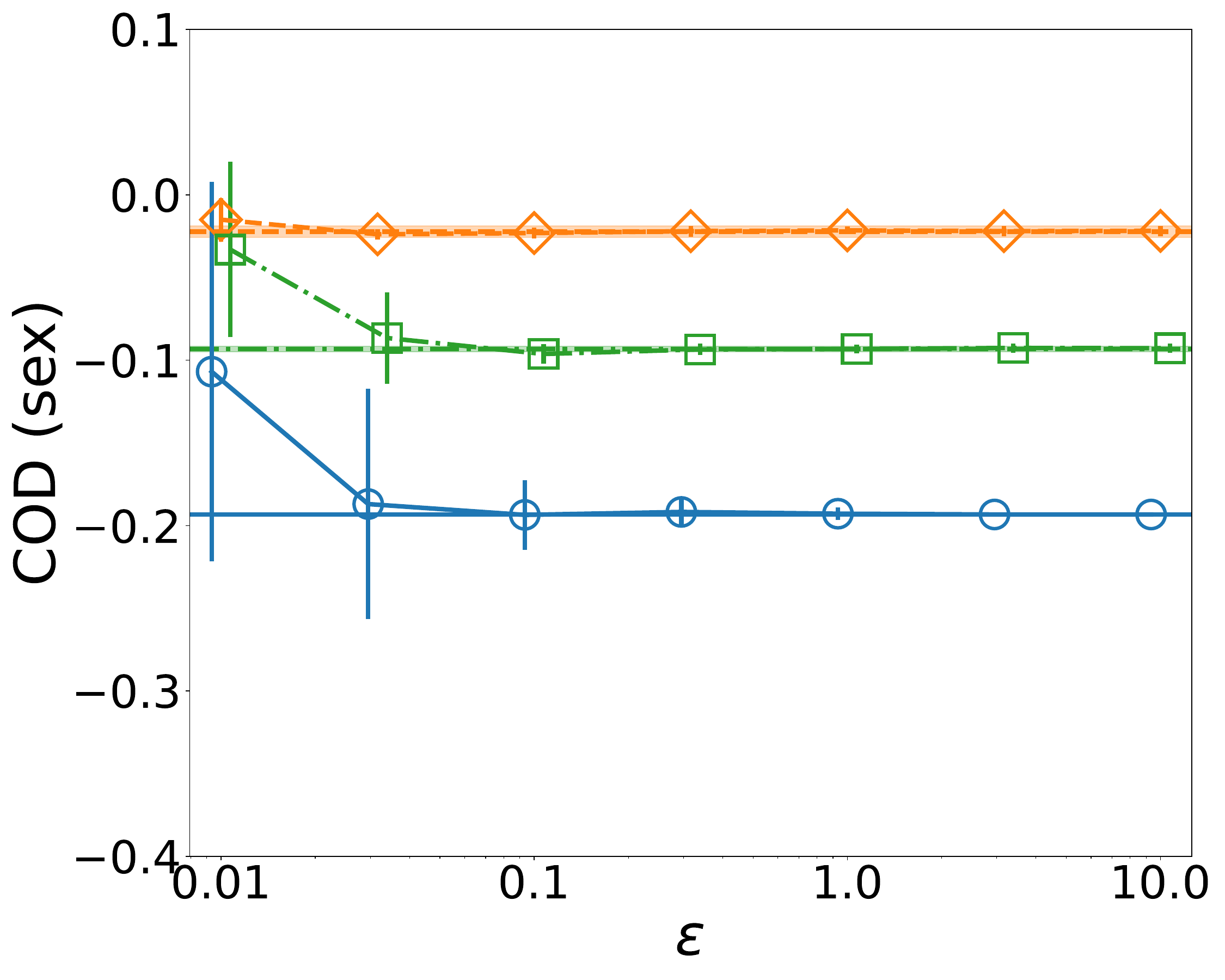}
    \includegraphics[width=0.325\linewidth]{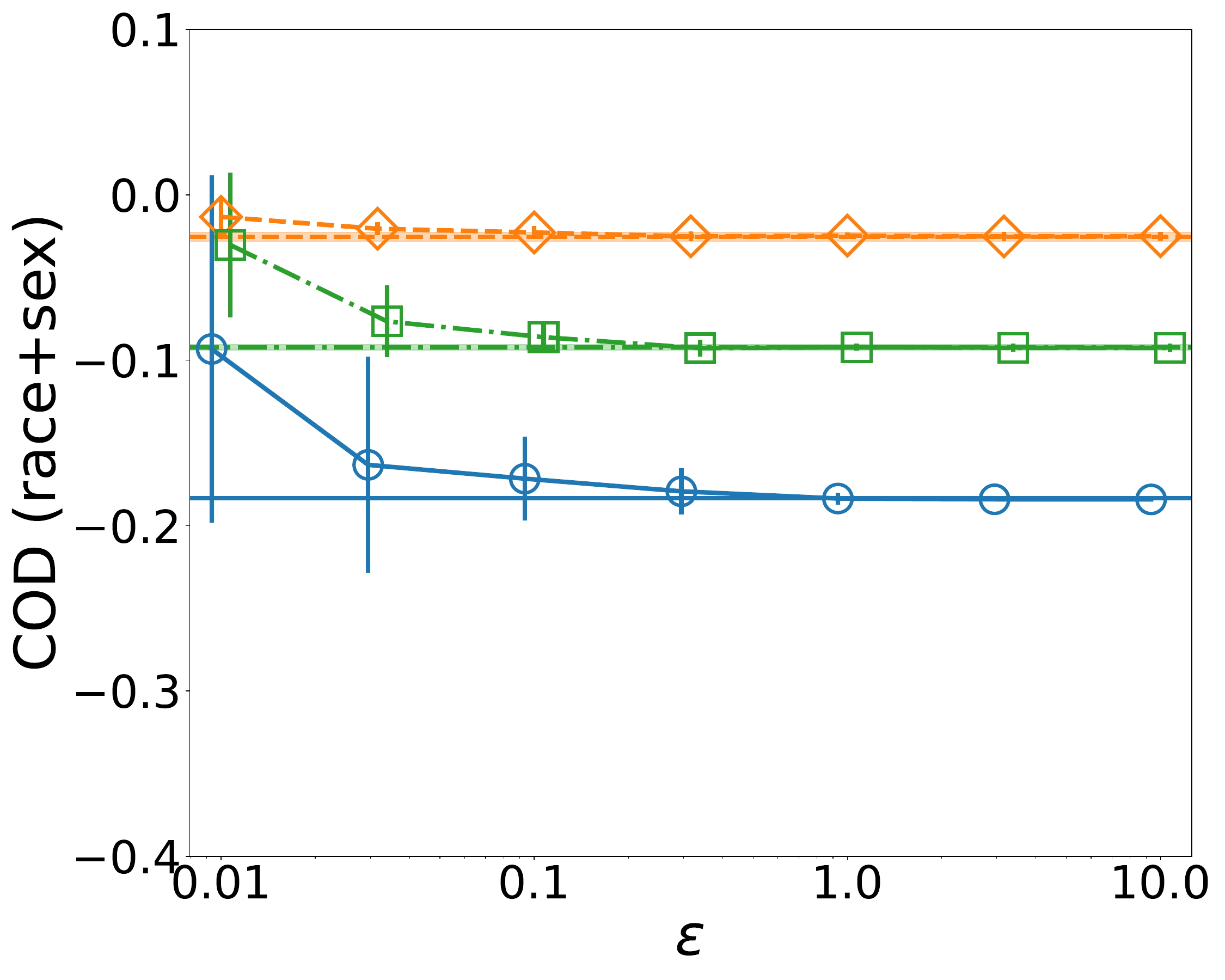}
    \caption{Mean $\pm$ 1 SD (error bars and shaded regions) COD, measured with race, sex, and race+sex as the protected attribute, for the Adult experiment.}
    \label{fig:Adult_COD}
\end{figure}

\begin{figure}[!htb]
    \includegraphics[width=0.325\linewidth]{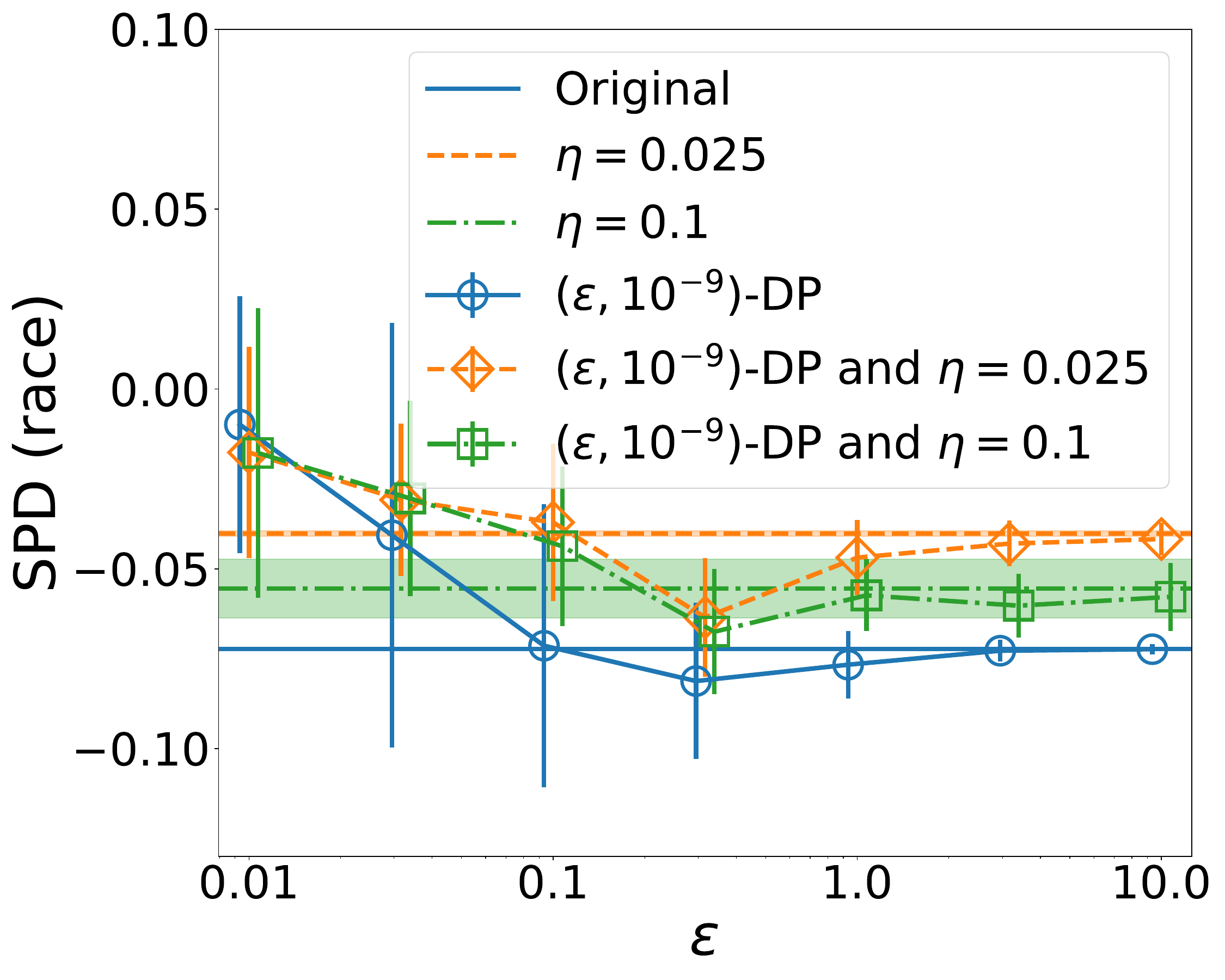}
    \includegraphics[width=0.325\linewidth]{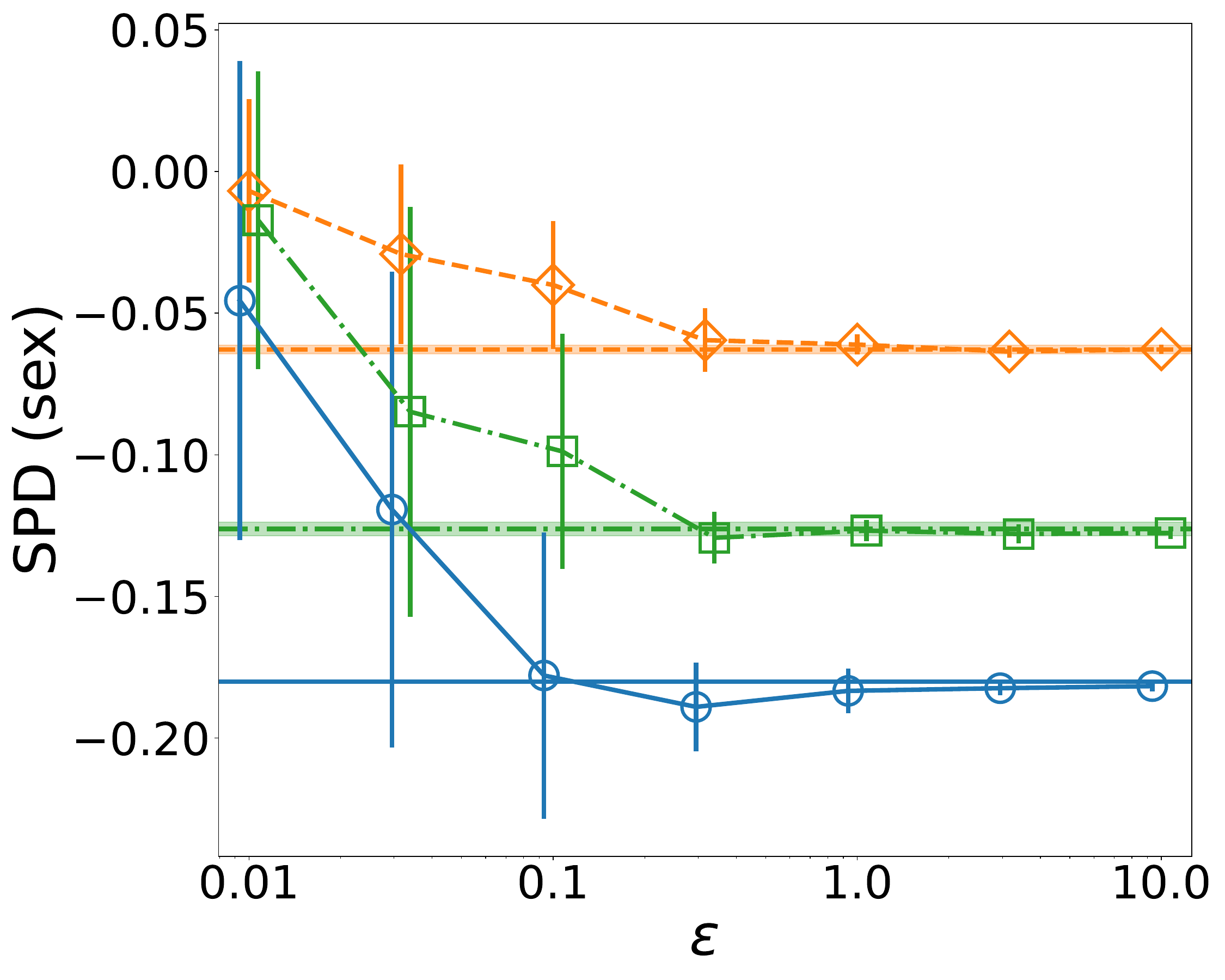}
    \includegraphics[width=0.325\linewidth]{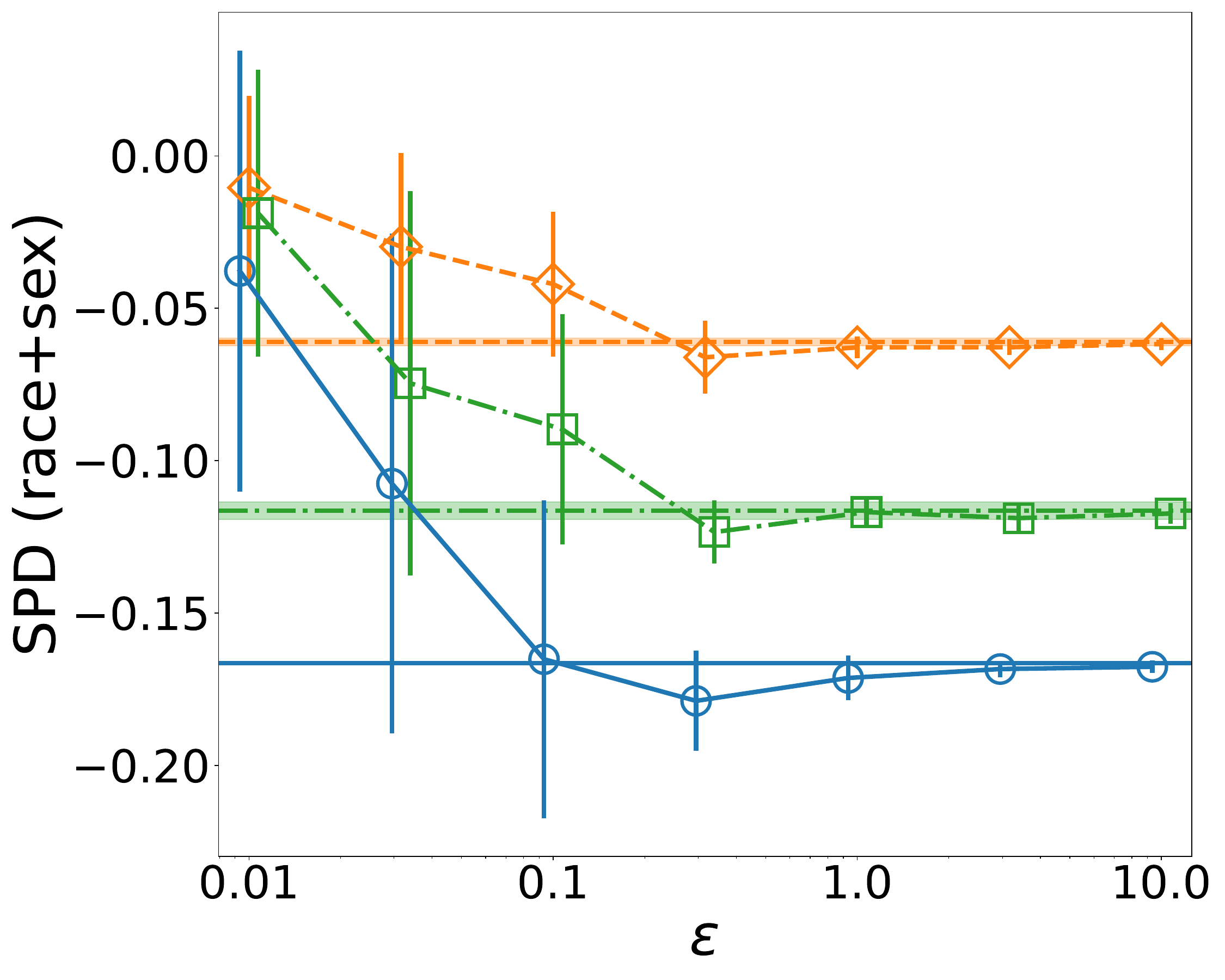}
    \caption{Mean $\pm$ 1 SD (error bars and shaded regions) SPD, with race, sex, and race+sex as the protected attribute, for the Adult experiment.}
    \label{fig:Adult_SPD}
\end{figure}

\begin{figure}[!htb]
    \includegraphics[width=0.325\linewidth]{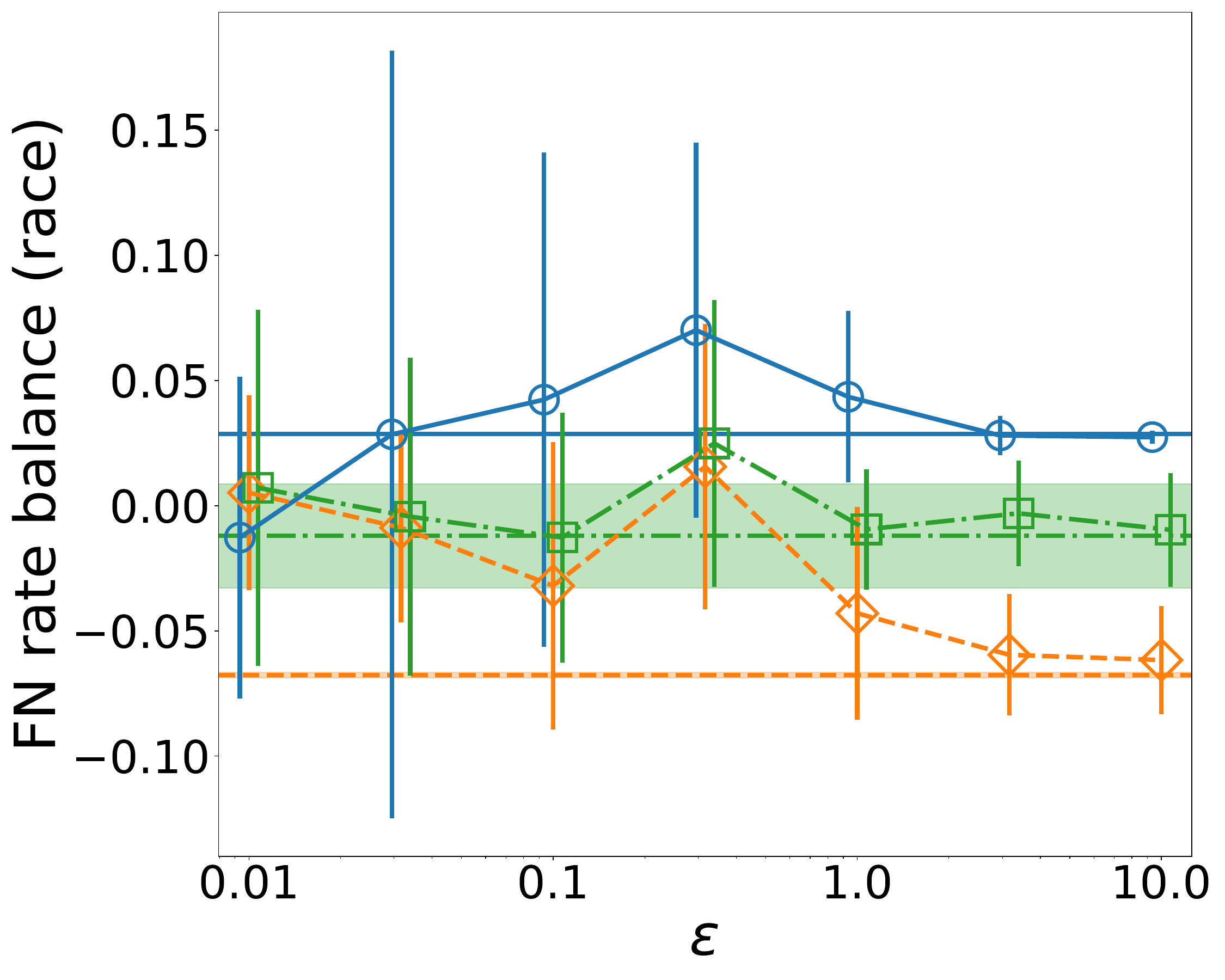}
    \includegraphics[width=0.325\linewidth]{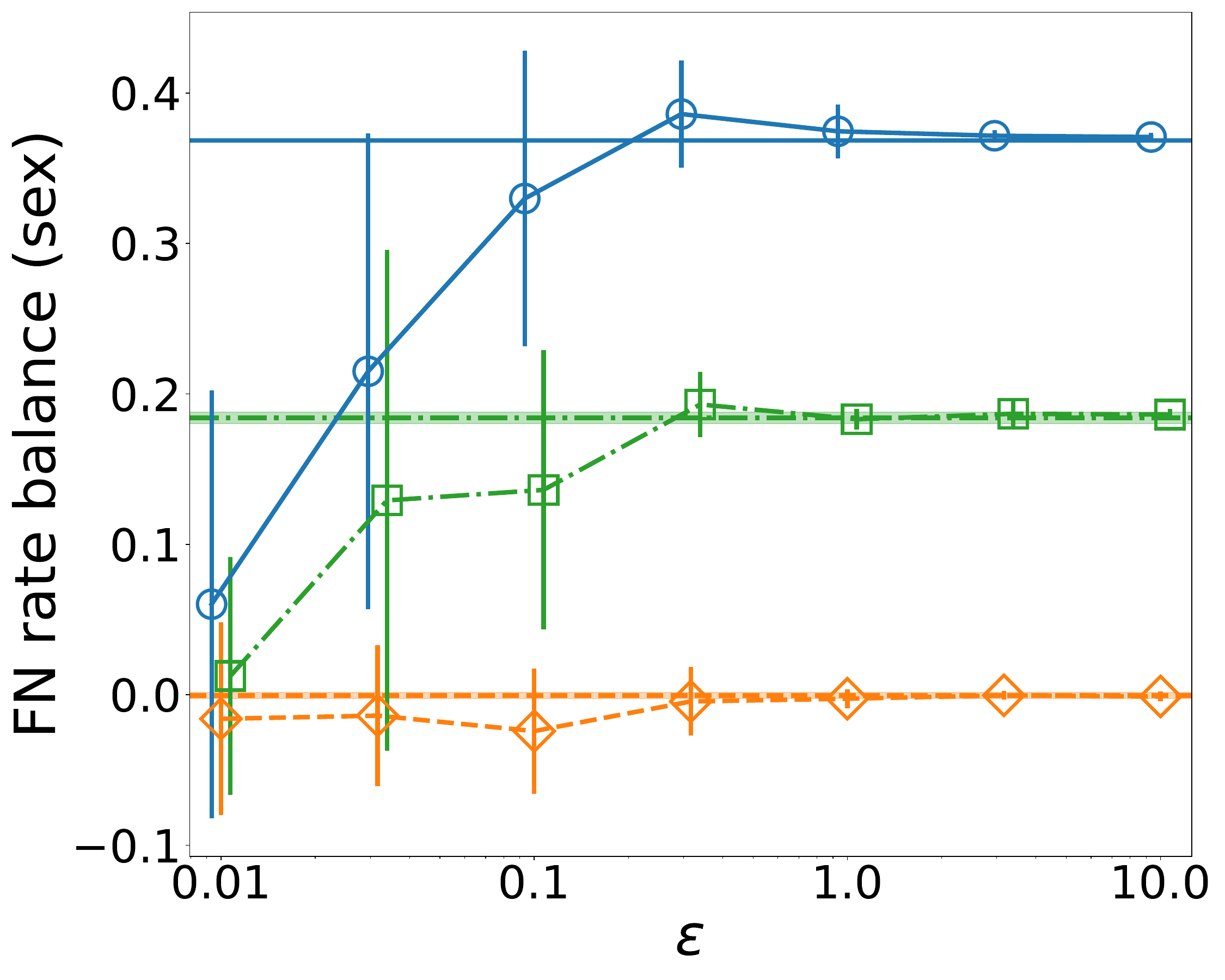}
    \includegraphics[width=0.325\linewidth]{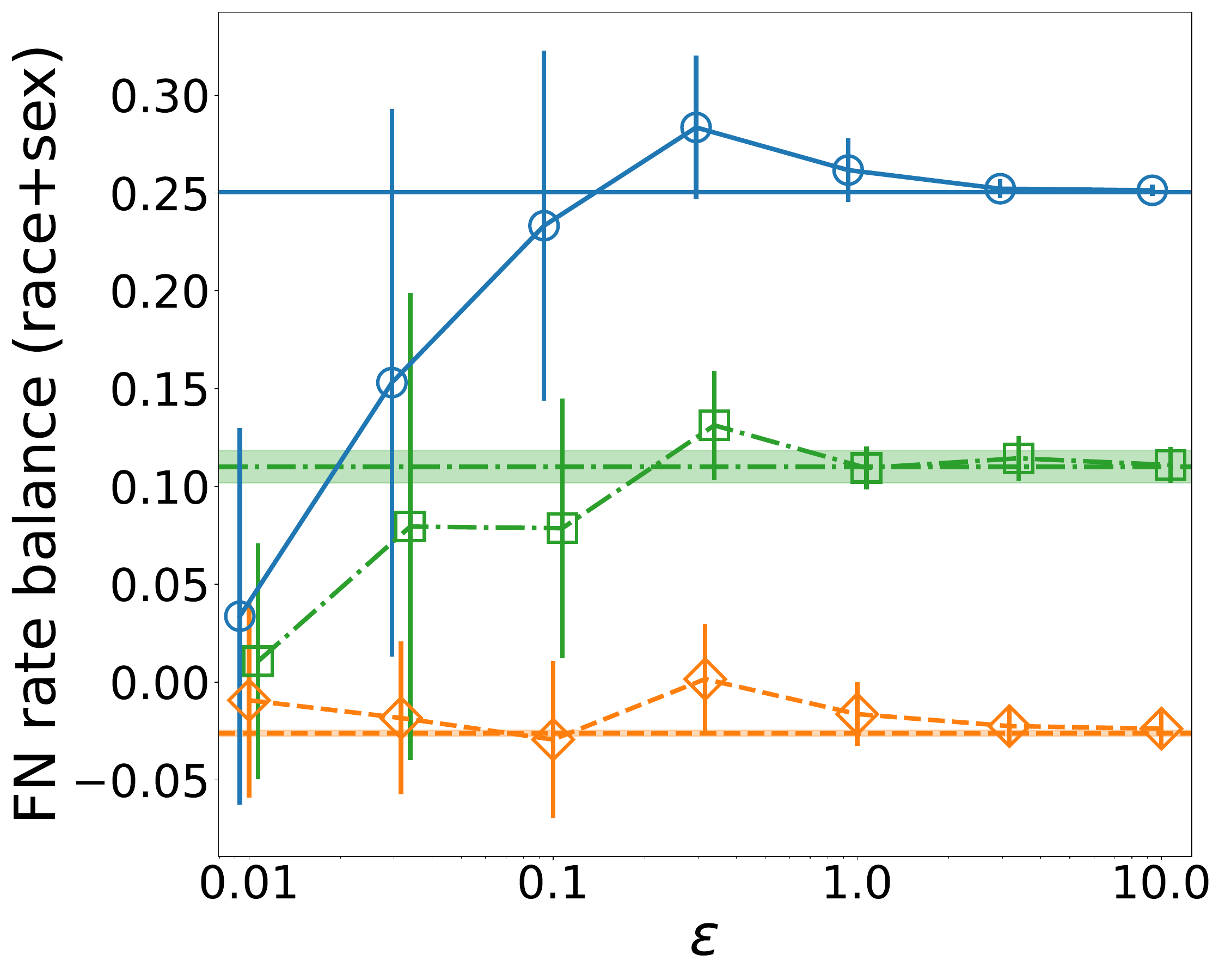}
    \caption{Mean $\pm$ 1 SD (error bars and shaded regions) FN rate balance, with race, sex, and race+sex as the protected attribute, for the Adult experiment.}
    \label{fig:Adult_FNR_balance}
\end{figure}

\begin{figure}[!htb]
    \includegraphics[width=0.325\linewidth]{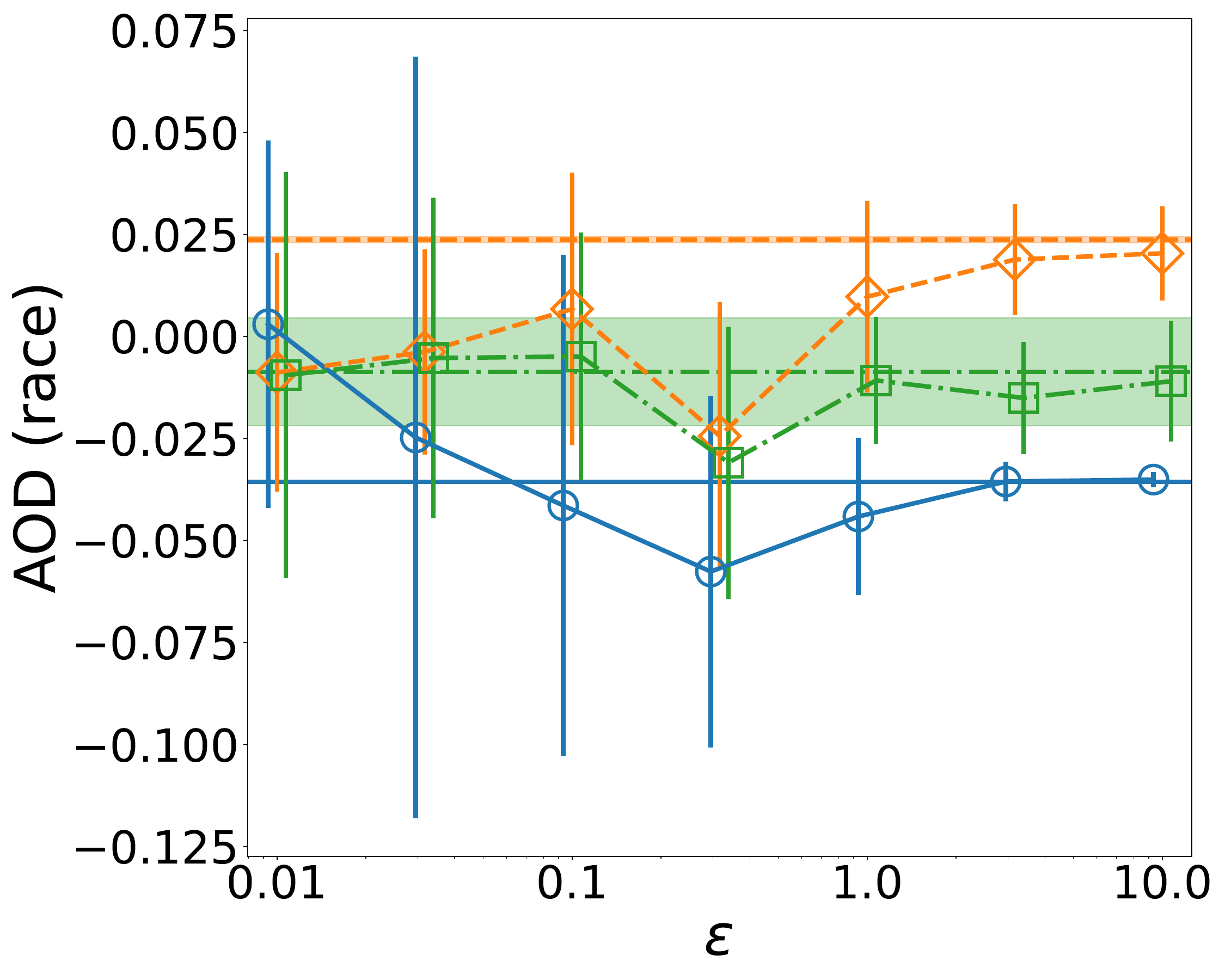}
    \includegraphics[width=0.325\linewidth]{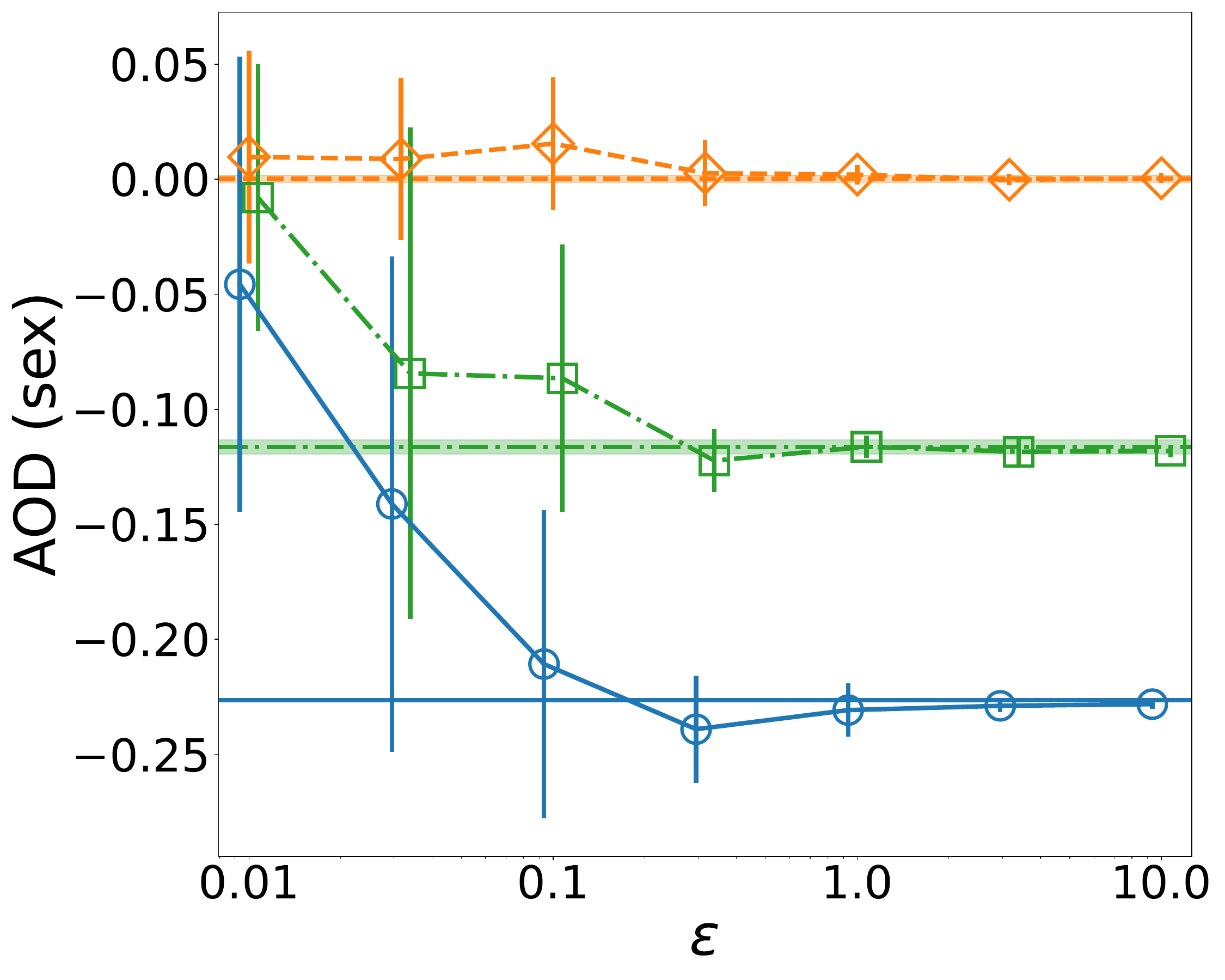}
    \includegraphics[width=0.325\linewidth]{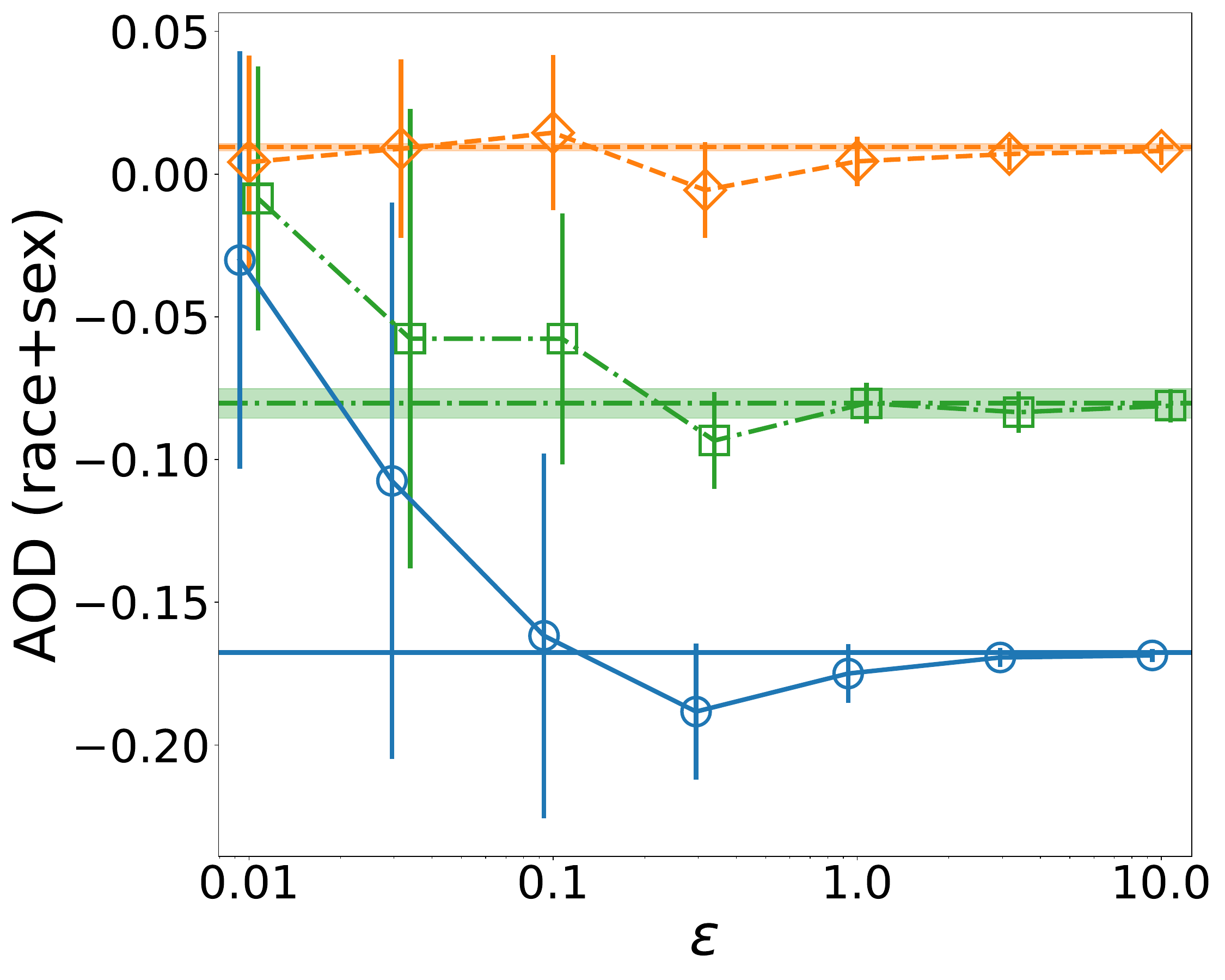}
    \caption{Mean $\pm$ 1 SD (error bars and shaded regions) AOD, with race, sex, and race+sex as the protected attribute, for the Adult experiment.}
    \label{fig:Adult_AOD}
\end{figure}

\begin{figure}[!htb]
    \includegraphics[width=0.325\linewidth]{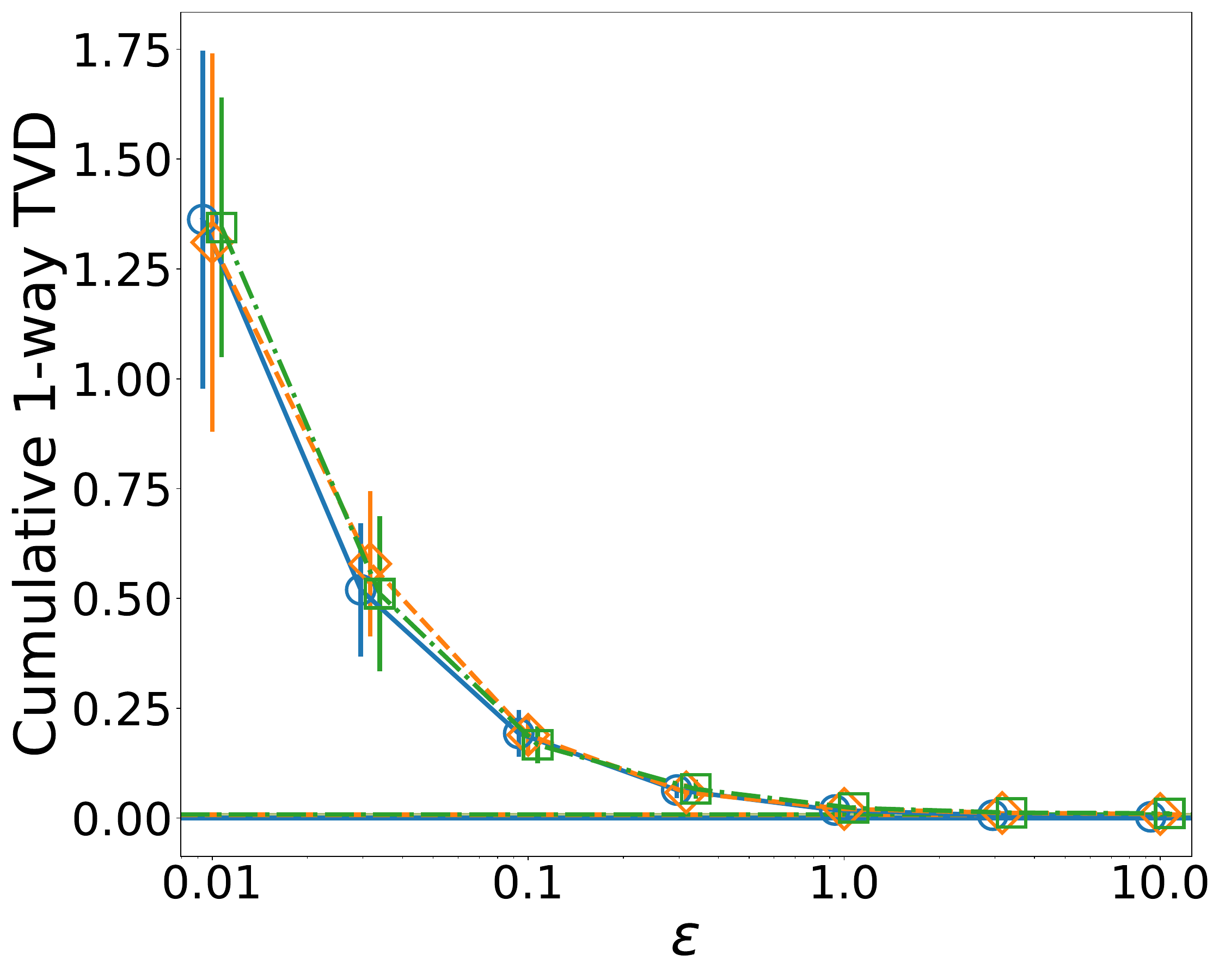}
    \includegraphics[width=0.325\linewidth]{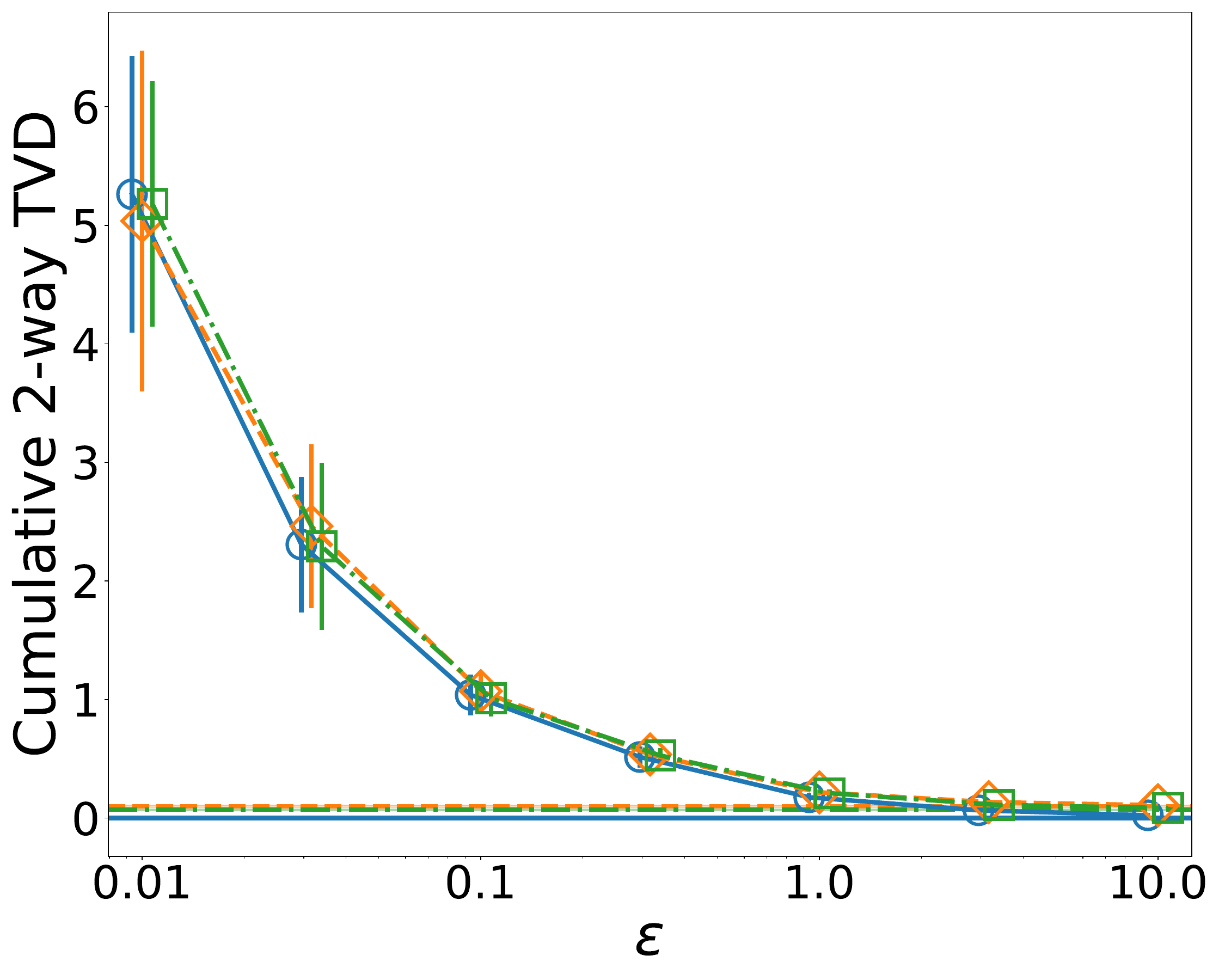}
    \includegraphics[width=0.325\linewidth]{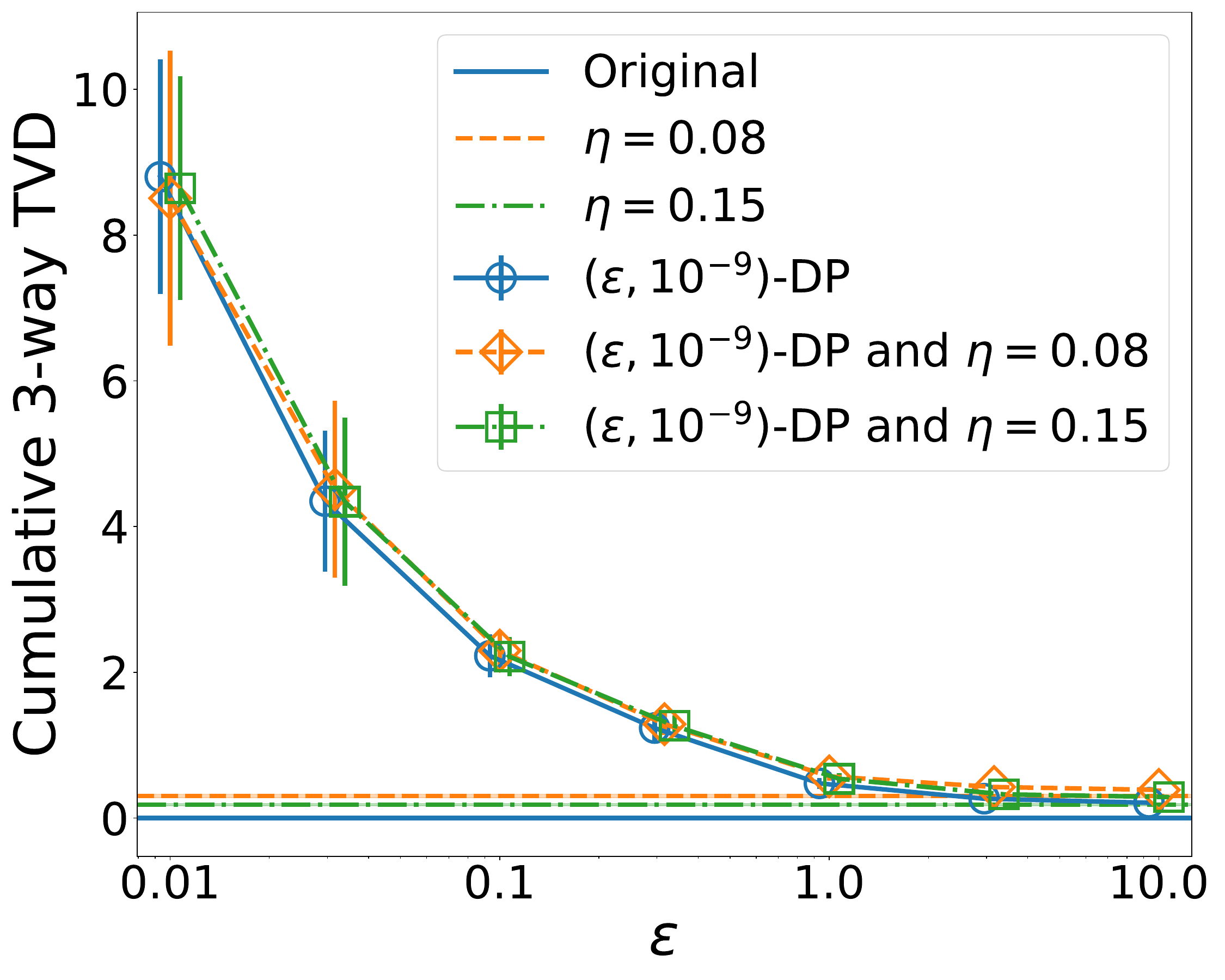}
    \caption{Mean $\pm$ 1 SD (error bars and shaded regions) summed TVD in each marginal set for 1-way, 2-way, and 3-way marginals between the synthetic data vs the original data for the COMPAS experiment.}
    \label{fig:compas_TVD}
\end{figure}

\begin{figure}[!htb]
    \centering
    \includegraphics[width=0.35\linewidth]{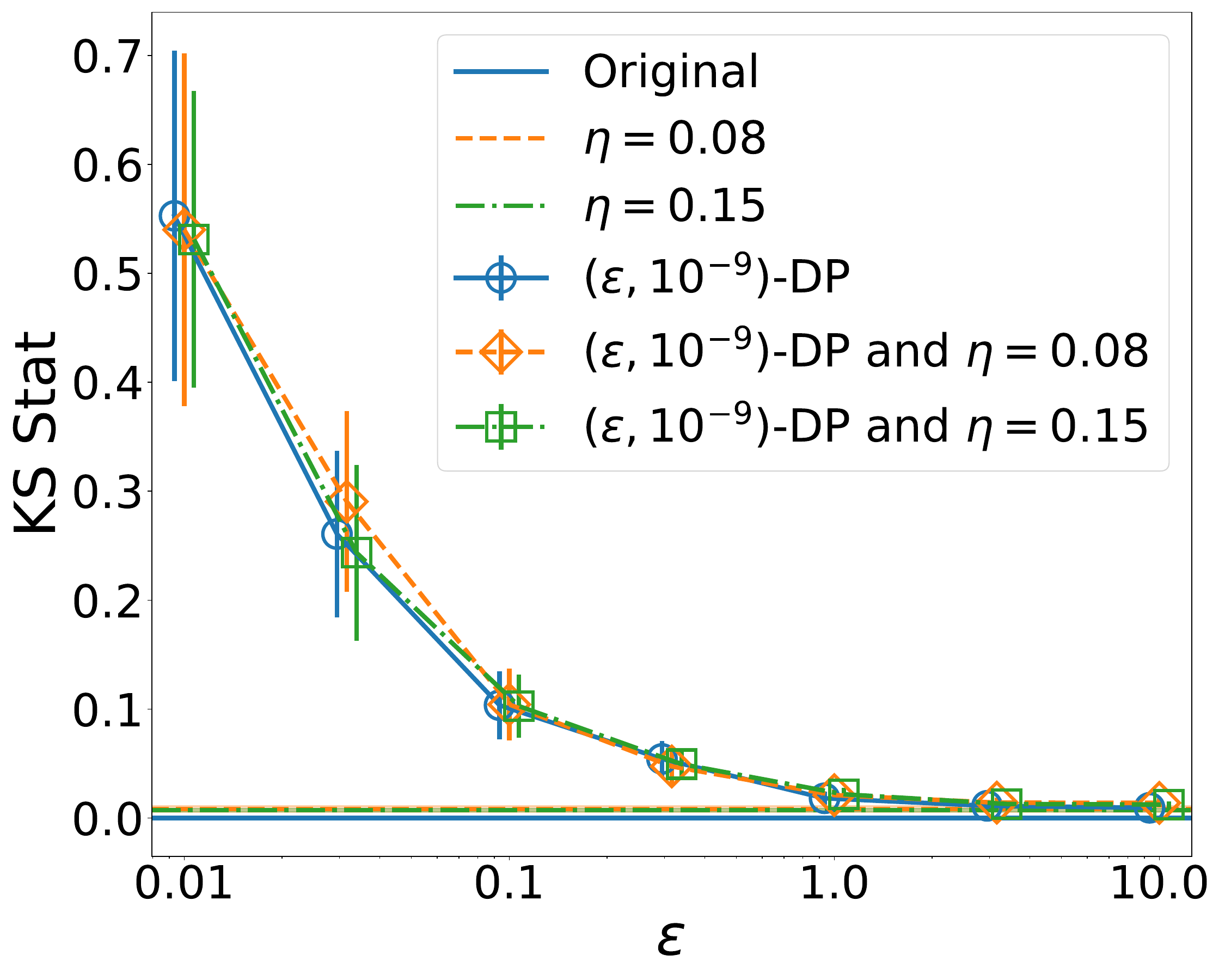}
    \includegraphics[width=0.35\linewidth]{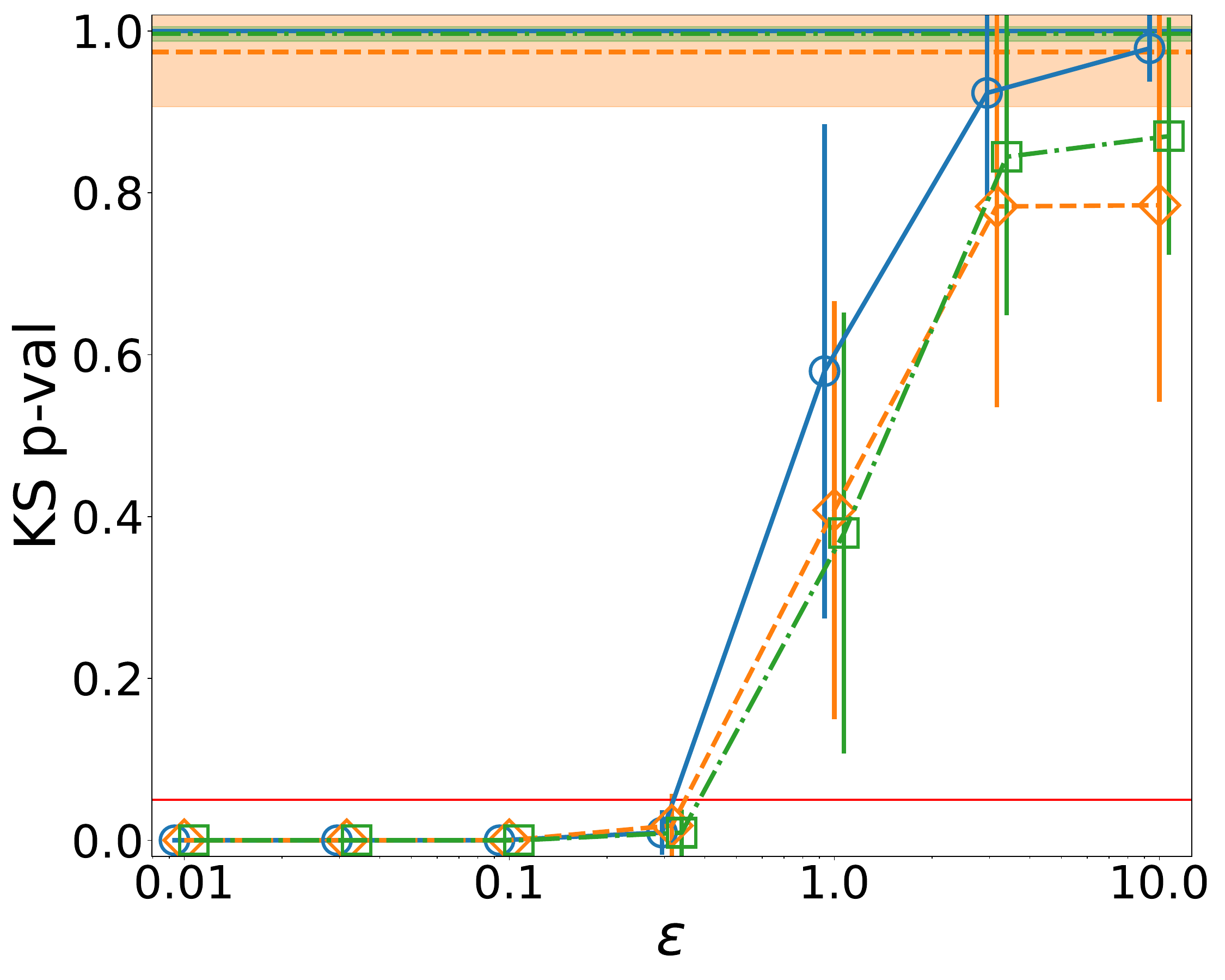}
    \caption{Mean $\pm$ 1 SD (error bars and shaded regions) test statistic and corresponding p-value for the KS test comparing original and synthetic datasets for the COMPAS experiment.
    Statistical significance threshold of $\alpha=0.05$ is marked in red in the plot on the right.}
    \label{fig:compas_KS}
\end{figure}

\begin{figure}[!htb]
    \centering
    \includegraphics[width=0.325\linewidth]{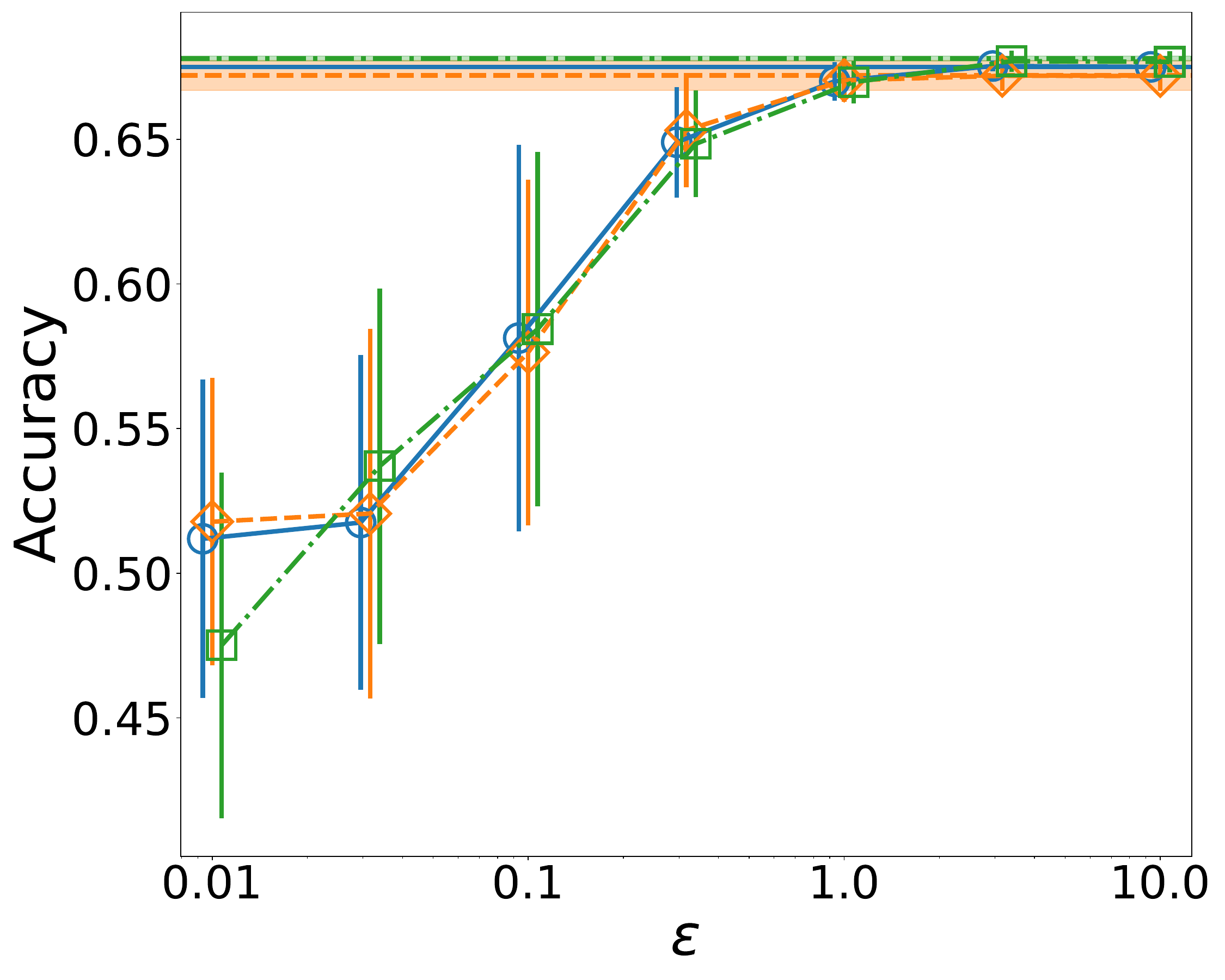}
    \includegraphics[width=0.325\linewidth]{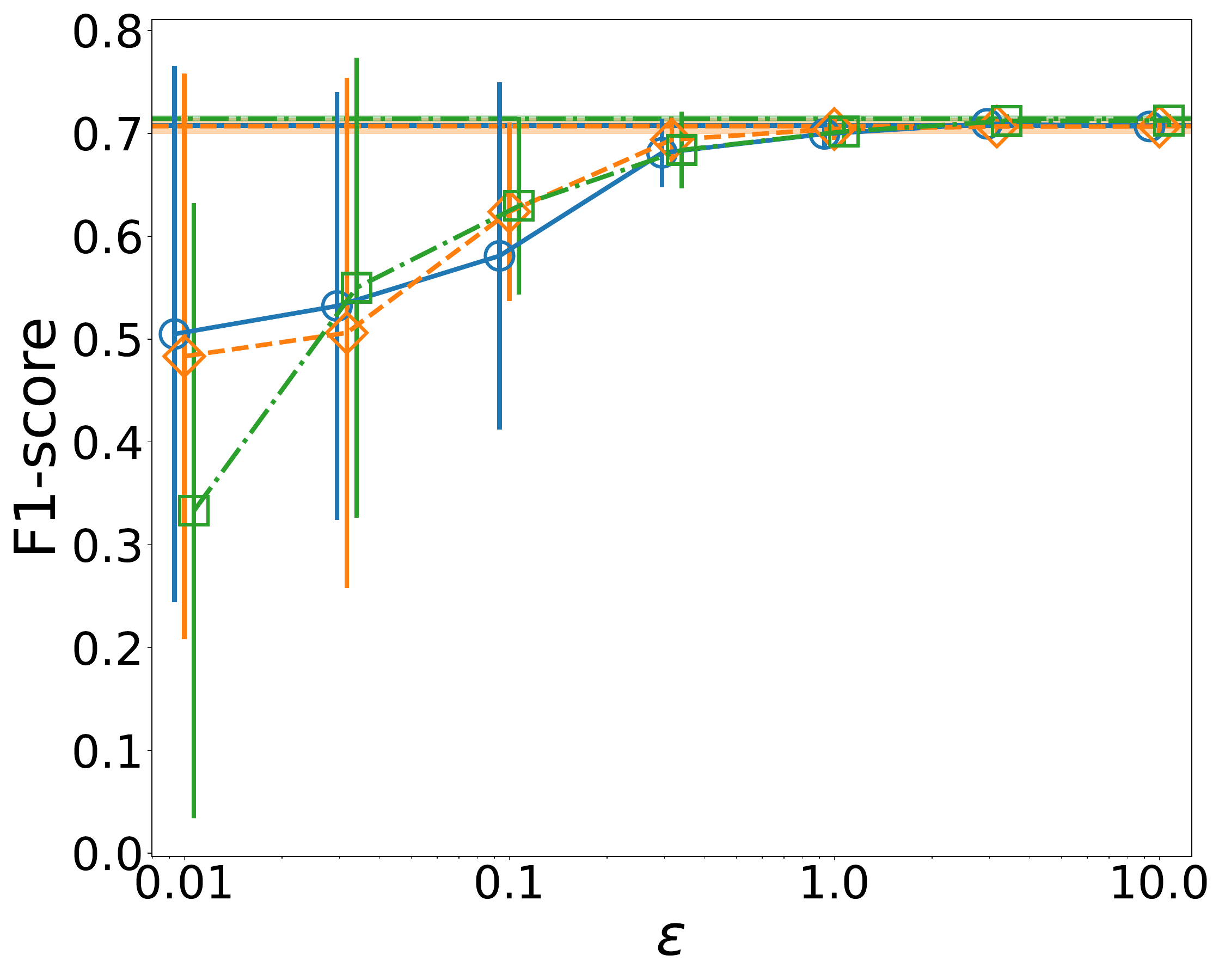}
    \includegraphics[width=0.325\linewidth]{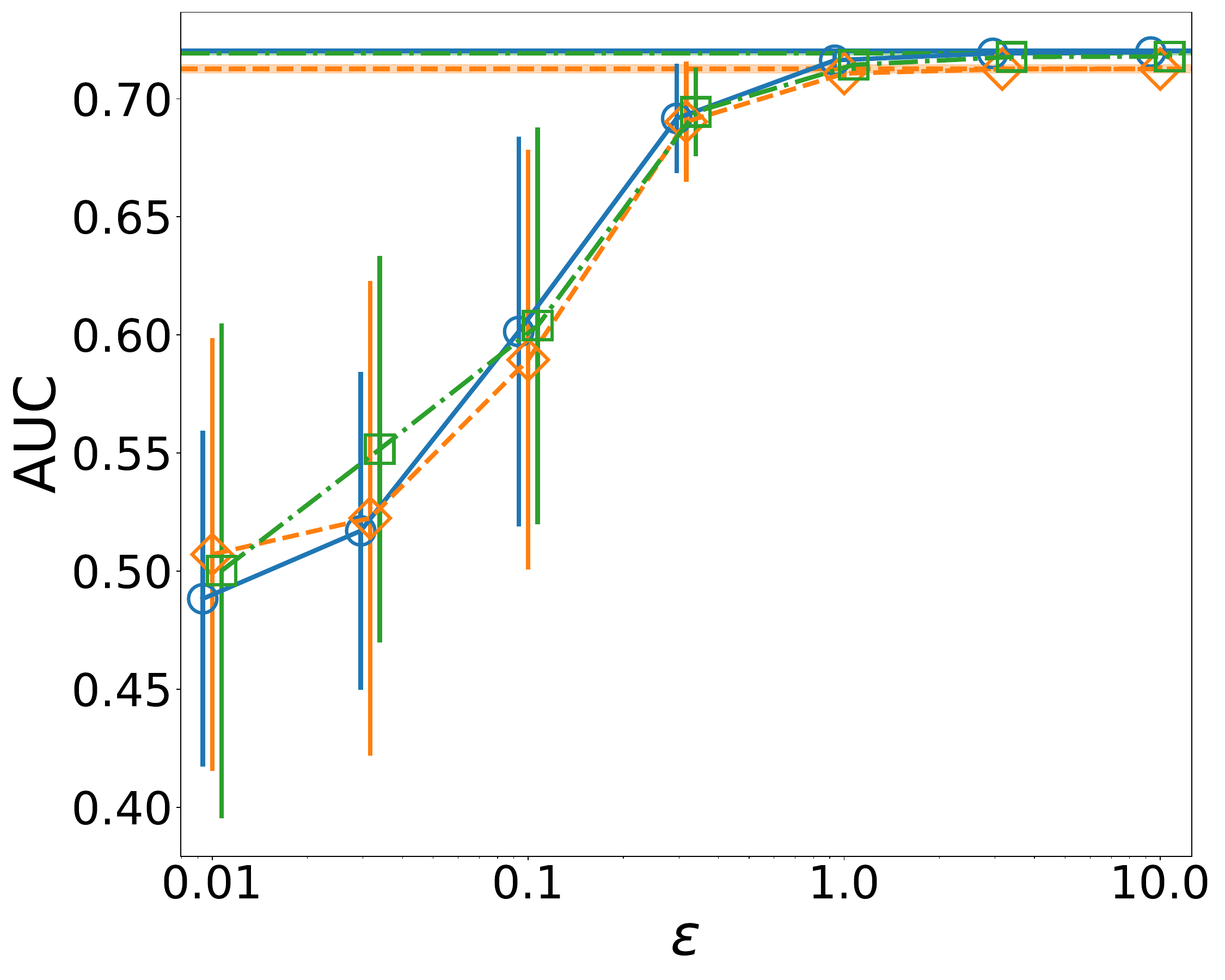}
    \caption{Mean $\pm$ 1 SD (error bars and shaded regions) prediction performance of the logistic regression model trained on SAFES synthetic data for the COMPAS experiment.}
    \label{fig:compas_pred}
\end{figure}

\begin{figure}[!htb]
    \centering
    \includegraphics[width=0.325\linewidth]{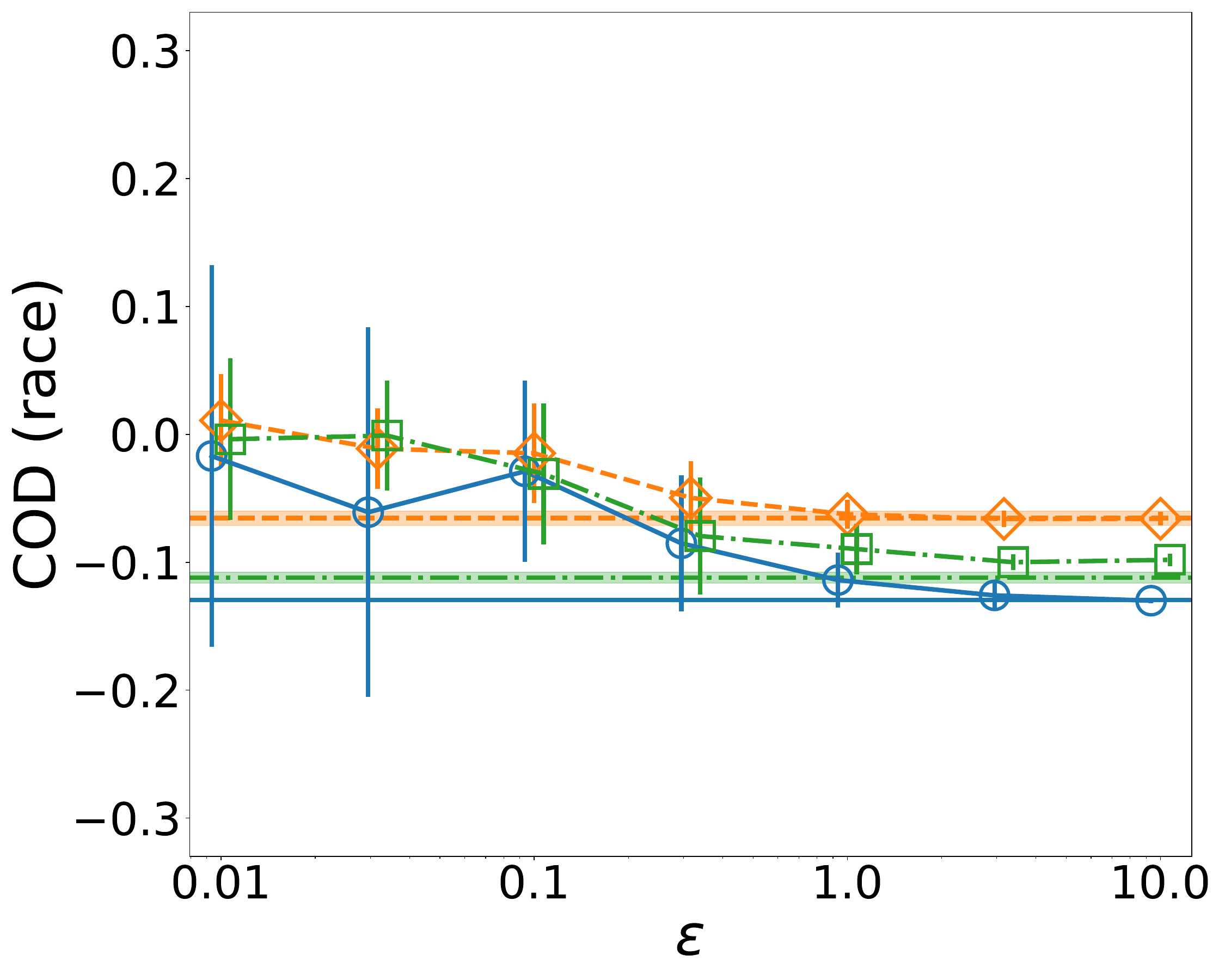}
    \includegraphics[width=0.325\linewidth]{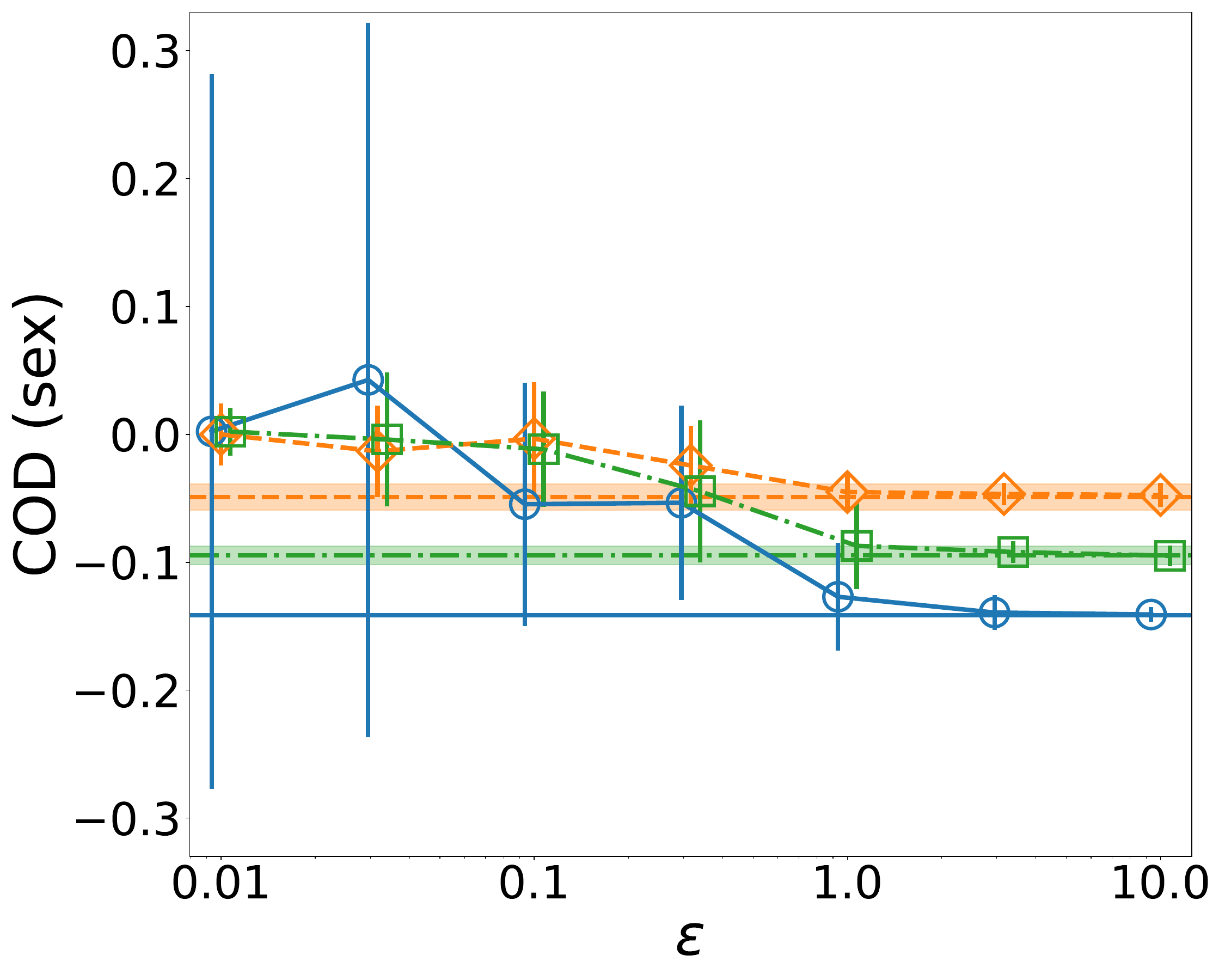}
    \includegraphics[width=0.325\linewidth]{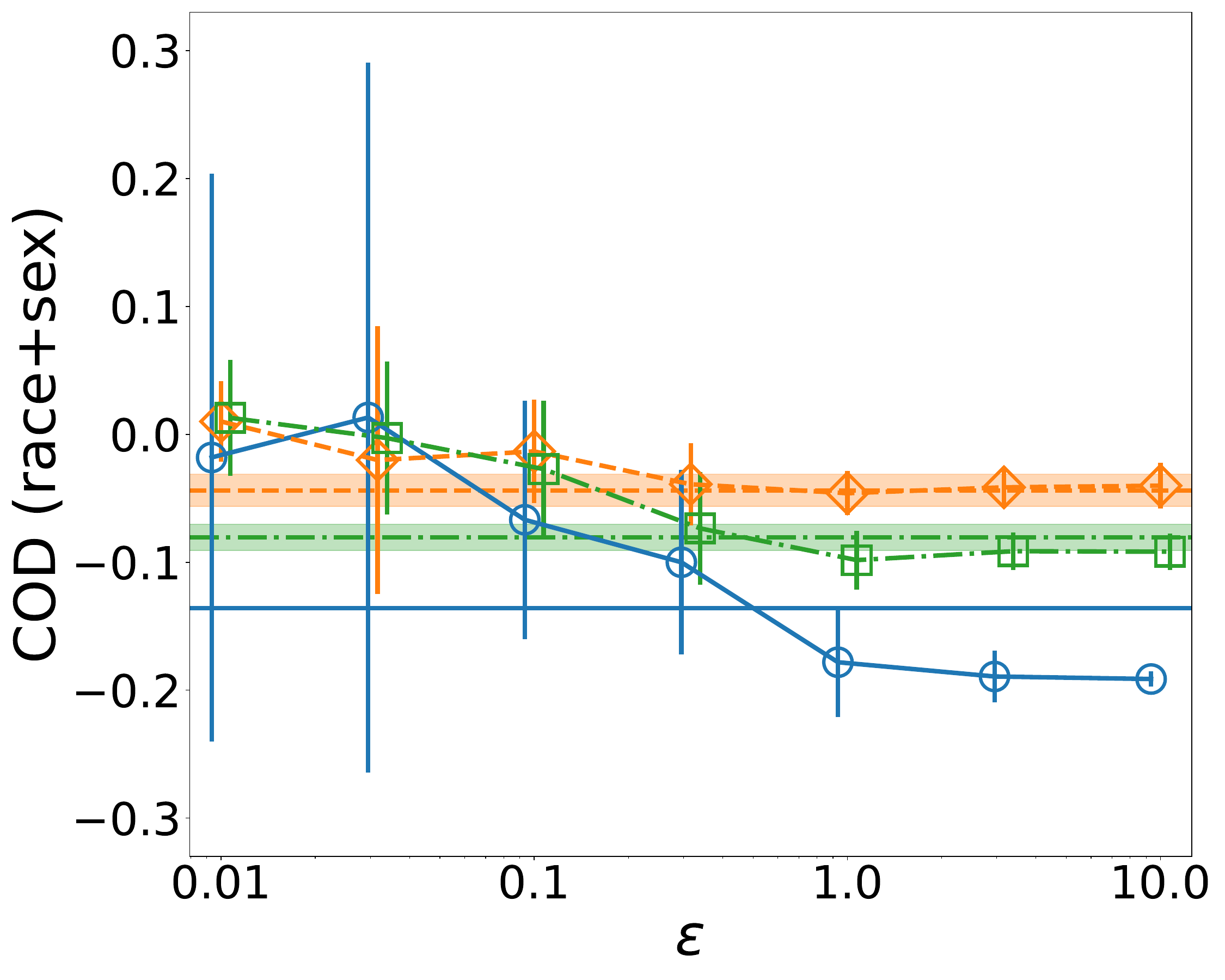}
    \caption{Mean $\pm$ 1 SD (error bars and shaded regions) COD, measured with race, sex, and race+sex as the protected attribute, for the COMPAS experiment.}
    \label{fig:compas_COD}
\end{figure}

\begin{figure}[!htb]
    \centering
    \includegraphics[width=0.325\linewidth]{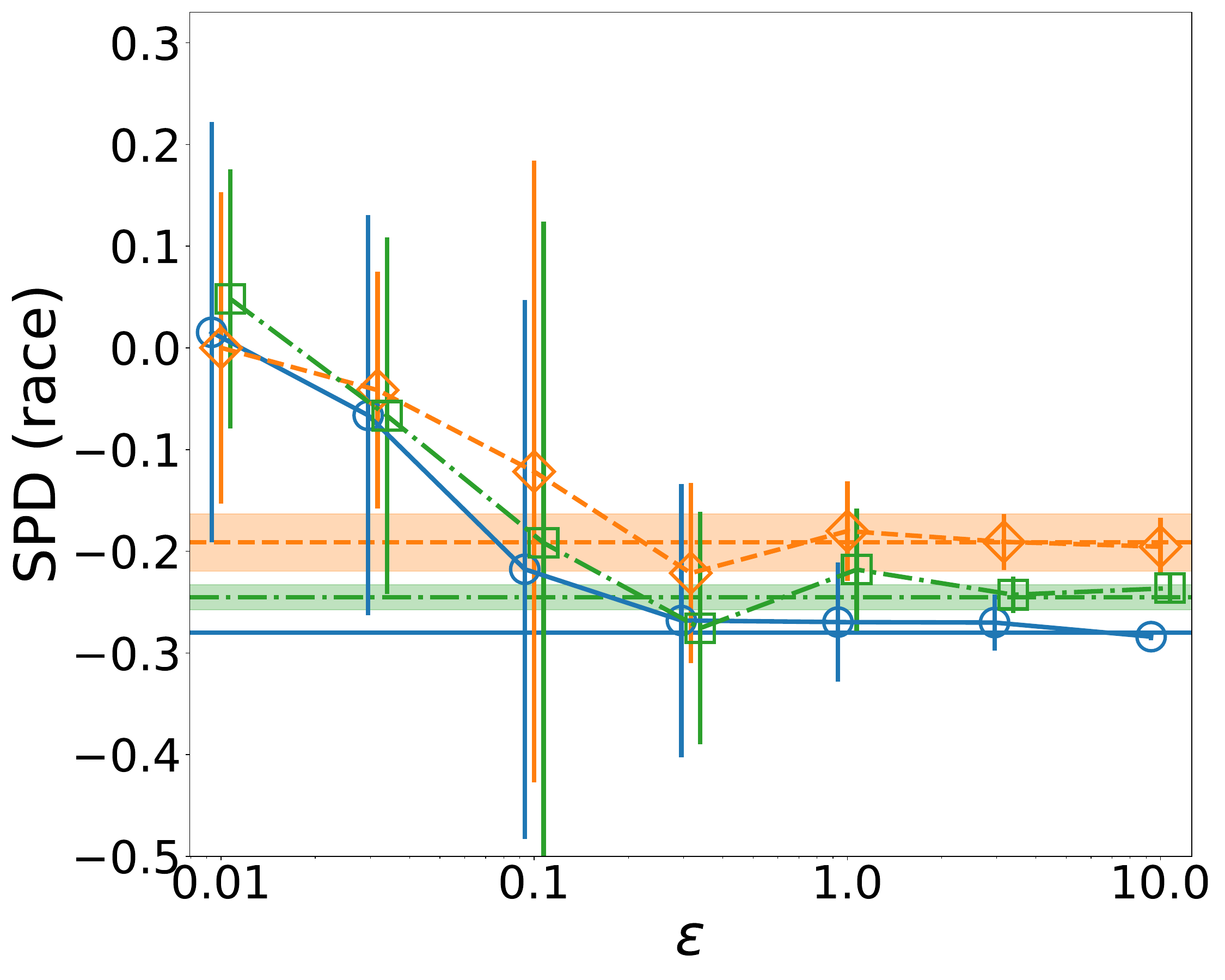}
    \includegraphics[width=0.325\linewidth]{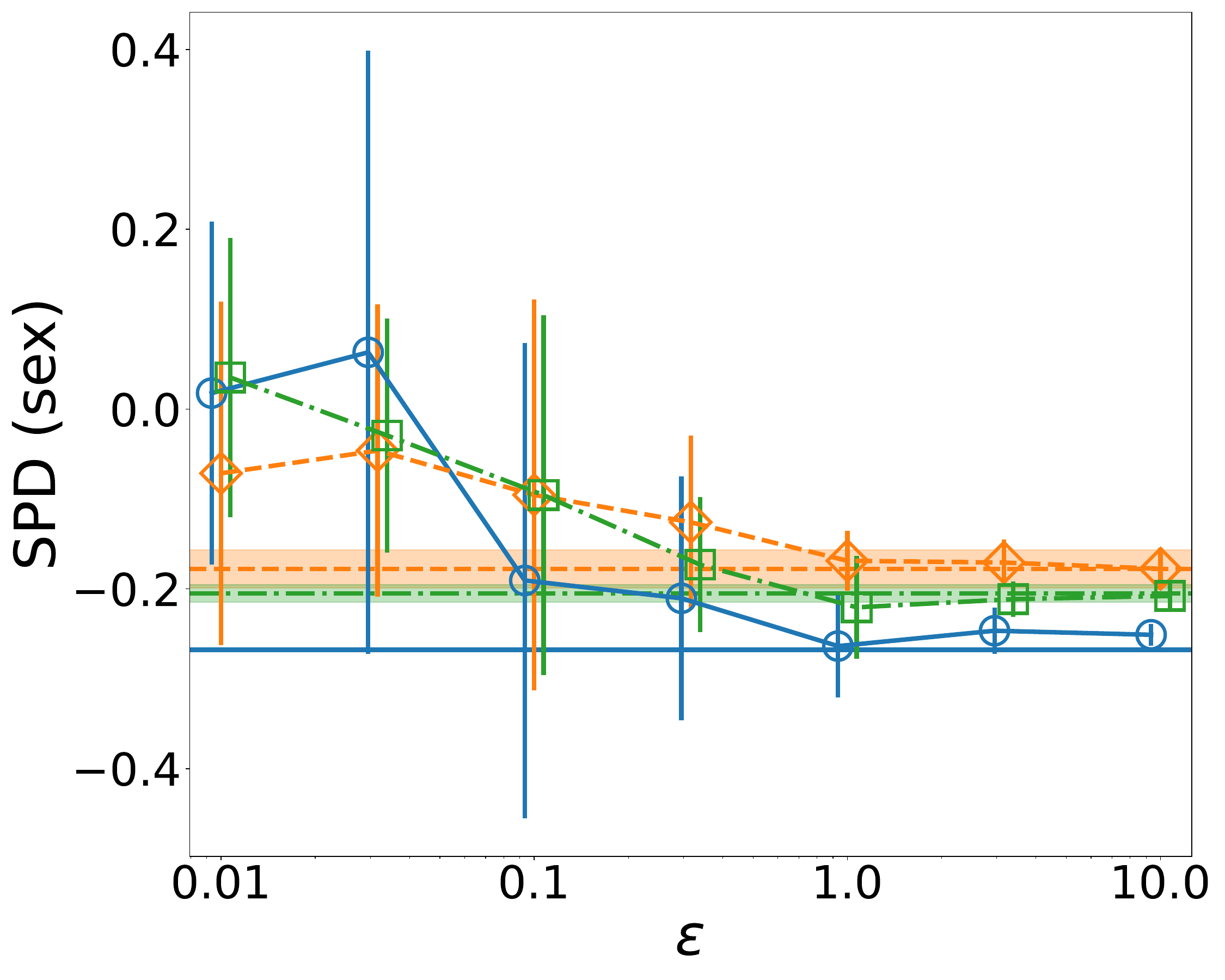}
    \includegraphics[width=0.325\linewidth]{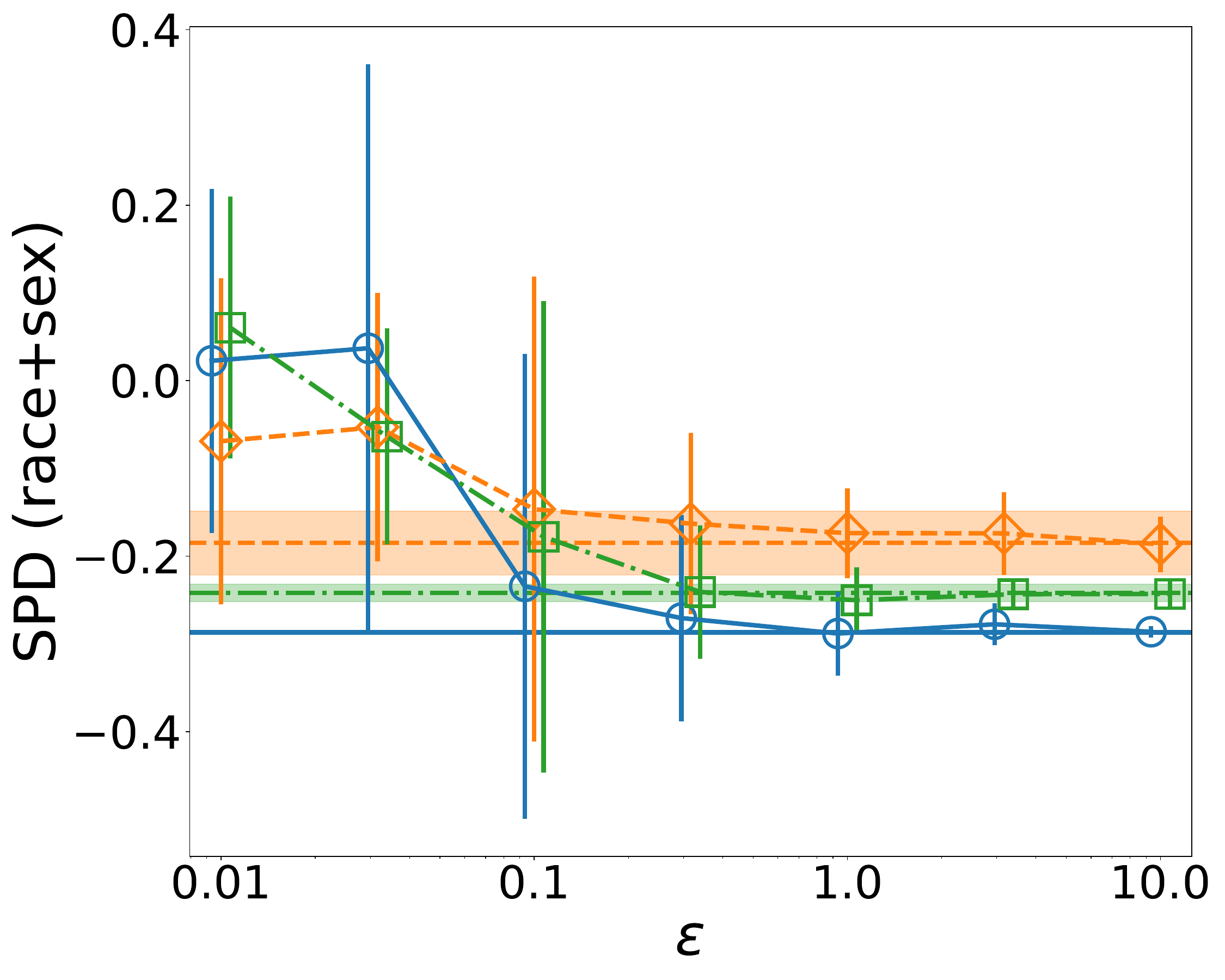}
    \caption{Mean $\pm$ 1 SD (error bars and shaded regions) SPD, with race, sex, and race+sex as the protected attribute, for the COMPAS experiment.}
    \label{fig:compas_SPD}
\end{figure}

\begin{figure}[!htb]
    \centering
    \includegraphics[width=0.325\linewidth]{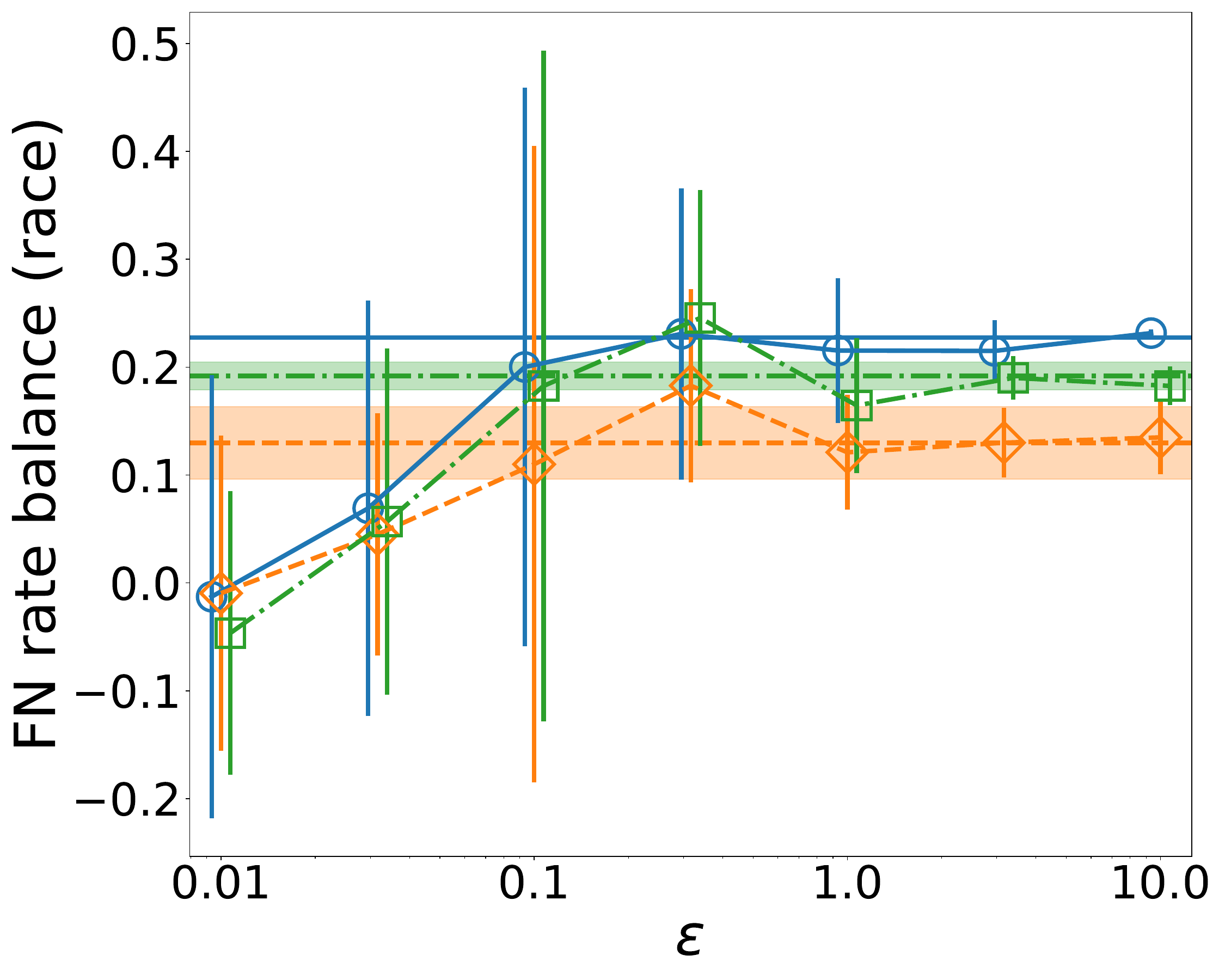}
    \includegraphics[width=0.325\linewidth]{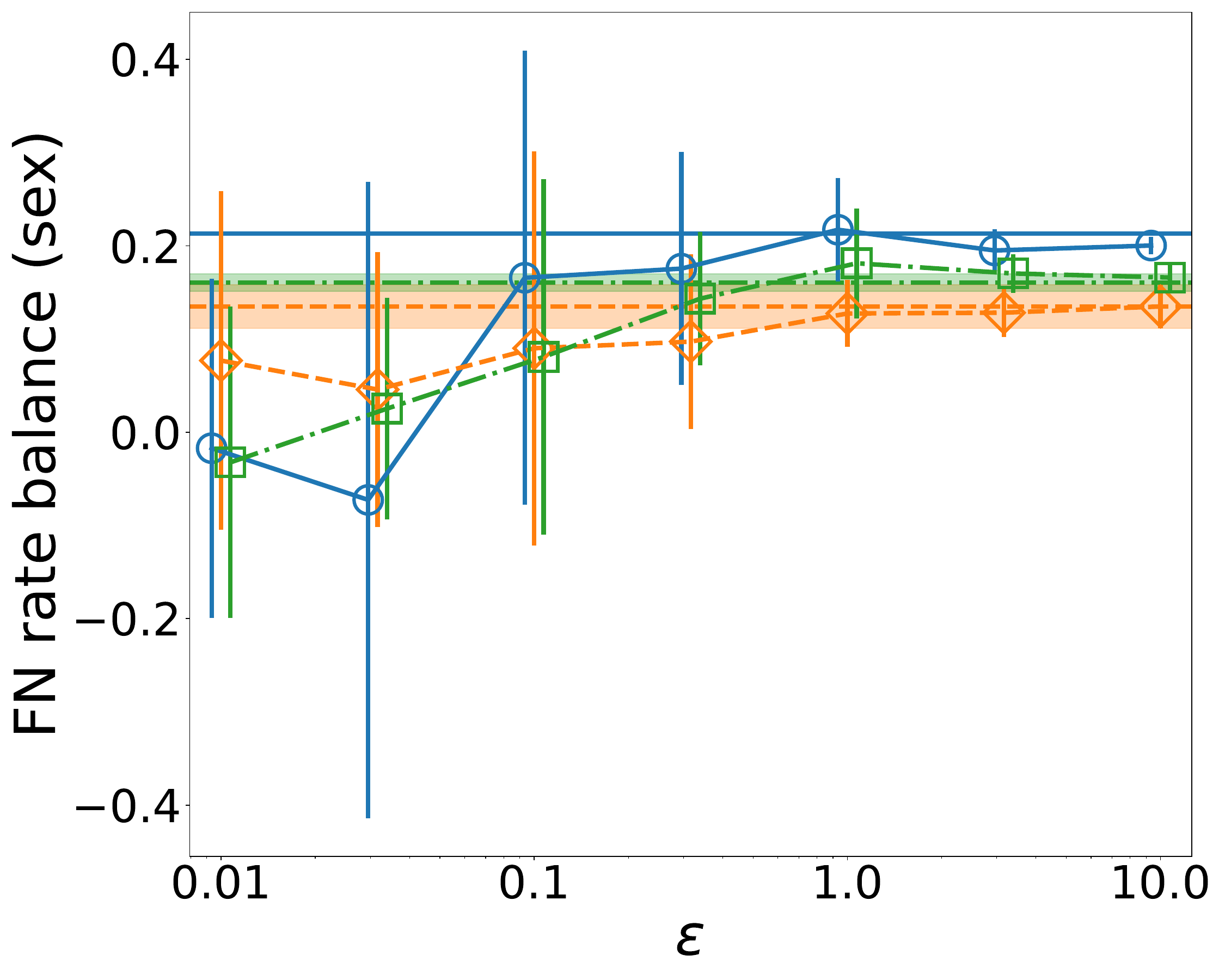}
    \includegraphics[width=0.325\linewidth]{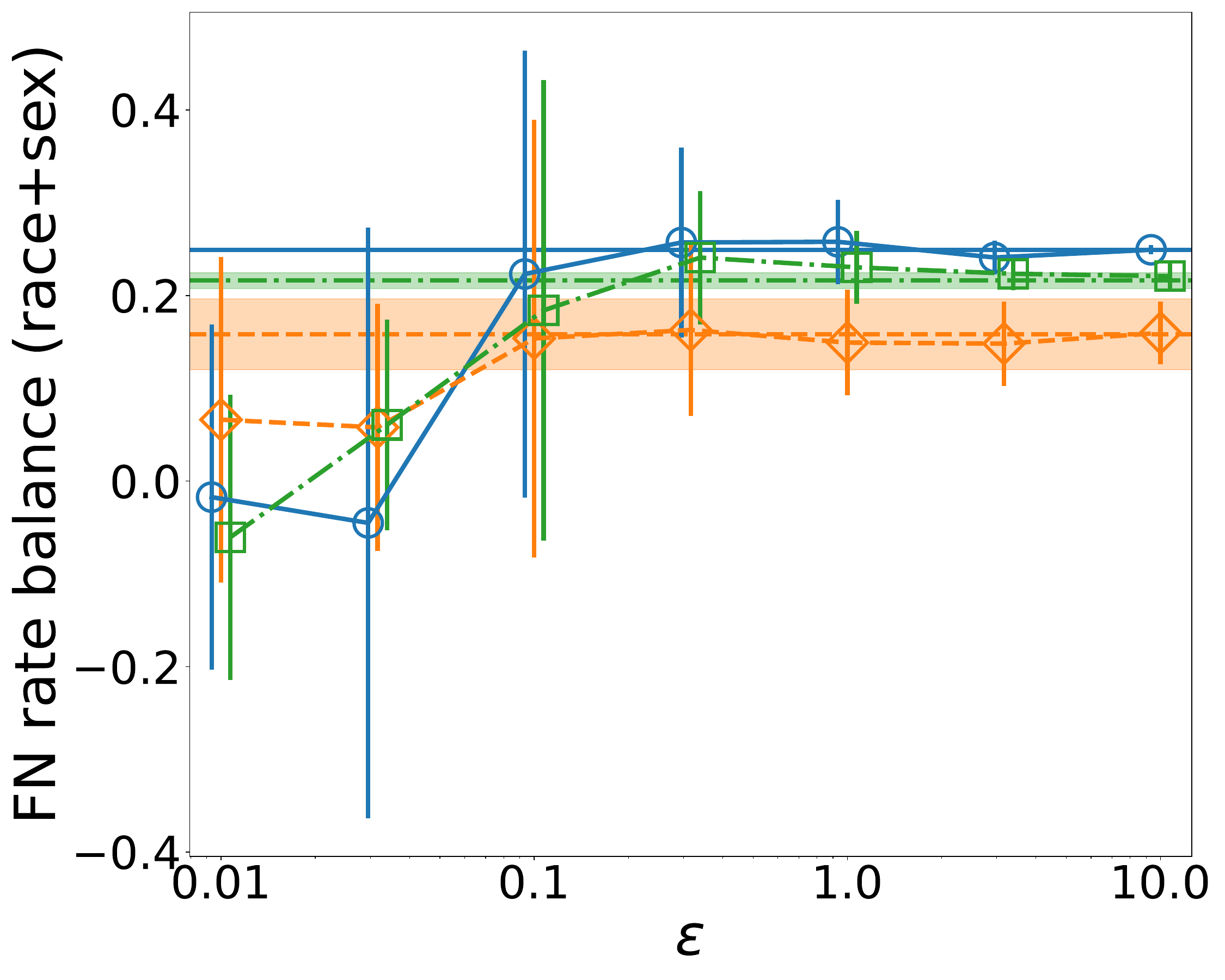}
    \caption{Mean $\pm$ 1 SD (error bars and shaded regions) FN rate balance, with race, sex, and race+sex as the protected attribute, for the COMPAS experiment.}
    \label{fig:compas_FNR_balance}
\end{figure}

\begin{figure}[!htb]
    \centering
    \includegraphics[width=0.325\linewidth]{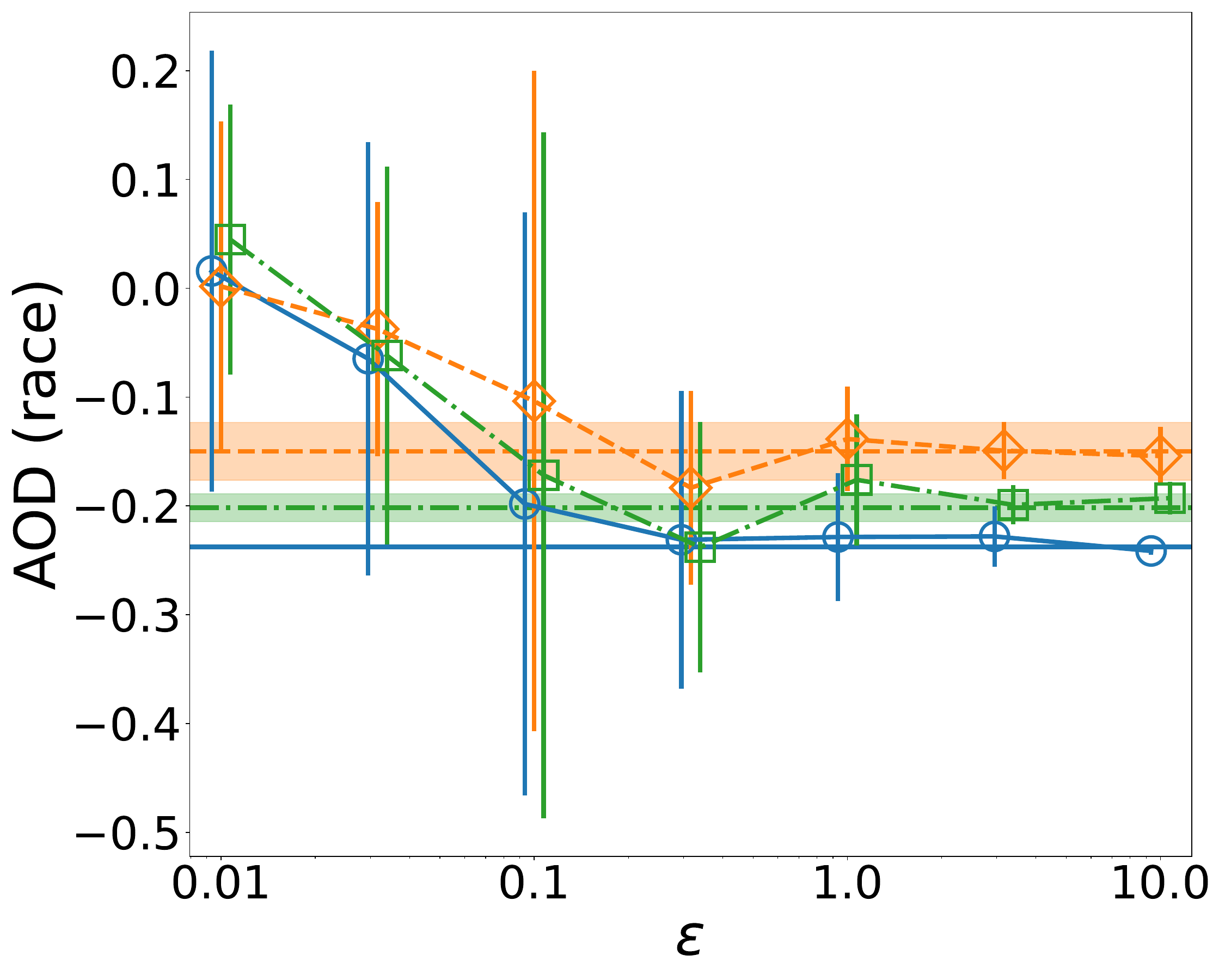}
    \includegraphics[width=0.325\linewidth]{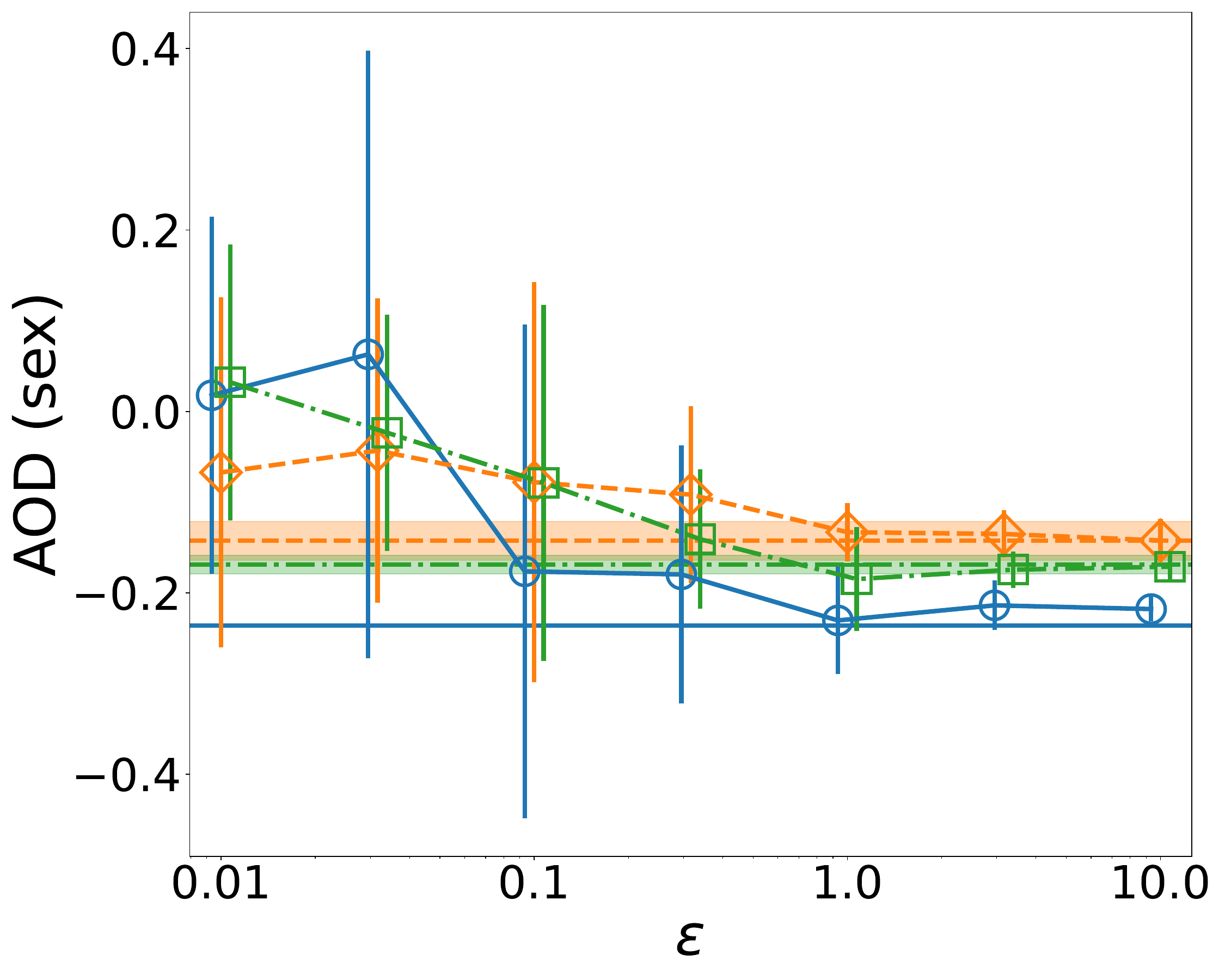}
    \includegraphics[width=0.325\linewidth]{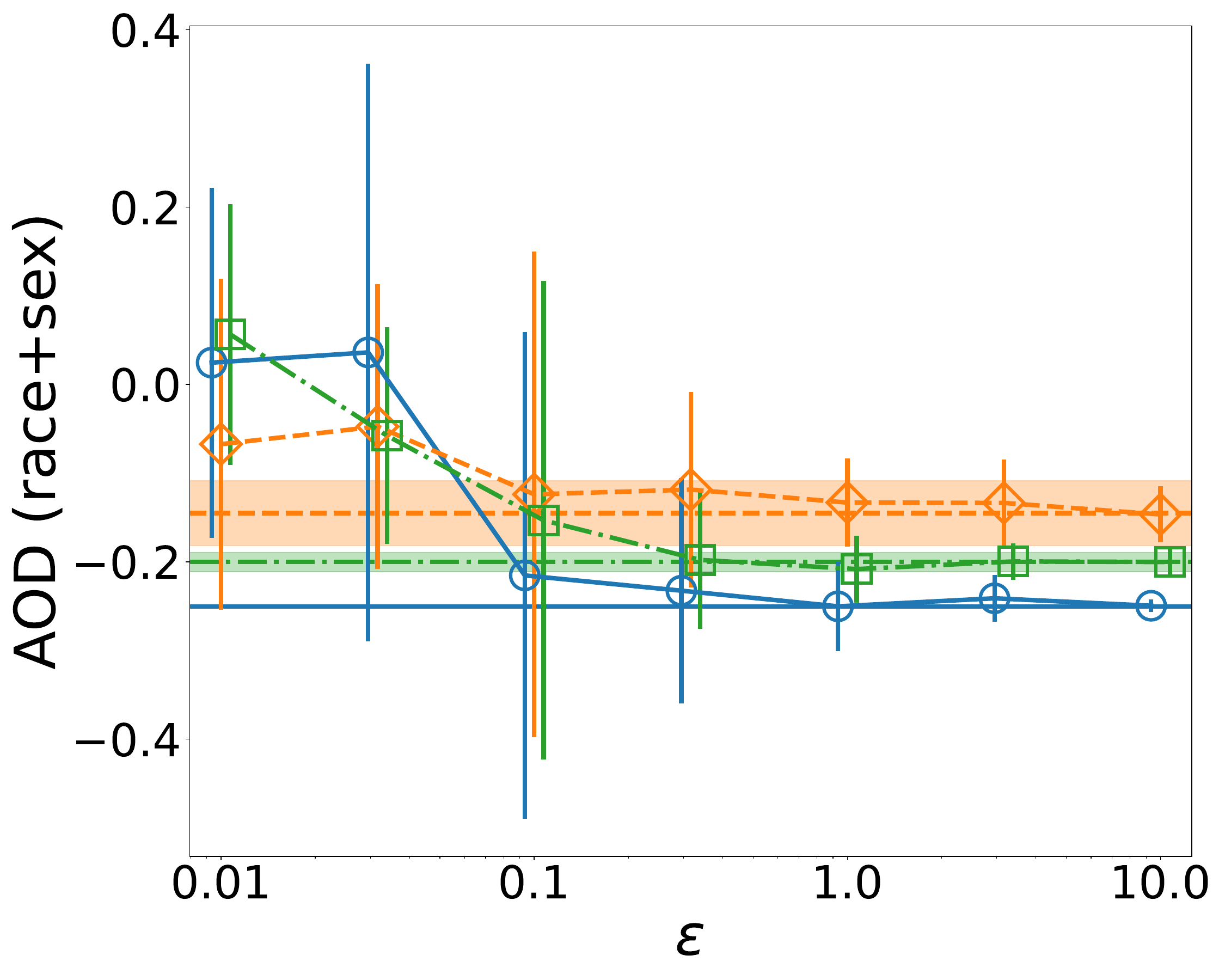}
    \caption{Mean $\pm$ 1 SD (error bars and shaded regions) AOD, with race, sex, and race+sex as the protected attribute, for the COMPAS experiment.}
    \label{fig:compas_AOD}
\end{figure}

\begin{figure}[!htb]
    \centering
    \includegraphics[width=0.325\linewidth]{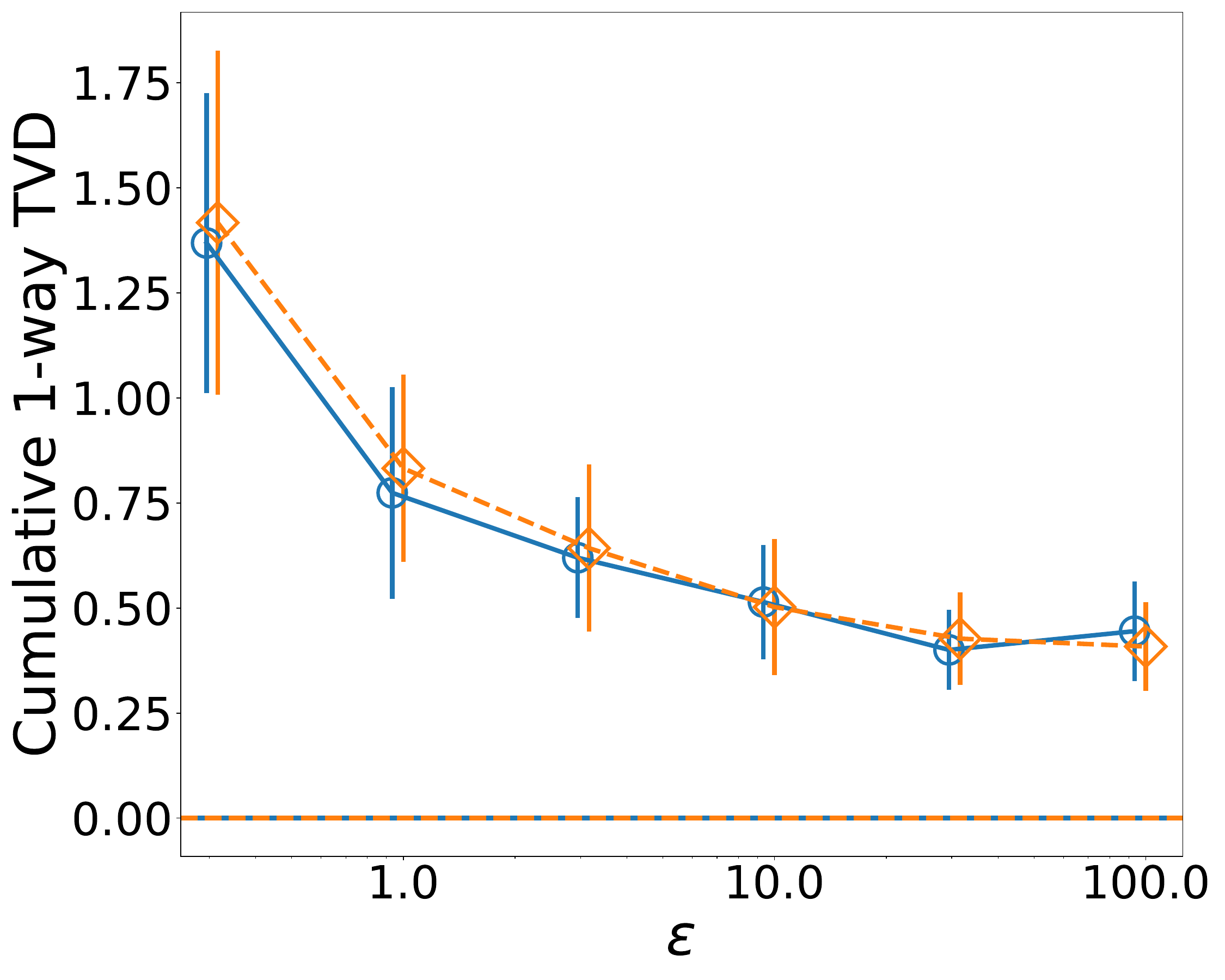}
    \includegraphics[width=0.325\linewidth]{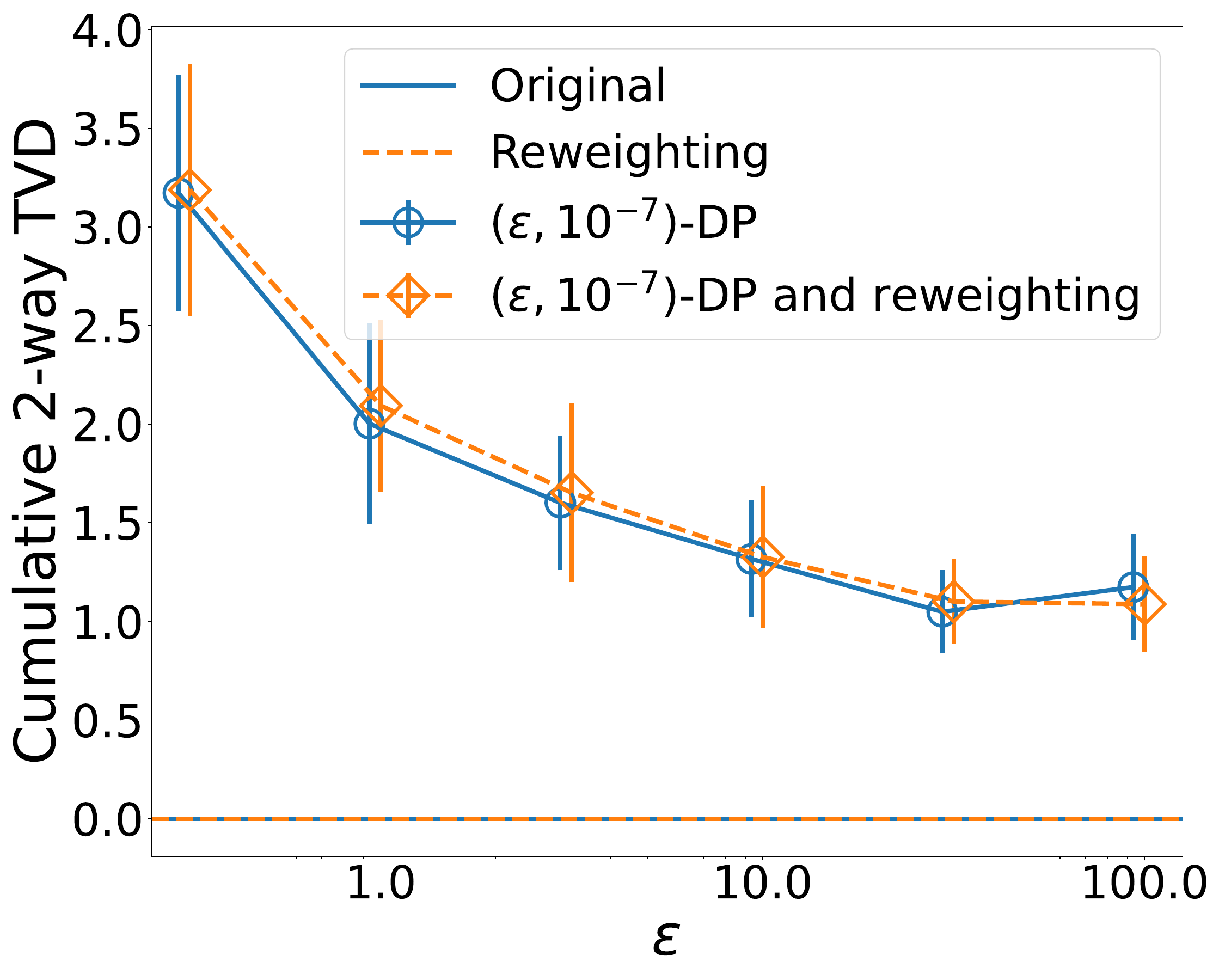}
    \includegraphics[width=0.325\linewidth]{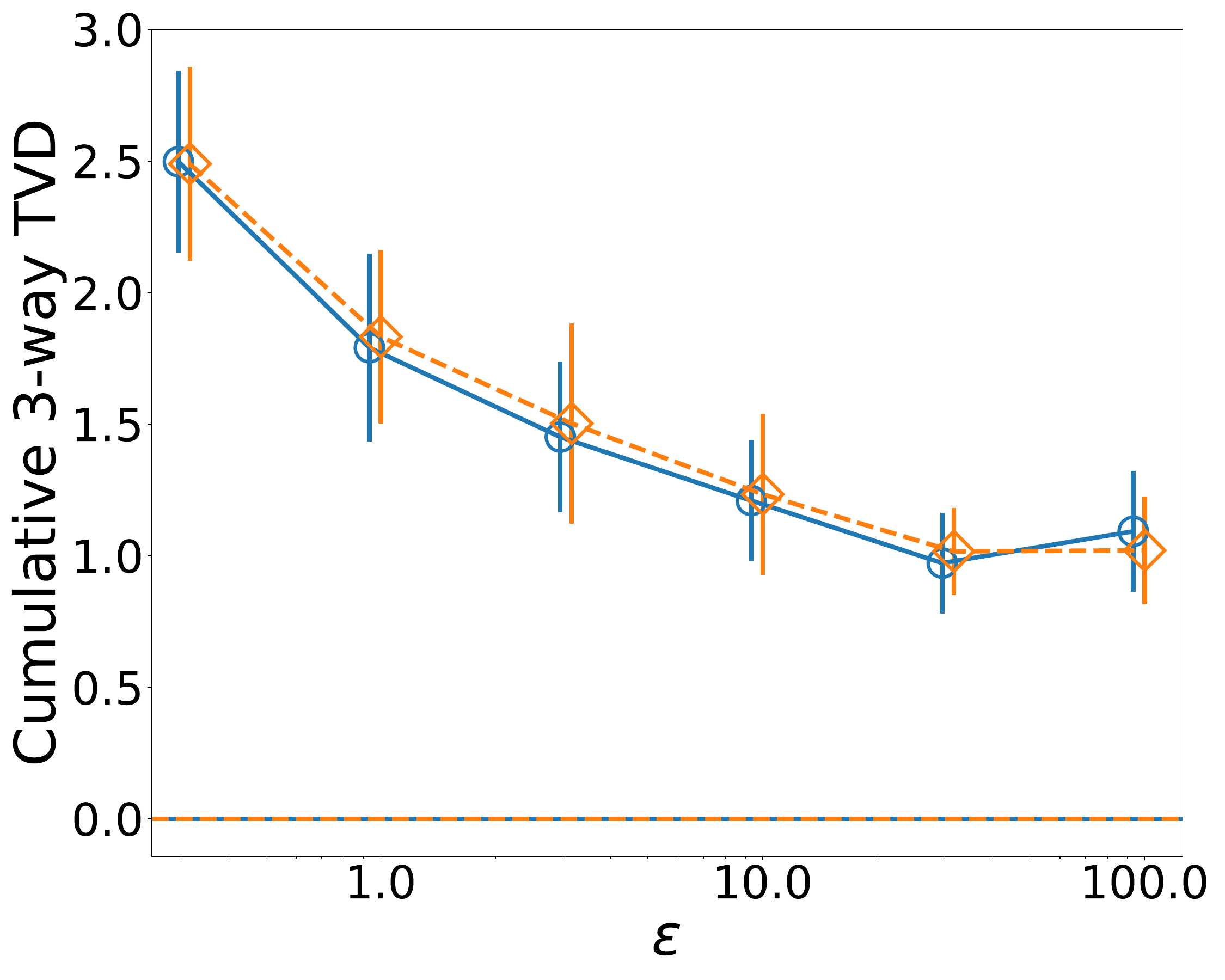}
    \caption{Mean $\pm$ 1 SD (error bars and shaded regions) summed TVD in each marginal set for 1-way, 2-way, and 3-way marginals between the synthetic data vs the original data for the DP-CTGAN + RW experiment in the Adult data with continuous age.}
    \label{fig:Adult_cont_TVD}
\end{figure}

\begin{figure}[!htb]
    \centering
    \includegraphics[width=0.35\linewidth]{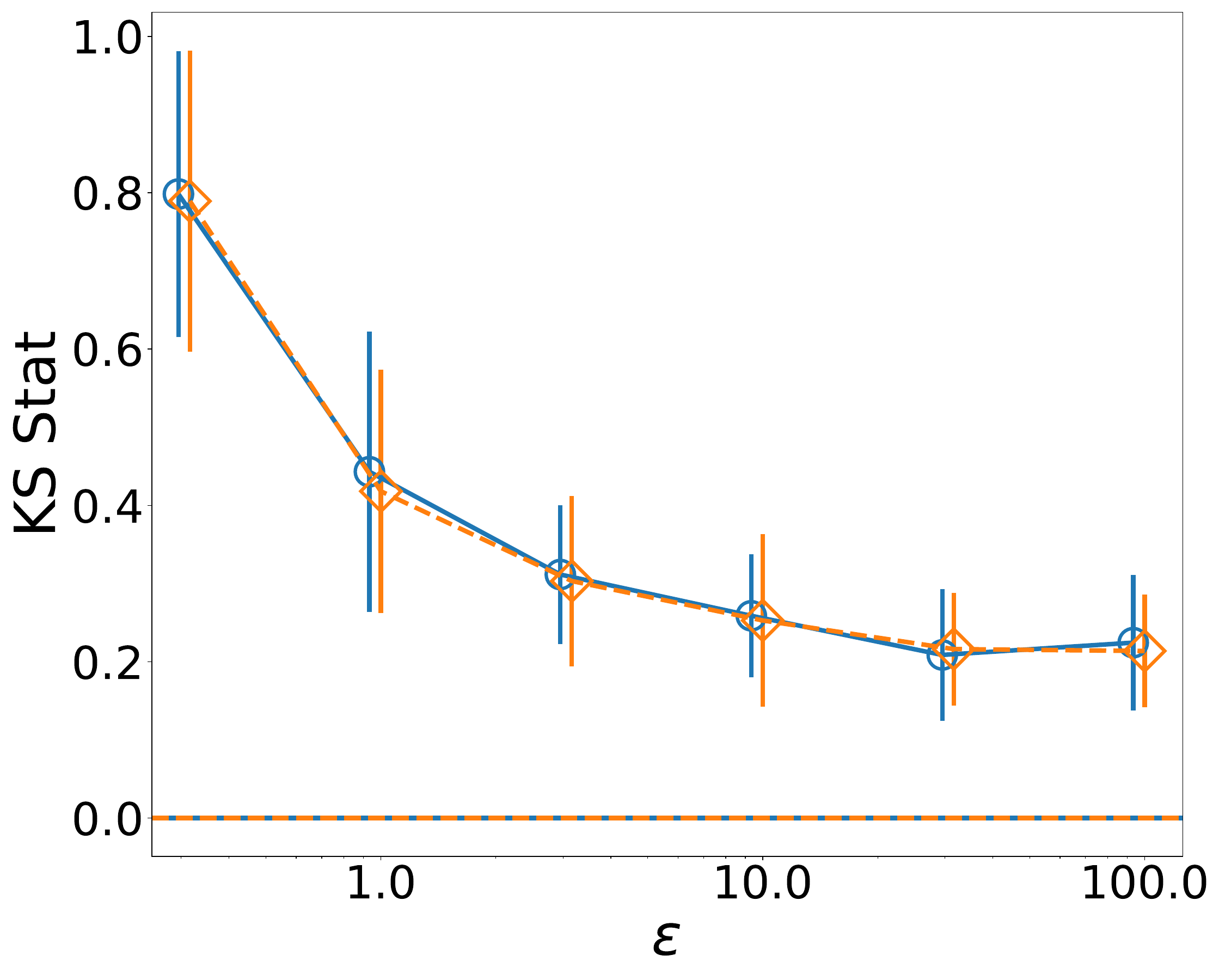}
    \includegraphics[width=0.35\linewidth]{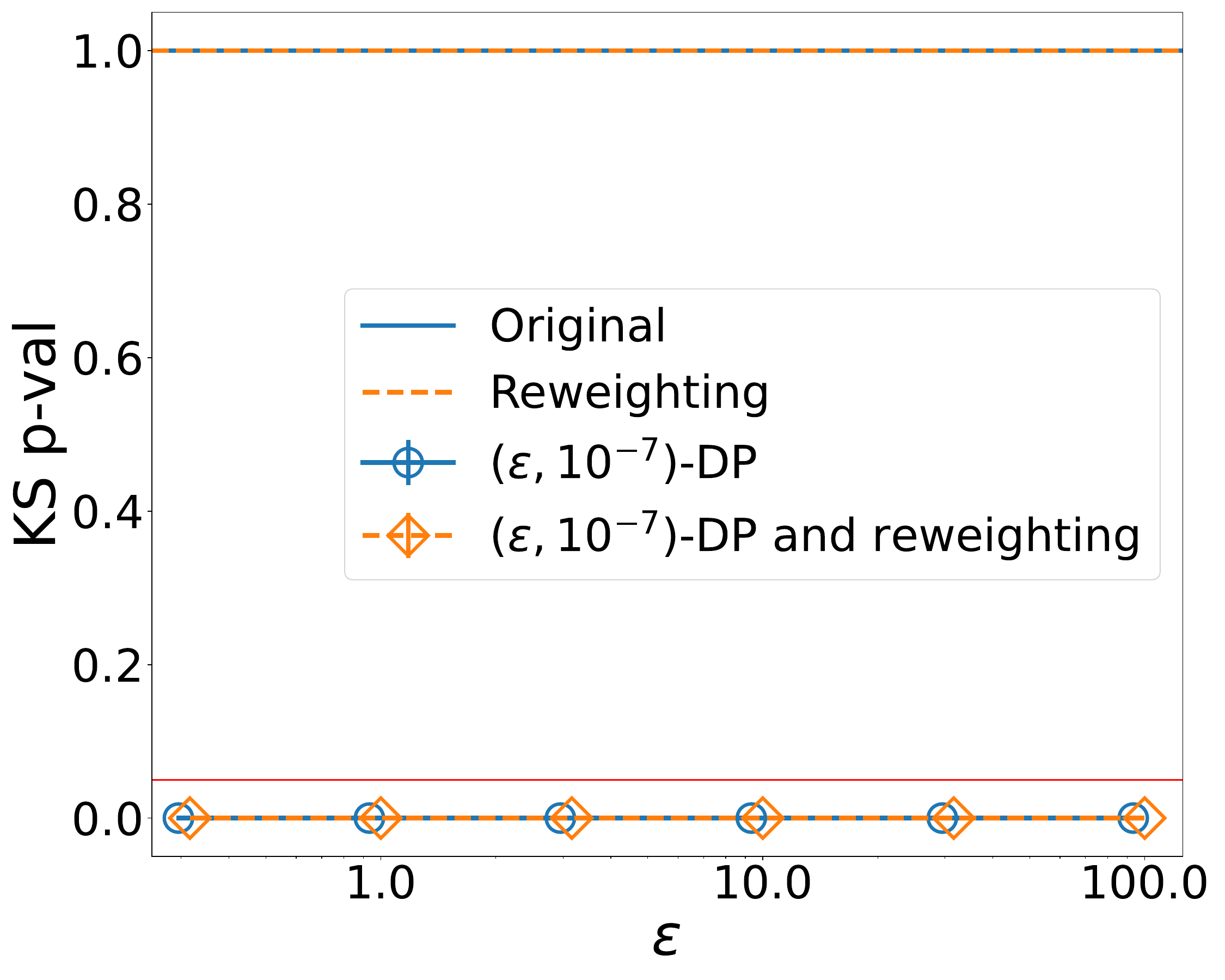}
    \caption{Mean $\pm$ 1 SD (error bars and shaded regions) test statistic and corresponding p-value for the KS test comparing original and synthetic datasets for the DP-CTGAN + RW experiment in the Adult data with continuous age.
    Statistical significance threshold of $\alpha=0.05$ is marked in red in the plot on the right.}
    \label{fig:Adult_cont_KS}
\end{figure}

\begin{figure}[!htb]
    \centering
    \includegraphics[width=0.325\linewidth]{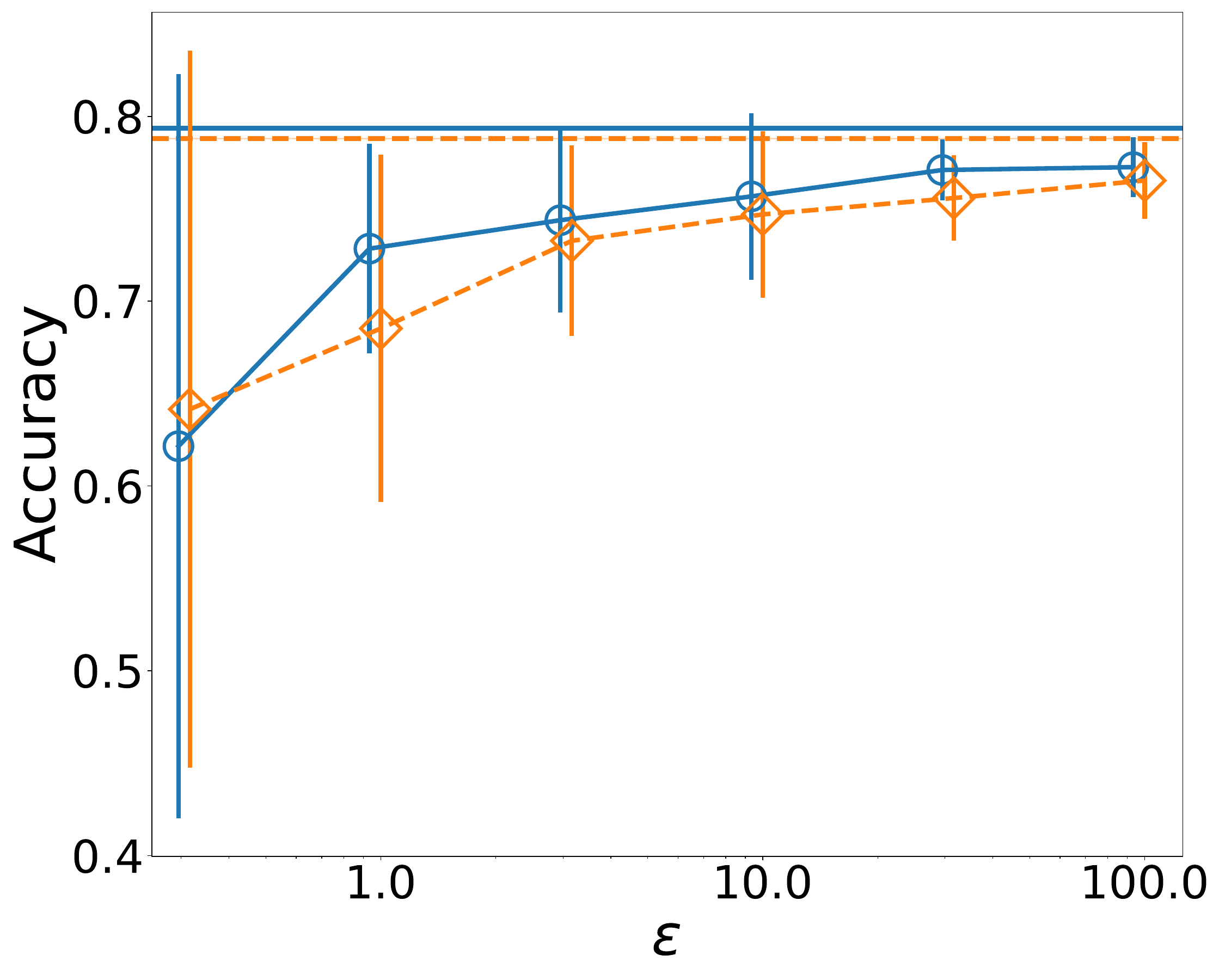}
    \includegraphics[width=0.325\linewidth]{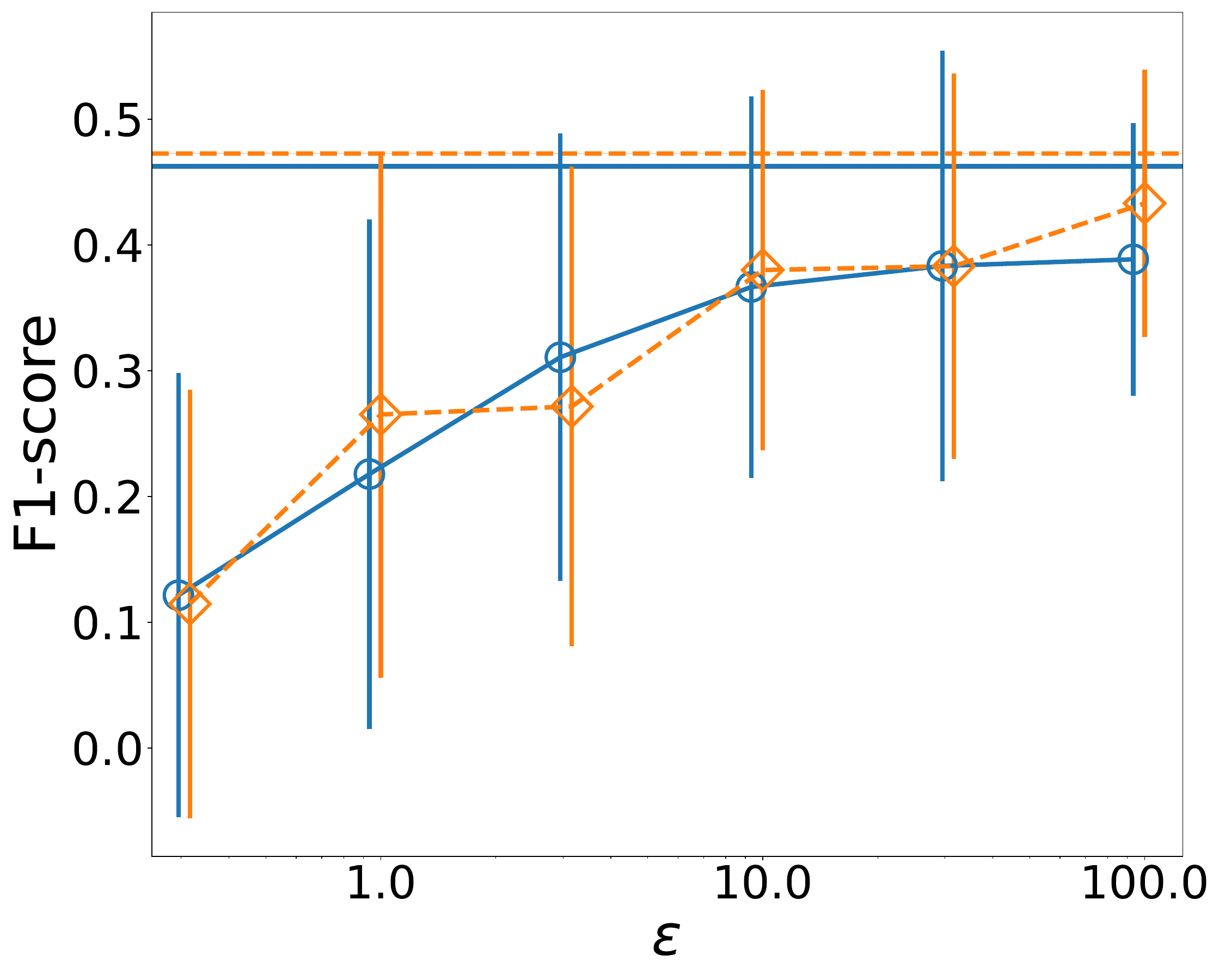}
    \includegraphics[width=0.325\linewidth]{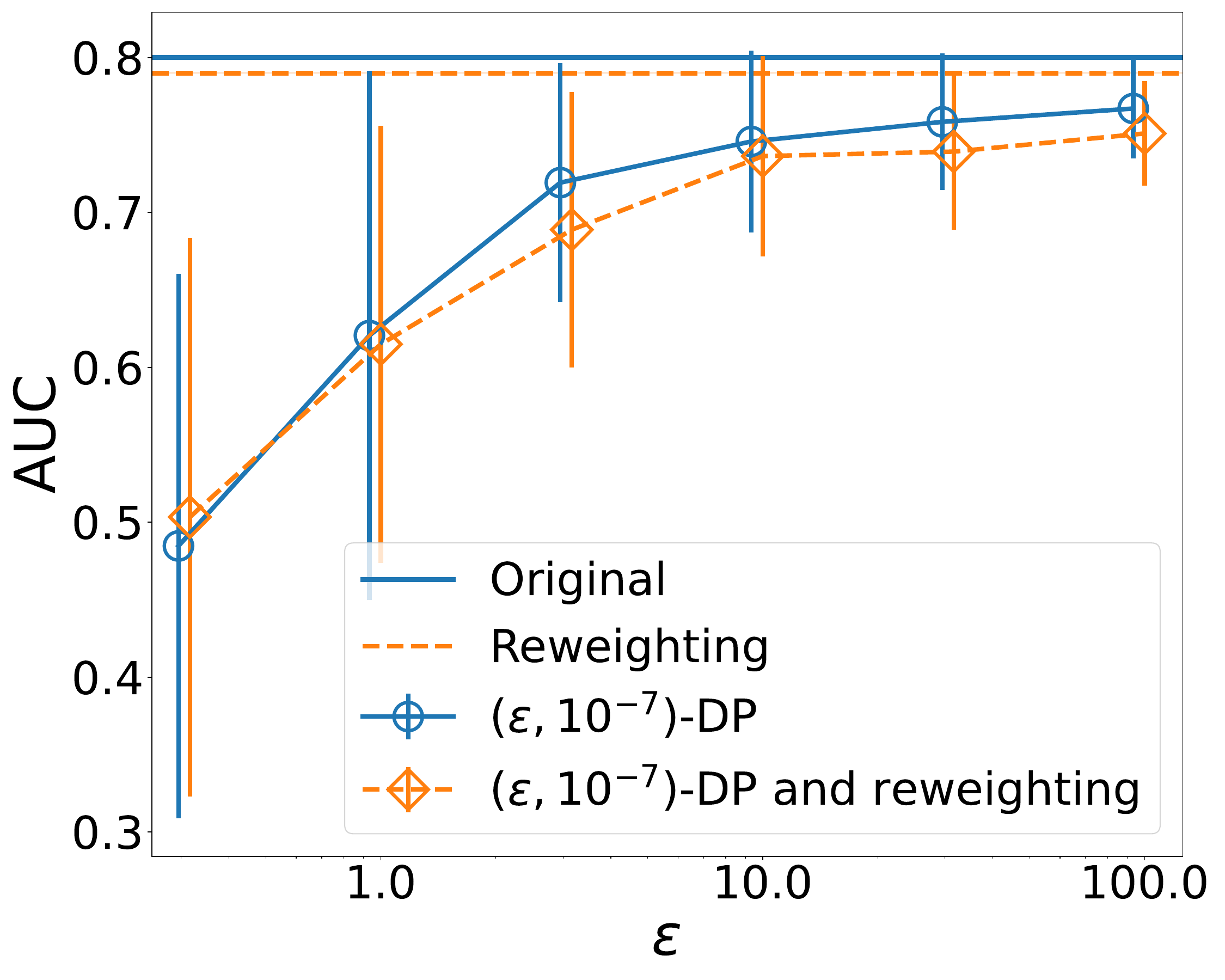}
    \caption{Mean $\pm$ 1 SD (error bars and shaded regions) prediction performance of the logistic regression model trained on SAFES synthetic data for the DP-CTGAN + RW experiment  in the Adult data with continuous age.}
    \label{fig:Adult_cont_pred}
\end{figure}

\begin{figure}[!htb]
    \includegraphics[width=0.325\linewidth]{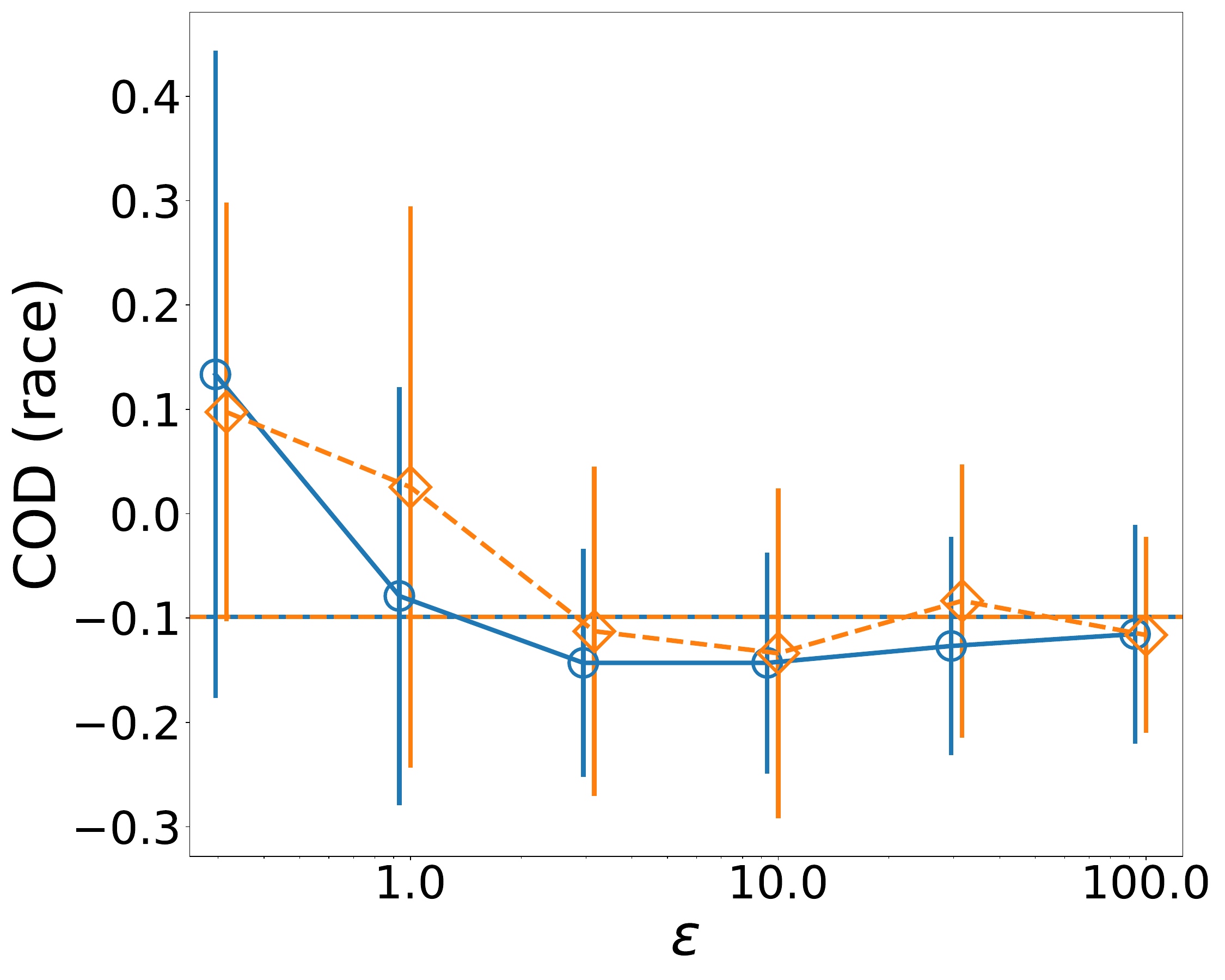}
    \includegraphics[width=0.325\linewidth]{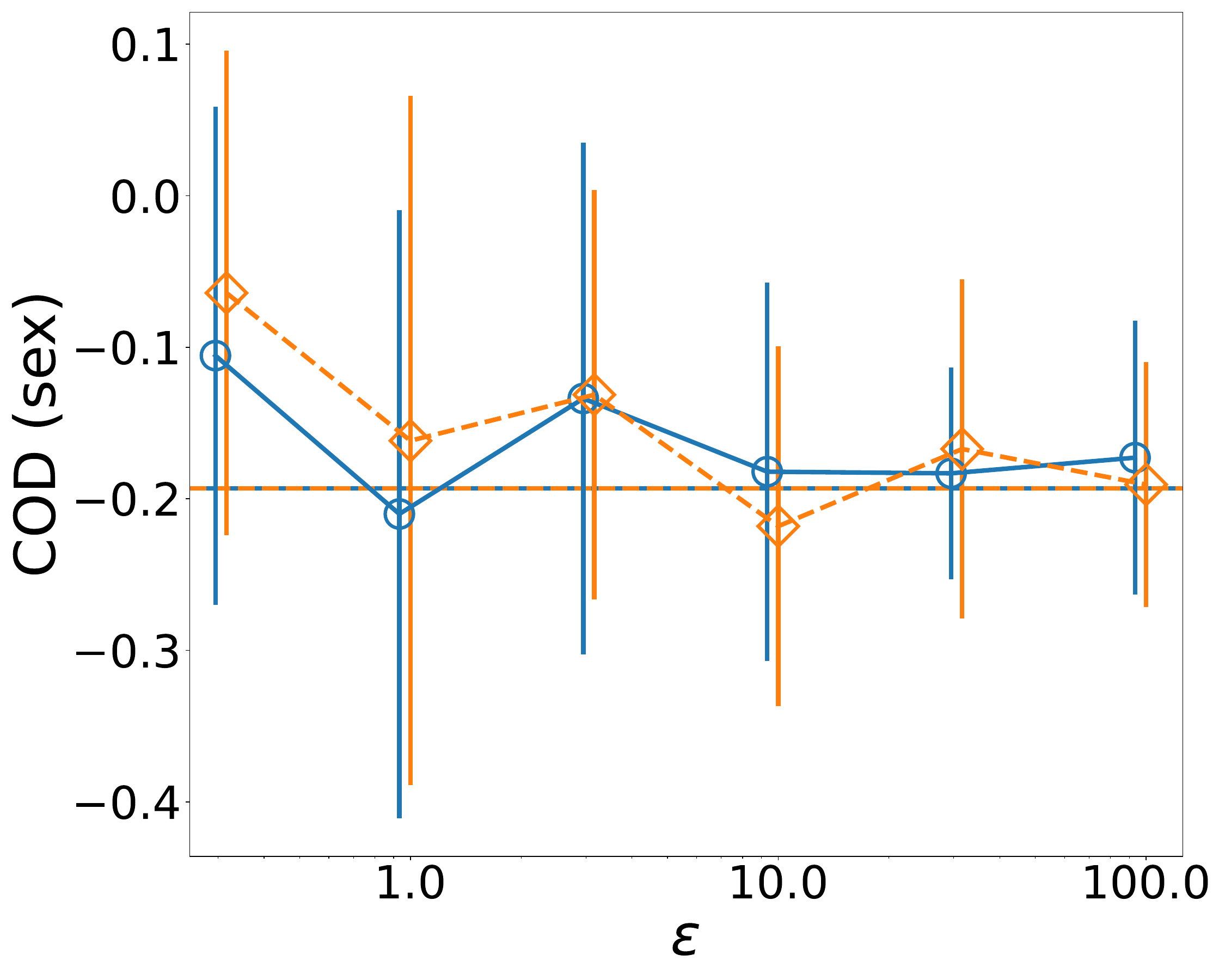}
    \includegraphics[width=0.325\linewidth]{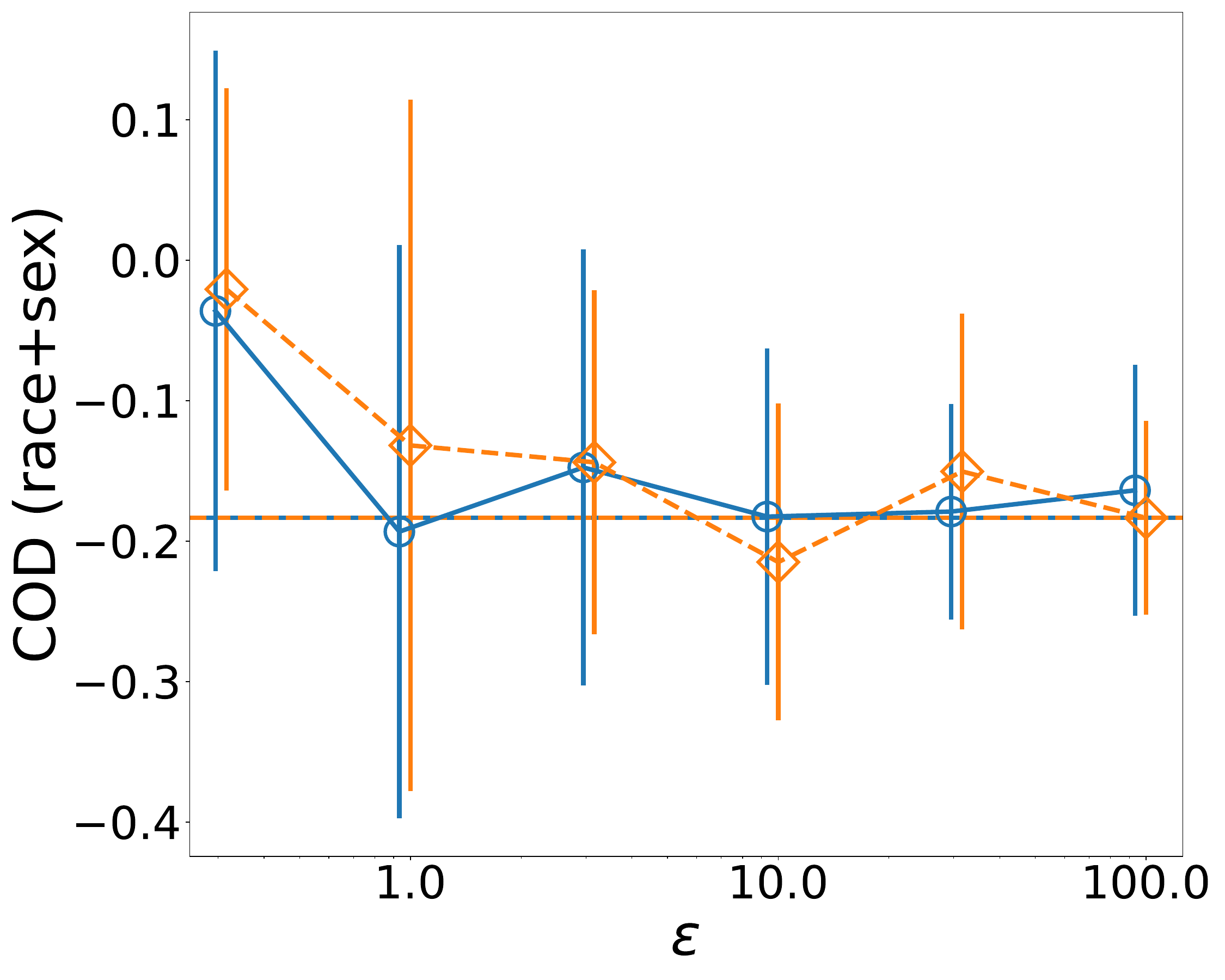}
    \caption{Mean $\pm$ 1 SD (error bars and shaded regions) COD, measured with race, sex, and race+sex as the protected attribute, for the DP-CTGAN + RW experiment  in the Adult data with continuous age.}
    \label{fig:Adult_cont_COD}
\end{figure}

\begin{figure}[!htb]
    \includegraphics[width=0.325\linewidth]{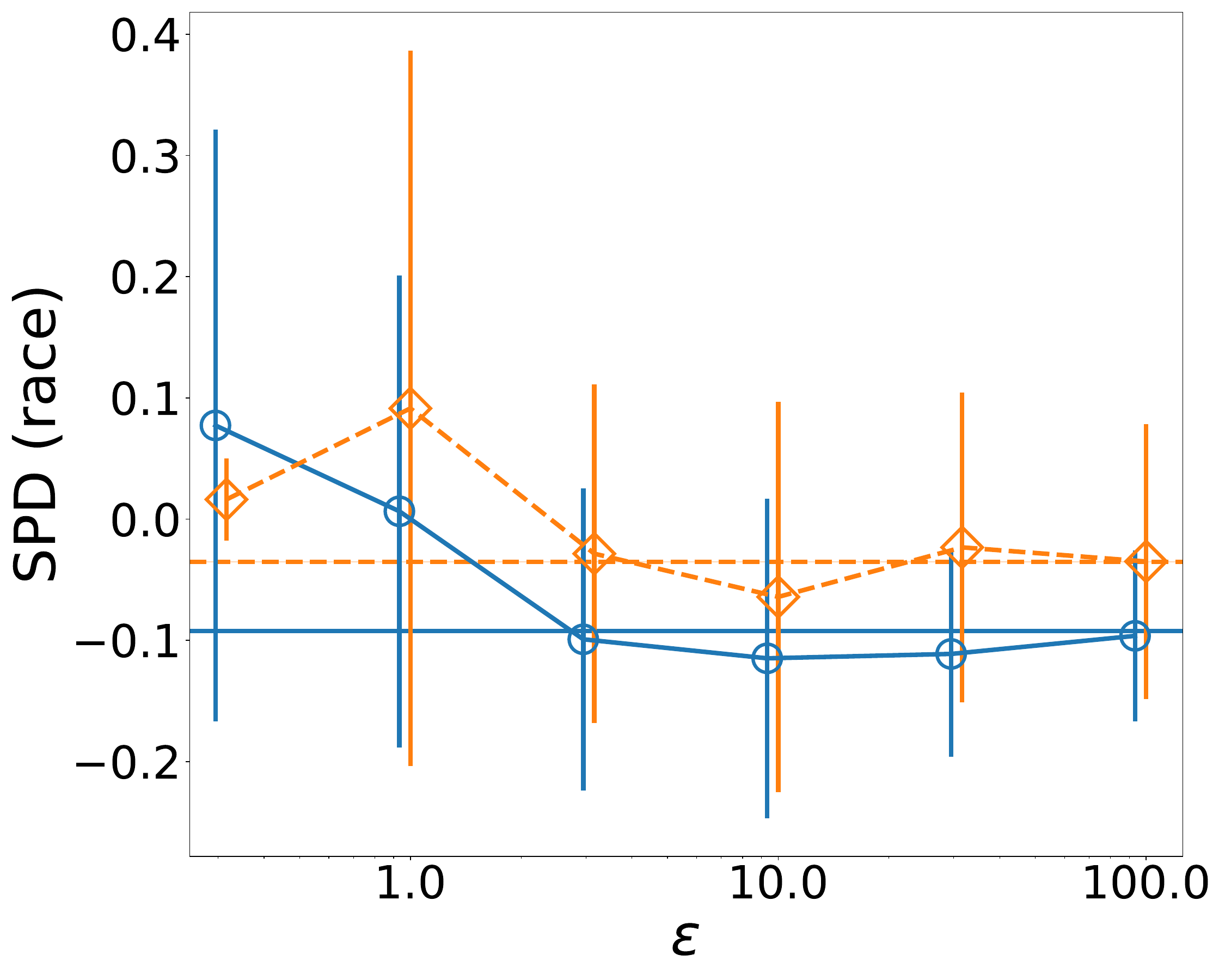}
    \includegraphics[width=0.325\linewidth]{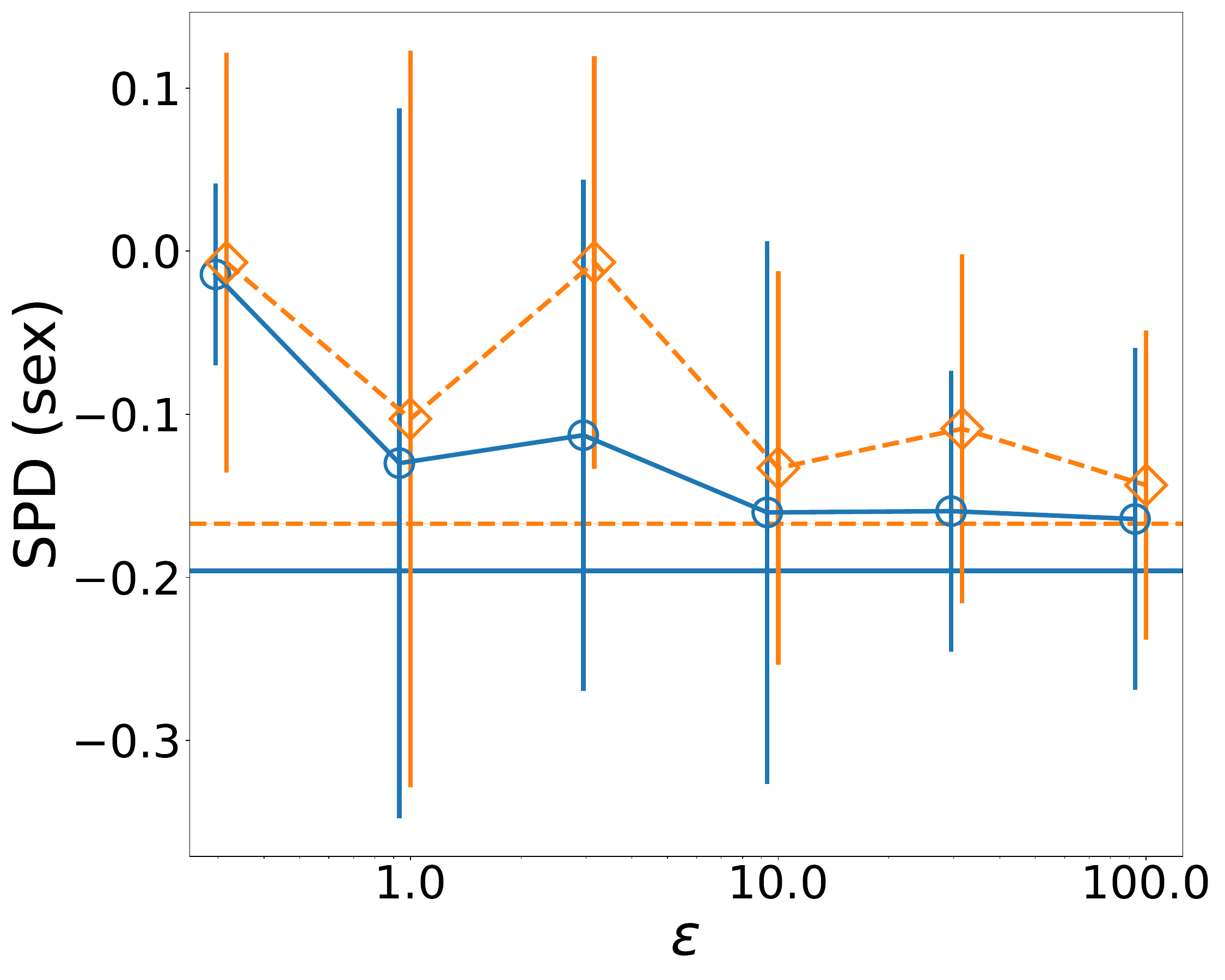}
    \includegraphics[width=0.325\linewidth]{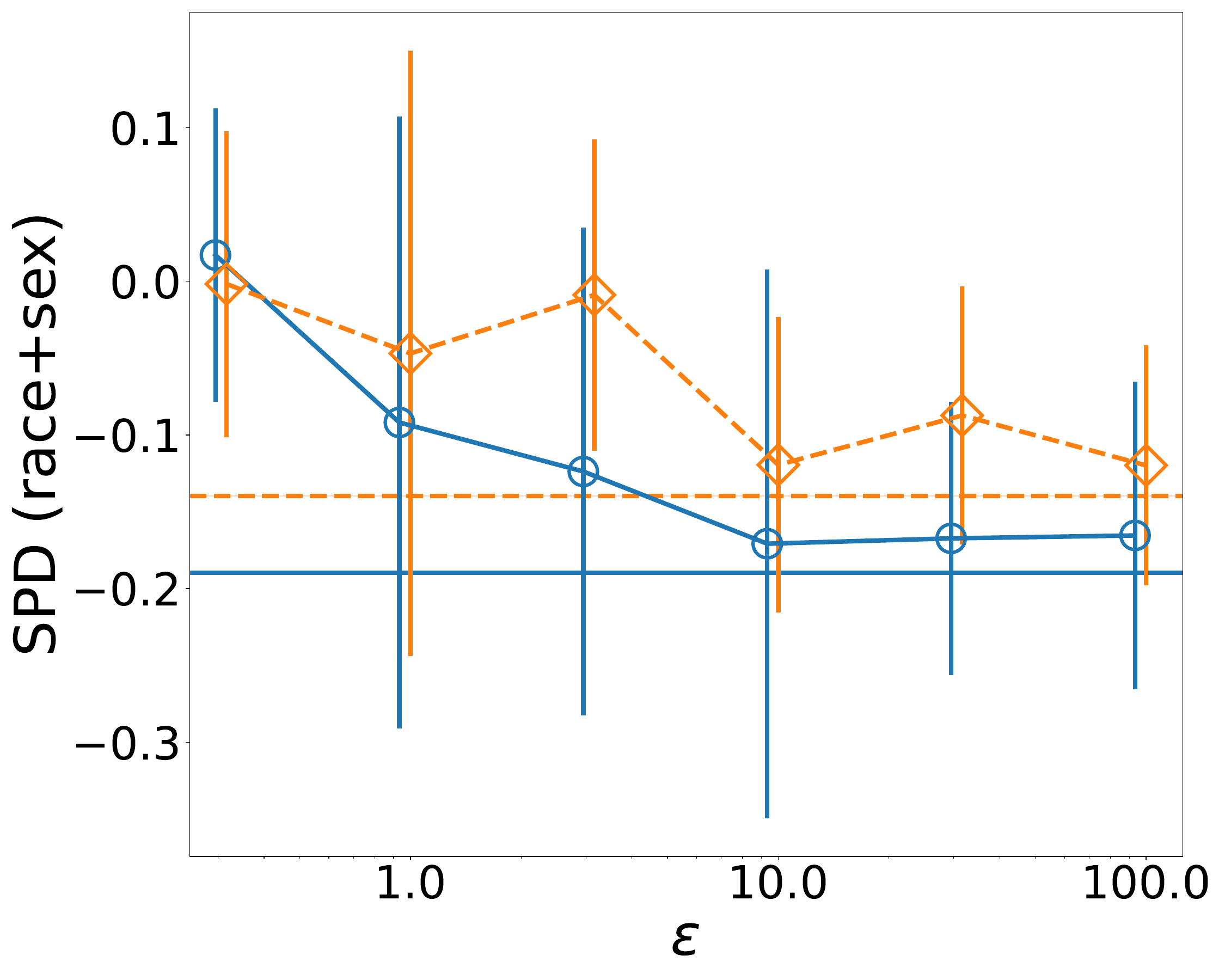}
    \caption{Mean $\pm$ 1 SD (error bars and shaded regions) SPD, with race, sex, and race+sex as the protected attribute, for the DP-CTGAN + RW experiment in the Adult data with continuous age.}
    \label{fig:Adult_cont_SPD}
\end{figure}

\begin{figure}[!htb]
    \includegraphics[width=0.325\linewidth]{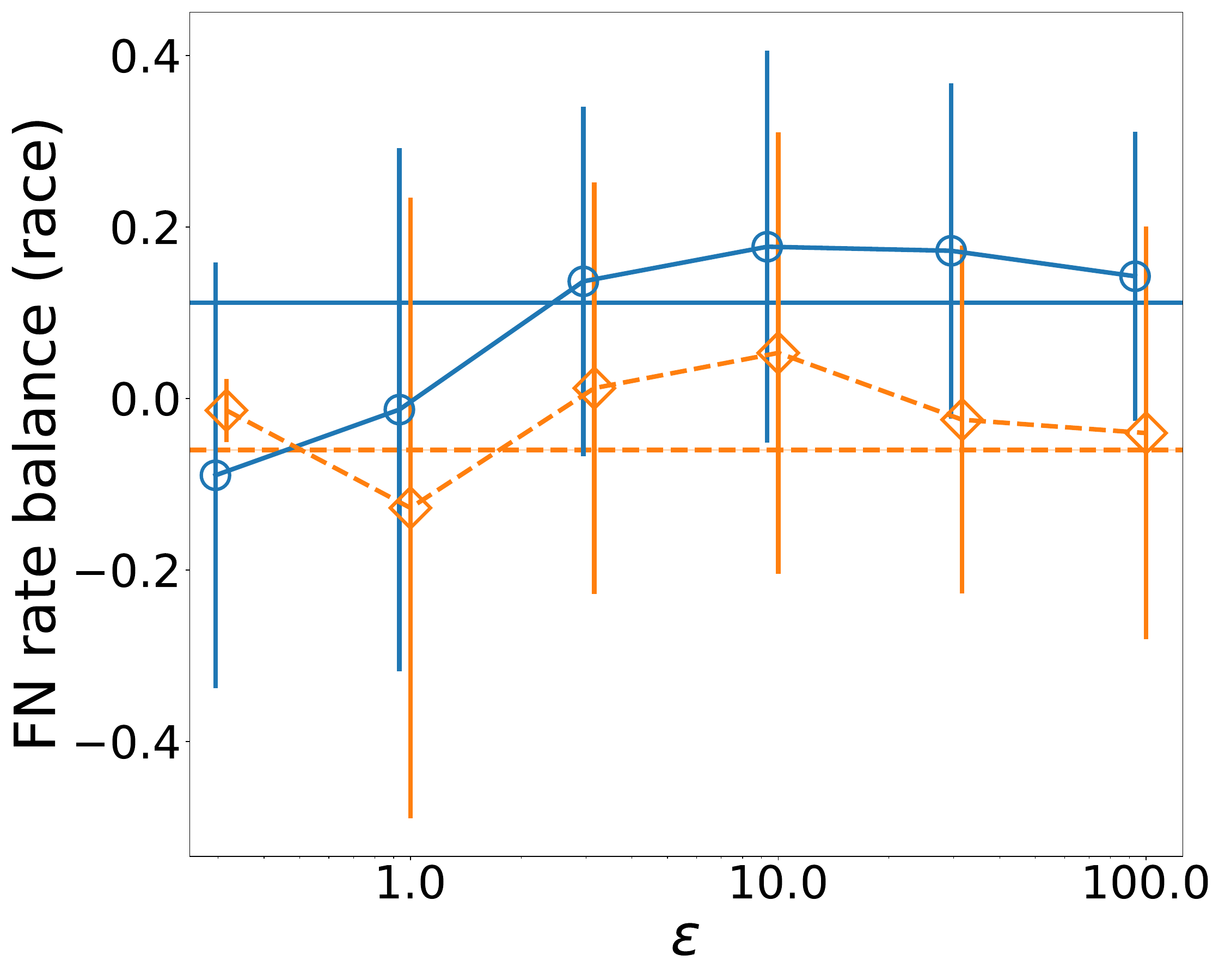}
    \includegraphics[width=0.325\linewidth]{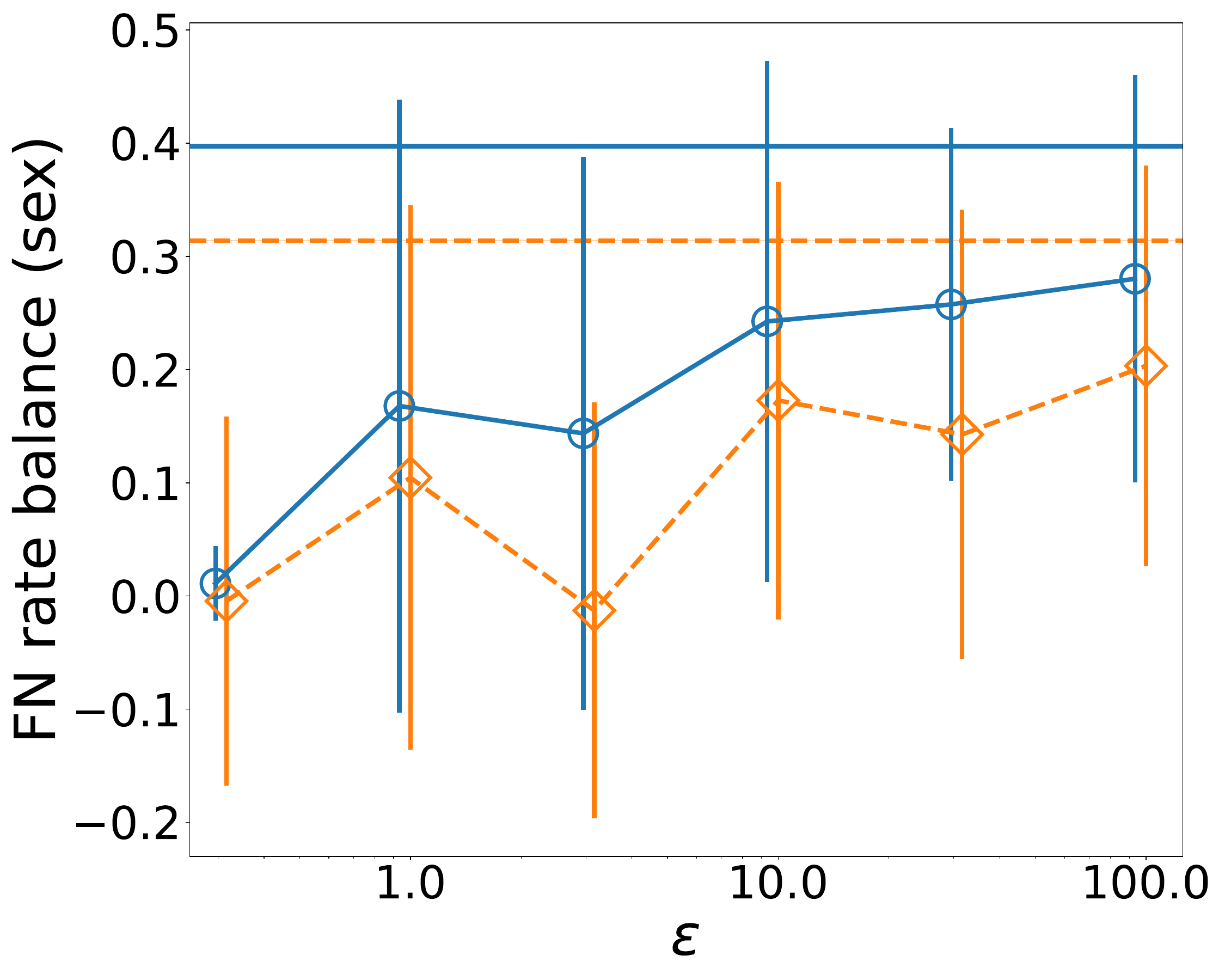}
    \includegraphics[width=0.325\linewidth]{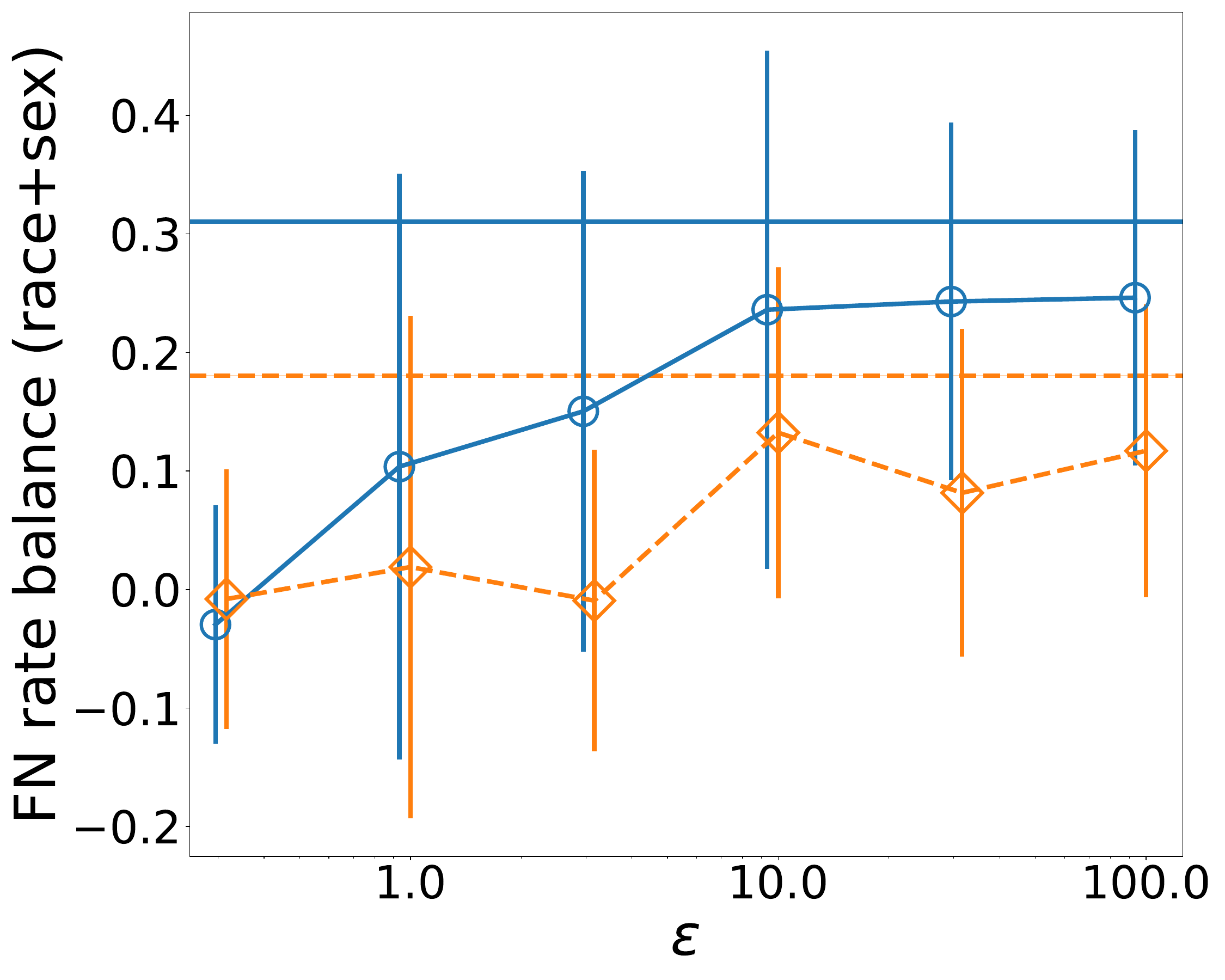}
    \caption{Mean $\pm$ 1 SD (error bars and shaded regions) FN rate balance, with race, sex, and race+sex as the protected attribute, for the DP-CTGAN + RW experiment.}
    \label{fig:Adult_cont_FNR_balance}
\end{figure}

\begin{figure}[!htb]
    \includegraphics[width=0.325\linewidth]{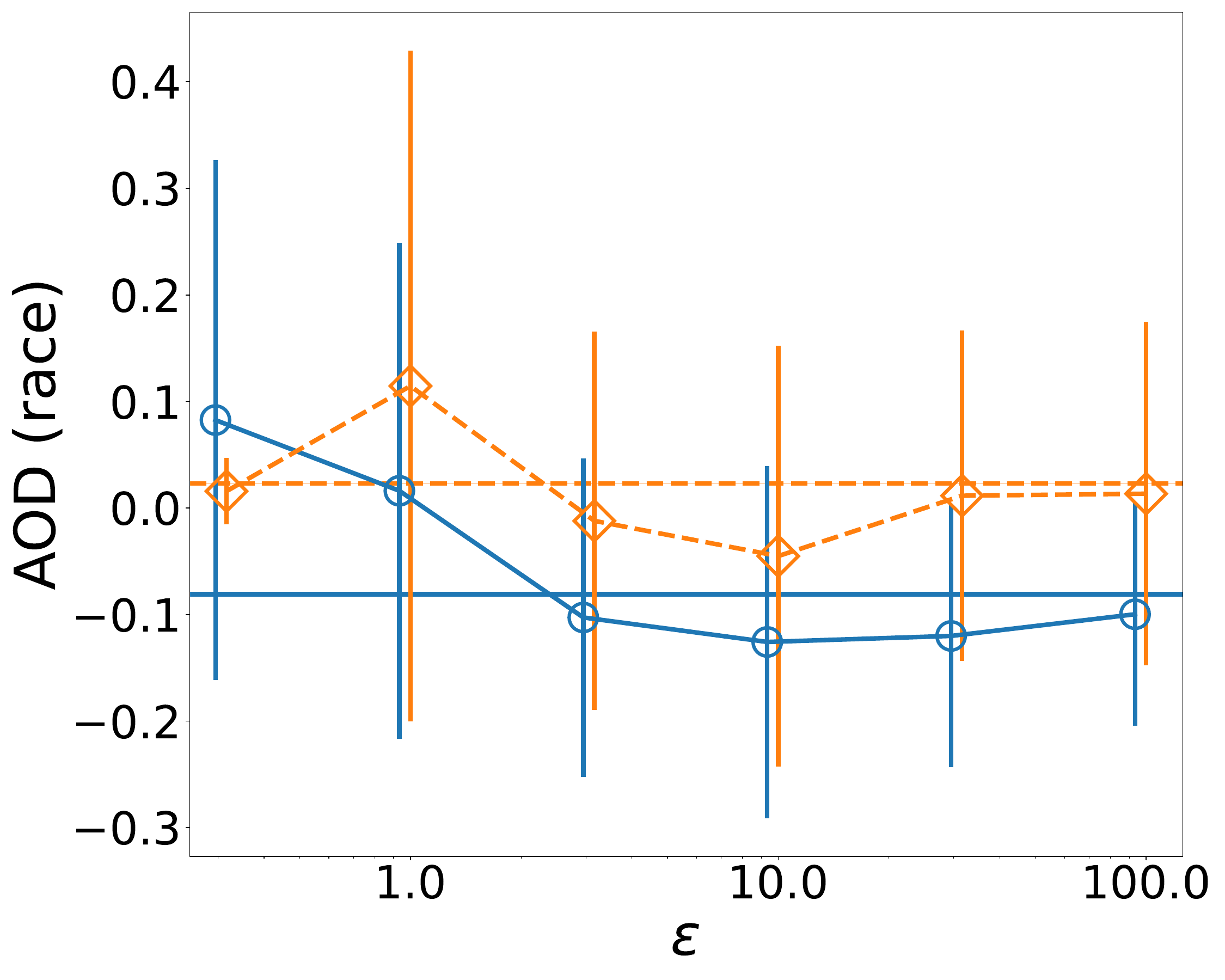}
    \includegraphics[width=0.325\linewidth]{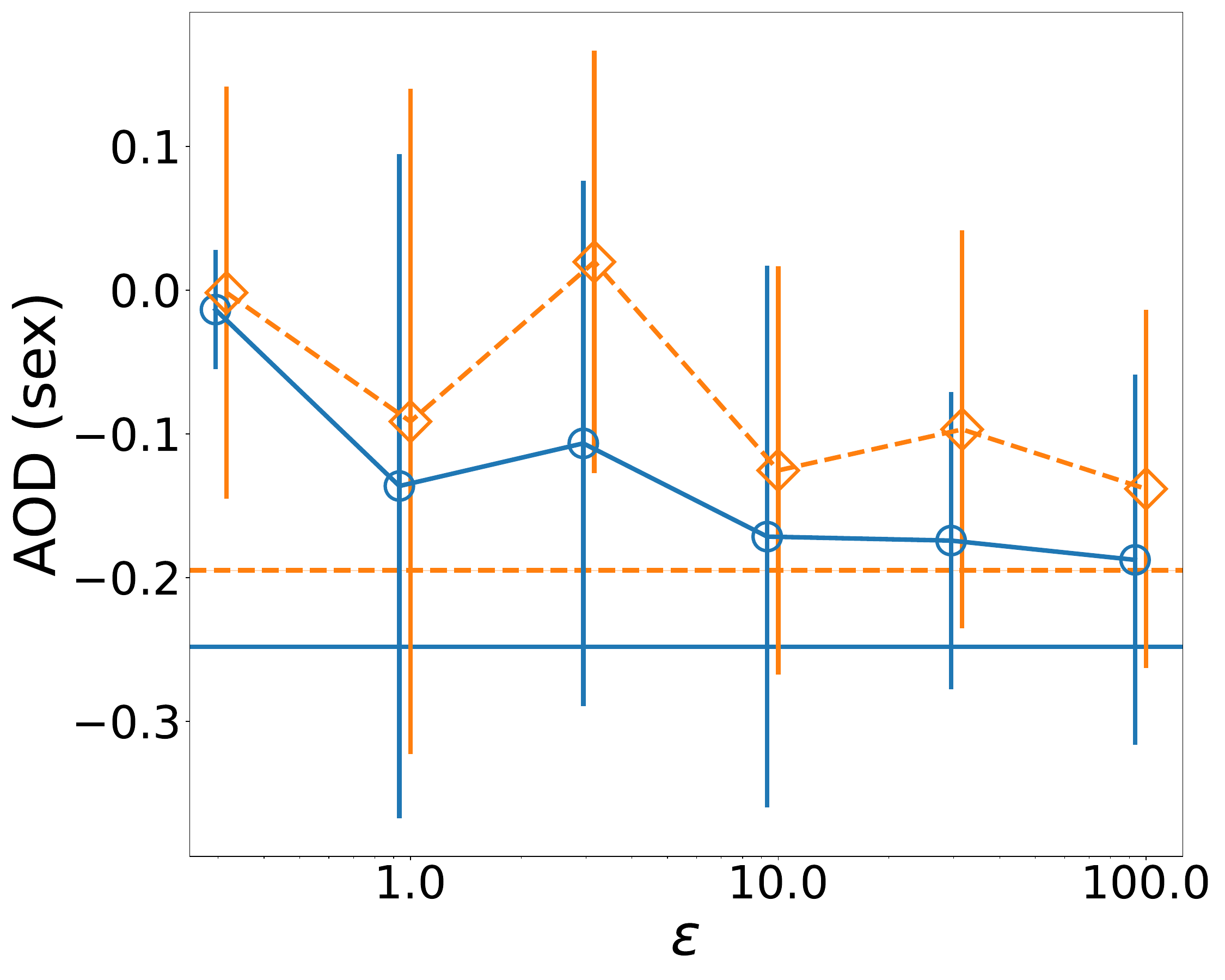}
    \includegraphics[width=0.325\linewidth]{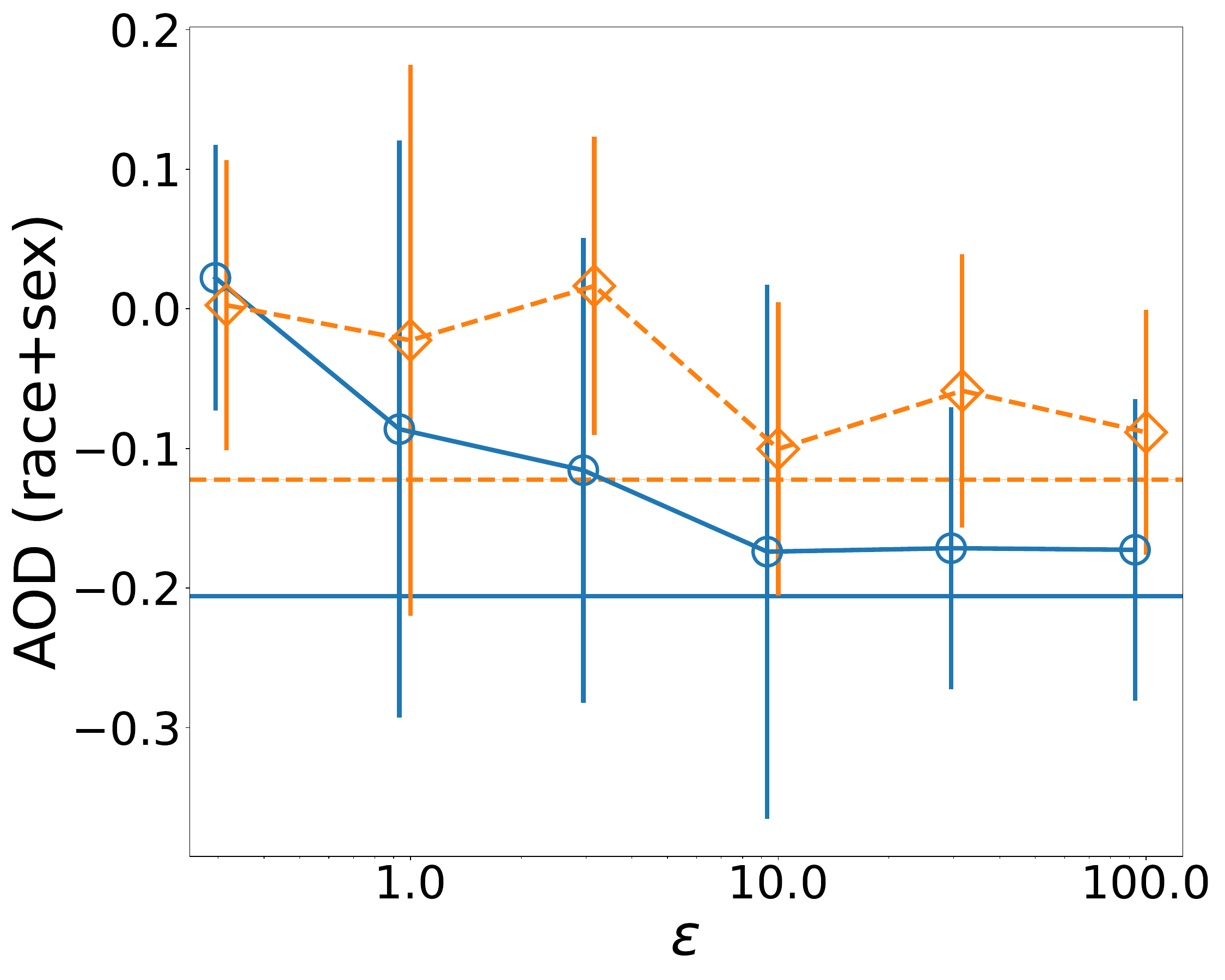}
    \caption{Mean $\pm$ 1 SD (error bars and shaded regions) AOD, with race, sex, and race+sex as the protected attribute, for the DP-CTGAN + RW experiment.}
    \label{fig:Adult_cont_AOD}
\end{figure}

We present several additional figures for the AIM + TOT experiments with the Adult (Figure  \ref{fig:Adult_TVD} through Figure  \ref{fig:Adult_AOD}) and COMPAS (Figure  \ref{fig:compas_TVD} through Figure  \ref{fig:compas_AOD}) datasets, as well as the DP-CTGAN + RW experiment on the Adult (with continuous age) dataset (Figure  \ref{fig:Adult_cont_TVD} through Figure  \ref{fig:Adult_cont_AOD}).

We note that the error bands for different values of $\eta$ in the fairness-only simulations can sometimes be fairly large, especially for the larger values of $\eta$ in our experiments.
Larger $\eta$ mean more flexibility in the permitted dataset distortions, and hence greater variability in the learned randomized fairness transformation.
Smaller $\eta$, on the other hand, would only permit very specific changes and thus have less variability.
This is most noticeable in the $\eta=0.1$ case of the right-hand plot of Figure  \ref{fig:Adult_KS}.
However, in that case, the entire error band of p-values falls in the ``fail-to-reject'' region of the plot, so they all essentially provide the same information.

\newpage

\newpage

\newpage

\newpage

\newpage

\newpage


\section{Tabular results}
We present the full set of tabular results for all privacy and utility metrics for all experiments.
Table \ref{tab:Adult_results_etaNone} shows the results of the AIM + TOT experiment on the Adult dataset for various $\varepsilon$ privacy parameters with no fairness transformation.
Tables \ref{tab:Adult_results_eta1} and \ref{tab:Adult_results_eta025} show the results of the AIM + TOT experiment on the Adult dataset for various $\varepsilon$ privacy parameters with a fairness transformation using $\eta=0.1$ and $\eta=0.025$, respectively.
Table \ref{tab:COMPAS_results_etaNone} shows the results of the AIM + TOT experiment on the COMPAS dataset for various $\varepsilon$ privacy parameters with no fairness transformation.
Tables \ref{tab:COMPAS_results_eta15} and \ref{tab:COMPAS_results_eta08} show the results of the AIM + TOT experiment on the COMPAS dataset for various $\varepsilon$ privacy parameters with a fairness transformation using $\eta=0.15$ and $\eta=0.08$, respectively.
Tables \ref{tab:Adult_continuous_results_no_fairness} and \ref{tab:Adult_continuous_results_with_fairness} contain the results of the DP-DP-CTGAN + RW experiment on the Adult dataset without and with the reweighting fairness data preprocessing, respectively.

\newpage
\begin{landscape}
\begin{table}
\caption{Full set of simulation results for the Adult dataset for various values of the privacy parameter $\varepsilon$ without the fairness transformation.
The other privacy parameter $\delta$ is assumed to be $\delta=10^{-9}$ throughout.
Values in the table are in the format ``mean (SD)'' of the output values from many runs of the evaluations.
At least 30 runs were performed in each case.
The final column represents a run that is also without the DP synthesis step as a baseline.
}
\label{tab:Adult_results_etaNone}
\resizebox{1.0\linewidth}{!}{
    \begin{tabular}{lrrrrrrrrr}
    \toprule
     & $\varepsilon$ & $10^{-2.0}$ & $10^{-1.5}$ & $10^{-1.0}$ & $10^{-0.5}$ & $10^{0.0}$ & $10^{0.5}$ & $10^{1.0}$ & None \\
    \midrule
    KS Stat &  & 0.344 (0.110) & 0.091 (0.027) & 0.030 (0.010) & 0.012 (0.005) & 0.006 (0.001) & 0.004 (0.001) & 0.003 (0.001) & 0.000 (0) \\
    \cline{1-10}
    KS p-val &  & 0.000 (0.000) & 0.000 (0.000) & 0.000 (0.000) & 0.058 (0.083) & 0.588 (0.211) & 0.906 (0.153) & 0.980 (0.060) & 1.000 (0) \\
    \cline{1-10}
    Cumulative 1-way TVD &  & 0.557 (0.094) & 0.164 (0.035) & 0.055 (0.011) & 0.022 (0.005) & 0.007 (0.001) & 0.003 (0.000) & 0.001 (0.000) & 0.000 (0) \\
    \cline{1-10}
    Cumulative 2-way TVD &  & 2.141 (0.311) & 0.775 (0.107) & 0.369 (0.058) & 0.133 (0.017) & 0.052 (0.007) & 0.025 (0.002) & 0.010 (0.001) & 0.000 (0) \\
    \cline{1-10}
    Cumulative 3-way TVD &  & 3.070 (0.428) & 1.272 (0.141) & 0.701 (0.110) & 0.267 (0.027) & 0.139 (0.013) & 0.100 (0.003) & 0.082 (0.001) & 0.000 (0) \\
    \cline{1-10}
    COD (race) &  & -0.025 (0.117) & -0.057 (0.076) & -0.065 (0.045) & -0.091 (0.028) & -0.099 (0.006) & -0.099 (0.002) & -0.099 (0.001) & -0.099 (0) \\
    \cline{1-10}
    COD (sex) &  & -0.107 (0.115) & -0.187 (0.070) & -0.194 (0.021) & -0.192 (0.009) & -0.193 (0.004) & -0.193 (0.001) & -0.193 (0.000) & -0.193 (0) \\
    \cline{1-10}
    COD (race + sex) &  & -0.093 (0.105) & -0.163 (0.065) & -0.172 (0.025) & -0.179 (0.014) & -0.184 (0.004) & -0.184 (0.001) & -0.184 (0.000) & -0.183 (0) \\
    \cline{1-10}
    Accuracy &  & 0.747 (0.025) & 0.770 (0.010) & 0.789 (0.007) & 0.795 (0.001) & 0.796 (0.001) & 0.796 (0.000) & 0.796 (0.000) & 0.796 (0) \\
    \cline{1-10}
    F1 score &  & 0.124 (0.173) & 0.304 (0.128) & 0.446 (0.063) & 0.463 (0.010) & 0.464 (0.006) & 0.464 (0.002) & 0.464 (0.002) & 0.463 (0) \\
    \cline{1-10}
    TP rate &  & 0.100 (0.151) & 0.222 (0.117) & 0.352 (0.073) & 0.359 (0.015) & 0.358 (0.009) & 0.359 (0.003) & 0.358 (0.002) & 0.356 (0) \\
    \cline{1-10}
    TN rate &  & 0.959 (0.058) & 0.949 (0.033) & 0.932 (0.023) & 0.938 (0.005) & 0.939 (0.003) & 0.939 (0.001) & 0.940 (0.001) & 0.940 (0) \\
    \cline{1-10}
    FN rate &  & 0.900 (0.151) & 0.778 (0.117) & 0.648 (0.073) & 0.641 (0.015) & 0.642 (0.009) & 0.641 (0.003) & 0.642 (0.002) & 0.644 (0) \\
    \cline{1-10}
    FP rate &  & 0.041 (0.058) & 0.051 (0.033) & 0.068 (0.023) & 0.062 (0.005) & 0.061 (0.003) & 0.061 (0.001) & 0.060 (0.001) & 0.060 (0) \\
    \cline{1-10}
    AUC &  & 0.607 (0.104) & 0.707 (0.076) & 0.791 (0.027) & 0.819 (0.002) & 0.823 (0.001) & 0.824 (0.000) & 0.824 (0.000) & 0.824 (0) \\
    \cline{1-10}
    SPD (race) &  & -0.010 (0.036) & -0.041 (0.059) & -0.071 (0.039) & -0.081 (0.022) & -0.077 (0.009) & -0.073 (0.003) & -0.072 (0.001) & -0.072 (0) \\
    \cline{1-10}
    SPD (sex) &  & -0.046 (0.085) & -0.119 (0.084) & -0.178 (0.051) & -0.189 (0.016) & -0.183 (0.008) & -0.182 (0.002) & -0.182 (0.002) & -0.180 (0) \\
    \cline{1-10}
    SPD (race + sex) &  & -0.038 (0.072) & -0.108 (0.082) & -0.165 (0.052) & -0.179 (0.016) & -0.171 (0.007) & -0.168 (0.003) & -0.168 (0.002) & -0.166 (0) \\
    \cline{1-10}
    AOD (race) &  & 0.003 (0.045) & -0.025 (0.093) & -0.041 (0.061) & -0.058 (0.043) & -0.044 (0.019) & -0.036 (0.005) & -0.035 (0.002) & -0.036 (0) \\
    \cline{1-10}
    AOD (sex) &  & -0.046 (0.099) & -0.141 (0.108) & -0.211 (0.067) & -0.239 (0.023) & -0.231 (0.012) & -0.229 (0.003) & -0.228 (0.002) & -0.226 (0) \\
    \cline{1-10}
    AOD (race + sex) &  & -0.030 (0.073) & -0.107 (0.097) & -0.162 (0.064) & -0.188 (0.024) & -0.175 (0.010) & -0.169 (0.003) & -0.169 (0.002) & -0.168 (0) \\
    \cline{1-10}
    FN rate balance (race) &  & -0.013 (0.064) & 0.028 (0.153) & 0.042 (0.099) & 0.070 (0.075) & 0.043 (0.034) & 0.028 (0.008) & 0.027 (0.003) & 0.029 (0) \\
    \cline{1-10}
    FN rate balance (sex) &  & 0.060 (0.142) & 0.215 (0.158) & 0.330 (0.098) & 0.386 (0.036) & 0.375 (0.018) & 0.372 (0.004) & 0.371 (0.003) & 0.368 (0) \\
    \cline{1-10}
    FN rate balance (race + sex) &  & 0.034 (0.096) & 0.153 (0.140) & 0.233 (0.090) & 0.284 (0.037) & 0.262 (0.016) & 0.252 (0.005) & 0.251 (0.003) & 0.250 (0) \\
    \cline{1-10}
    FP rate balance (race) &  & -0.007 (0.030) & -0.021 (0.037) & -0.040 (0.027) & -0.045 (0.012) & -0.045 (0.005) & -0.043 (0.002) & -0.043 (0.001) & -0.042 (0) \\
    \cline{1-10}
    FP rate balance (sex) &  & -0.031 (0.061) & -0.067 (0.060) & -0.092 (0.038) & -0.092 (0.012) & -0.087 (0.006) & -0.086 (0.002) & -0.086 (0.001) & -0.084 (0) \\
    \cline{1-10}
    FP rate balance (race + sex) &  & -0.027 (0.054) & -0.062 (0.059) & -0.090 (0.041) & -0.093 (0.013) & -0.088 (0.006) & -0.086 (0.002) & -0.086 (0.002) & -0.085 (0) \\
    \bottomrule
    \end{tabular}}
\end{table}
\end{landscape}

\newpage
\begin{landscape}
\begin{table}
\caption{Full set of simulation results for the Adult dataset for $\eta=0.1$ and for various values of the privacy parameter $\varepsilon$.
The other privacy parameter $\delta$ is assumed to be $\delta=10^{-9}$ throughout.
Values in the table are in the format ``mean (standard deviation)'' of the output values from many runs of the evaluations.
At least 30 runs were performed in each case.
The final column represents a run without the DP synthesis step as a baseline.
}
\label{tab:Adult_results_eta1}
\resizebox{1.0\linewidth}{!}{
\begin{tabular}{lrrrrrrrrr}
    \toprule
     & $\varepsilon$ & $10^{-2.0}$ & $10^{-1.5}$ & $10^{-1.0}$ & $10^{-0.5}$ & $10^{0.0}$ & $10^{0.5}$ & $10^{1.0}$ & None \\
    \midrule
    KS Stat &  & 0.327 (0.119) & 0.091 (0.031) & 0.034 (0.009) & 0.015 (0.005) & 0.010 (0.004) & 0.007 (0.002) & 0.006 (0.003) & 0.004 (0.003) \\
    \cline{1-10}
    KS p-val &  & 0.000 (0.000) & 0.000 (0.000) & 0.000 (0.000) & 0.026 (0.051) & 0.183 (0.286) & 0.423 (0.304) & 0.518 (0.372) & 0.795 (0.321) \\
    \cline{1-10}
    Cumulative 1-way TVD &  & 0.555 (0.119) & 0.170 (0.040) & 0.054 (0.013) & 0.022 (0.006) & 0.008 (0.002) & 0.004 (0.001) & 0.003 (0.001) & 0.002 (0.001) \\
    \cline{1-10}
    Cumulative 2-way TVD &  & 2.119 (0.411) & 0.813 (0.106) & 0.405 (0.052) & 0.177 (0.016) & 0.109 (0.007) & 0.086 (0.003) & 0.075 (0.003) & 0.070 (0.002) \\
    \cline{1-10}
    Cumulative 3-way TVD &  & 3.031 (0.555) & 1.320 (0.141) & 0.773 (0.106) & 0.367 (0.023) & 0.271 (0.010) & 0.240 (0.006) & 0.225 (0.005) & 0.180 (0.004) \\
    \cline{1-10}
    COD (race) &  & -0.015 (0.038) & -0.025 (0.029) & -0.029 (0.022) & -0.047 (0.014) & -0.046 (0.004) & -0.048 (0.004) & -0.047 (0.003) & -0.046 (0.003) \\
    \cline{1-10}
    COD (sex) &  & -0.033 (0.053) & -0.087 (0.028) & -0.096 (0.006) & -0.093 (0.003) & -0.093 (0.003) & -0.093 (0.003) & -0.093 (0.003) & -0.093 (0.002) \\
    \cline{1-10}
    COD (race + sex) &  & -0.030 (0.044) & -0.077 (0.022) & -0.086 (0.008) & -0.093 (0.005) & -0.092 (0.002) & -0.093 (0.002) & -0.093 (0.003) & -0.092 (0.002) \\
    \cline{1-10}
    Accuracy &  & 0.746 (0.017) & 0.768 (0.013) & 0.785 (0.009) & 0.794 (0.001) & 0.794 (0.001) & 0.794 (0.001) & 0.794 (0.000) & 0.794 (0.000) \\
    \cline{1-10}
    F1 score &  & 0.131 (0.171) & 0.315 (0.099) & 0.407 (0.077) & 0.468 (0.010) & 0.471 (0.003) & 0.471 (0.002) & 0.471 (0.001) & 0.471 (0.001) \\
    \cline{1-10}
    TP rate &  & 0.109 (0.156) & 0.226 (0.090) & 0.308 (0.083) & 0.368 (0.014) & 0.373 (0.005) & 0.372 (0.003) & 0.372 (0.002) & 0.372 (0.002) \\
    \cline{1-10}
    TN rate &  & 0.954 (0.060) & 0.946 (0.031) & 0.940 (0.019) & 0.933 (0.005) & 0.932 (0.002) & 0.932 (0.001) & 0.932 (0.001) & 0.932 (0.001) \\
    \cline{1-10}
    FN rate &  & 0.891 (0.156) & 0.774 (0.090) & 0.692 (0.083) & 0.632 (0.014) & 0.627 (0.005) & 0.628 (0.003) & 0.628 (0.002) & 0.628 (0.002) \\
    \cline{1-10}
    FP rate &  & 0.046 (0.060) & 0.054 (0.031) & 0.060 (0.019) & 0.067 (0.005) & 0.068 (0.002) & 0.068 (0.001) & 0.068 (0.001) & 0.068 (0.001) \\
    \cline{1-10}
    AUC &  & 0.613 (0.088) & 0.717 (0.026) & 0.777 (0.030) & 0.811 (0.003) & 0.813 (0.002) & 0.813 (0.001) & 0.814 (0.001) & 0.813 (0.001) \\
    \cline{1-10}
    SPD (race) &  & -0.018 (0.040) & -0.030 (0.027) & -0.044 (0.022) & -0.068 (0.017) & -0.057 (0.010) & -0.060 (0.009) & -0.058 (0.009) & -0.055 (0.008) \\
    \cline{1-10}
    SPD (sex) &  & -0.017 (0.053) & -0.085 (0.072) & -0.099 (0.041) & -0.129 (0.009) & -0.127 (0.004) & -0.128 (0.003) & -0.128 (0.002) & -0.126 (0.002) \\
    \cline{1-10}
    SPD (race + sex) &  & -0.019 (0.047) & -0.075 (0.063) & -0.090 (0.038) & -0.123 (0.010) & -0.117 (0.005) & -0.119 (0.004) & -0.117 (0.003) & -0.116 (0.003) \\
    \cline{1-10}
    AOD (race) &  & -0.009 (0.050) & -0.005 (0.039) & -0.005 (0.030) & -0.031 (0.033) & -0.011 (0.016) & -0.015 (0.014) & -0.011 (0.015) & -0.009 (0.013) \\
    \cline{1-10}
    AOD (sex) &  & -0.008 (0.058) & -0.084 (0.107) & -0.087 (0.058) & -0.122 (0.014) & -0.116 (0.005) & -0.118 (0.005) & -0.118 (0.003) & -0.116 (0.003) \\
    \cline{1-10}
    AOD (race + sex) &  & -0.009 (0.046) & -0.058 (0.081) & -0.058 (0.044) & -0.093 (0.017) & -0.080 (0.007) & -0.083 (0.007) & -0.081 (0.006) & -0.080 (0.005) \\
    \cline{1-10}
    FN rate balance (race) &  & 0.007 (0.071) & -0.004 (0.063) & -0.013 (0.050) & 0.025 (0.057) & -0.009 (0.024) & -0.003 (0.021) & -0.010 (0.023) & -0.012 (0.021) \\
    \cline{1-10}
    FN rate balance (sex) &  & 0.013 (0.079) & 0.129 (0.166) & 0.136 (0.093) & 0.193 (0.022) & 0.183 (0.007) & 0.187 (0.009) & 0.186 (0.004) & 0.184 (0.004) \\
    \cline{1-10}
    FN rate balance (race + sex) &  & 0.011 (0.060) & 0.079 (0.119) & 0.079 (0.066) & 0.131 (0.028) & 0.109 (0.011) & 0.114 (0.011) & 0.111 (0.009) & 0.110 (0.008) \\
    \cline{1-10}
    FP rate balance (race) &  & -0.012 (0.031) & -0.015 (0.018) & -0.022 (0.014) & -0.037 (0.010) & -0.031 (0.007) & -0.033 (0.007) & -0.032 (0.007) & -0.029 (0.006) \\
    \cline{1-10}
    FP rate balance (sex) &  & -0.004 (0.040) & -0.040 (0.049) & -0.037 (0.024) & -0.052 (0.007) & -0.049 (0.003) & -0.050 (0.003) & -0.050 (0.002) & -0.049 (0.002) \\
    \cline{1-10}
    FP rate balance (race + sex) &  & -0.006 (0.035) & -0.036 (0.044) & -0.037 (0.023) & -0.055 (0.007) & -0.051 (0.003) & -0.052 (0.003) & -0.051 (0.003) & -0.050 (0.002) \\
    \bottomrule
    \end{tabular}}
\end{table}
\end{landscape}

\newpage
\begin{landscape}
\begin{table}
\caption{Full set of simulation results for the Adult dataset for $\eta=0.025$ and for various values of the privacy parameter $\varepsilon$.
The other privacy parameter $\delta$ is assumed to be $\delta=10^{-9}$ throughout.
Values in the table are in the format ``mean (standard deviation)'' of the output values from many runs of the evaluations.
At least 30 runs were performed in each case.
The final column represents a run without the DP synthesis step as a baseline.
}
\label{tab:Adult_results_eta025}
\resizebox{1.0\linewidth}{!}{
    \begin{tabular}{lrrrrrrrrr}
    \toprule
     & $\varepsilon$ & $10^{-2.0}$ & $10^{-1.5}$ & $10^{-1.0}$ & $10^{-0.5}$ & $10^{0.0}$ & $10^{0.5}$ & $10^{1.0}$ & None \\
    \midrule
    KS Stat &  & 0.350 (0.128) & 0.086 (0.025) & 0.047 (0.013) & 0.029 (0.004) & 0.029 (0.003) & 0.028 (0.002) & 0.028 (0.002) & 0.027 (0.001) \\
    \cline{1-10}
    KS p-val &  & 0.000 (0.000) & 0.000 (0.000) & 0.000 (0.000) & 0.000 (0.000) & 0.000 (0.000) & 0.000 (0.000) & 0.000 (0.000) & 0.000 (0.000) \\
    \cline{1-10}
    Cumulative 1-way TVD &  & 0.593 (0.120) & 0.183 (0.036) & 0.086 (0.014) & 0.056 (0.005) & 0.046 (0.003) & 0.045 (0.002) & 0.044 (0.002) & 0.042 (0.001) \\
    \cline{1-10}
    Cumulative 2-way TVD &  & 2.252 (0.429) & 0.825 (0.101) & 0.478 (0.056) & 0.263 (0.013) & 0.222 (0.009) & 0.213 (0.006) & 0.209 (0.006) & 0.202 (0.004) \\
    \cline{1-10}
    Cumulative 3-way TVD &  & 3.204 (0.583) & 1.348 (0.128) & 0.871 (0.099) & 0.483 (0.016) & 0.418 (0.012) & 0.398 (0.007) & 0.392 (0.006) & 0.365 (0.006) \\
    \cline{1-10}
    COD (race) &  & -0.003 (0.013) & -0.004 (0.008) & -0.011 (0.009) & -0.018 (0.004) & -0.017 (0.005) & -0.018 (0.004) & -0.018 (0.003) & -0.018 (0.003) \\
    \cline{1-10}
    COD (sex) &  & -0.015 (0.013) & -0.024 (0.003) & -0.023 (0.003) & -0.022 (0.003) & -0.022 (0.002) & -0.022 (0.003) & -0.022 (0.003) & -0.022 (0.003) \\
    \cline{1-10}
    COD (race + sex) &  & -0.013 (0.012) & -0.021 (0.004) & -0.023 (0.004) & -0.025 (0.003) & -0.025 (0.002) & -0.025 (0.003) & -0.025 (0.003) & -0.025 (0.003) \\
    \cline{1-10}
    Accuracy &  & 0.731 (0.032) & 0.761 (0.014) & 0.776 (0.006) & 0.783 (0.003) & 0.785 (0.002) & 0.786 (0.001) & 0.786 (0.001) & 0.786 (0.000) \\
    \cline{1-10}
    F1 score &  & 0.162 (0.158) & 0.329 (0.046) & 0.408 (0.070) & 0.474 (0.018) & 0.482 (0.003) & 0.484 (0.002) & 0.483 (0.002) & 0.483 (0.001) \\
    \cline{1-10}
    TP rate &  & 0.134 (0.141) & 0.241 (0.054) & 0.322 (0.084) & 0.397 (0.024) & 0.407 (0.005) & 0.407 (0.002) & 0.405 (0.003) & 0.405 (0.001) \\
    \cline{1-10}
    TN rate &  & 0.926 (0.073) & 0.931 (0.028) & 0.925 (0.024) & 0.910 (0.007) & 0.908 (0.003) & 0.910 (0.001) & 0.911 (0.001) & 0.911 (0.001) \\
    \cline{1-10}
    FN rate &  & 0.866 (0.141) & 0.759 (0.054) & 0.678 (0.084) & 0.603 (0.024) & 0.593 (0.005) & 0.593 (0.002) & 0.595 (0.003) & 0.595 (0.001) \\
    \cline{1-10}
    FP rate &  & 0.074 (0.073) & 0.069 (0.028) & 0.075 (0.024) & 0.090 (0.007) & 0.092 (0.003) & 0.090 (0.001) & 0.089 (0.001) & 0.089 (0.001) \\
    \cline{1-10}
    AUC &  & 0.598 (0.087) & 0.694 (0.028) & 0.758 (0.029) & 0.789 (0.004) & 0.788 (0.002) & 0.790 (0.002) & 0.790 (0.002) & 0.789 (0.001) \\
    \cline{1-10}
    SPD (race) &  & -0.018 (0.029) & -0.031 (0.021) & -0.037 (0.022) & -0.064 (0.017) & -0.047 (0.010) & -0.043 (0.006) & -0.042 (0.005) & -0.040 (0.001) \\
    \cline{1-10}
    SPD (sex) &  & -0.007 (0.032) & -0.029 (0.032) & -0.040 (0.023) & -0.060 (0.011) & -0.061 (0.004) & -0.064 (0.002) & -0.063 (0.002) & -0.063 (0.001) \\
    \cline{1-10}
    SPD (race + sex) &  & -0.010 (0.030) & -0.030 (0.031) & -0.042 (0.024) & -0.066 (0.012) & -0.063 (0.004) & -0.063 (0.003) & -0.062 (0.002) & -0.061 (0.001) \\
    \cline{1-10}
    AOD (race) &  & -0.009 (0.029) & -0.004 (0.025) & 0.007 (0.033) & -0.024 (0.033) & 0.010 (0.023) & 0.019 (0.014) & 0.020 (0.012) & 0.024 (0.001) \\
    \cline{1-10}
    AOD (sex) &  & 0.010 (0.046) & 0.009 (0.035) & 0.015 (0.029) & 0.003 (0.014) & 0.002 (0.004) & -0.000 (0.002) & 0.000 (0.002) & 0.000 (0.002) \\
    \cline{1-10}
    AOD (race + sex) &  & 0.004 (0.037) & 0.009 (0.031) & 0.014 (0.027) & -0.006 (0.017) & 0.004 (0.009) & 0.007 (0.006) & 0.008 (0.005) & 0.009 (0.001) \\
    \cline{1-10}
    FN rate balance (race) &  & 0.005 (0.039) & -0.009 (0.038) & -0.032 (0.057) & 0.016 (0.057) & -0.043 (0.042) & -0.060 (0.024) & -0.062 (0.022) & -0.068 (0.001) \\
    \cline{1-10}
    FN rate balance (sex) &  & -0.016 (0.064) & -0.014 (0.047) & -0.024 (0.042) & -0.004 (0.023) & -0.003 (0.006) & -0.000 (0.003) & -0.001 (0.003) & -0.000 (0.002) \\
    \cline{1-10}
    FN rate balance (race + sex) &  & -0.009 (0.050) & -0.018 (0.039) & -0.029 (0.040) & 0.002 (0.028) & -0.016 (0.016) & -0.022 (0.010) & -0.024 (0.009) & -0.026 (0.002) \\
    \cline{1-10}
    FP rate balance (race) &  & -0.012 (0.023) & -0.016 (0.017) & -0.018 (0.014) & -0.033 (0.010) & -0.023 (0.005) & -0.022 (0.003) & -0.021 (0.002) & -0.020 (0.001) \\
    \cline{1-10}
    FP rate balance (sex) &  & 0.003 (0.029) & 0.004 (0.024) & 0.007 (0.016) & 0.001 (0.006) & 0.001 (0.003) & -0.001 (0.002) & -0.000 (0.001) & -0.000 (0.002) \\
    \cline{1-10}
    FP rate balance (race + sex) &  & -0.001 (0.026) & -0.001 (0.025) & -0.000 (0.015) & -0.010 (0.006) & -0.007 (0.002) & -0.008 (0.002) & -0.008 (0.001) & -0.007 (0.001) \\
    \bottomrule
    \end{tabular}}
\end{table}
\end{landscape}

\begin{landscape}
\begin{table}
\caption{Full set of simulation results for the COMPAS dataset for various values of the privacy parameter $\varepsilon$ without the fairness transformation.
The other privacy parameter $\delta$ is assumed to be $\delta=10^{-9}$ throughout.
Values in the table are in the format ``mean (standard deviation)'' of the output values from many runs of the evaluations.
At least 30 successful runs were performed in each case.
The final column represents a run that is also without the DP synthesis step as a baseline.
}
\label{tab:COMPAS_results_etaNone}
\resizebox{1.0\linewidth}{!}{
    \begin{tabular}{lrrrrrrrrr}
    \toprule
     & $\varepsilon$ & $10^{-2.0}$ & $10^{-1.5}$ & $10^{-1.0}$ & $10^{-0.5}$ & $10^{0.0}$ & $10^{0.5}$ & $10^{1.0}$ & None \\
    \midrule
    KS Stat &  & 0.553 (0.152) & 0.261 (0.077) & 0.104 (0.031) & 0.054 (0.016) & 0.018 (0.006) & 0.011 (0.003) & 0.009 (0.002) & 0.000 (0.000) \\
    \cline{1-10}
    KS p-val &  & 0.000 (0.000) & 0.000 (0.000) & 0.000 (0.000) & 0.010 (0.027) & 0.580 (0.306) & 0.924 (0.134) & 0.979 (0.041) & 1.000 (0.000) \\
    \cline{1-10}
    Cumulative 1-way TVD &  & 1.362 (0.385) & 0.520 (0.151) & 0.193 (0.053) & 0.064 (0.019) & 0.018 (0.005) & 0.007 (0.002) & 0.003 (0.001) & 0.000 (0.000) \\
    \cline{1-10}
    Cumulative 2-way TVD &  & 5.262 (1.167) & 2.308 (0.572) & 1.039 (0.171) & 0.515 (0.090) & 0.175 (0.034) & 0.066 (0.012) & 0.023 (0.003) & 0.000 (0.000) \\
    \cline{1-10}
    Cumulative 3-way TVD &  & 8.800 (1.609) & 4.349 (0.968) & 2.227 (0.292) & 1.236 (0.218) & 0.476 (0.077) & 0.266 (0.022) & 0.210 (0.004) & 0.000 (0.000) \\
    \cline{1-10}
    COD (race) &  & -0.017 (0.149) & -0.061 (0.145) & -0.029 (0.071) & -0.085 (0.053) & -0.114 (0.022) & -0.126 (0.010) & -0.130 (0.002) & -0.129 (0.000) \\
    \cline{1-10}
    COD (sex) &  & 0.002 (0.279) & 0.043 (0.279) & -0.055 (0.095) & -0.053 (0.076) & -0.127 (0.042) & -0.139 (0.014) & -0.141 (0.006) & -0.141 (0.000) \\
    \cline{1-10}
    COD (race + sex) &  & -0.018 (0.222) & 0.013 (0.277) & -0.067 (0.093) & -0.100 (0.072) & -0.178 (0.043) & -0.189 (0.020) & -0.191 (0.006) & -0.136 (0.000) \\
    \cline{1-10}
    Accuracy &  & 0.512 (0.055) & 0.517 (0.058) & 0.581 (0.067) & 0.649 (0.019) & 0.670 (0.007) & 0.675 (0.002) & 0.675 (0.000) & 0.675 (0.000) \\
    \cline{1-10}
    F1 score &  & 0.505 (0.261) & 0.532 (0.208) & 0.581 (0.169) & 0.681 (0.033) & 0.699 (0.014) & 0.709 (0.004) & 0.707 (0.001) & 0.708 (0.000) \\
    \cline{1-10}
    TP rate &  & 0.618 (0.383) & 0.602 (0.302) & 0.589 (0.239) & 0.682 (0.079) & 0.694 (0.033) & 0.714 (0.010) & 0.706 (0.002) & 0.709 (0.000) \\
    \cline{1-10}
    TN rate &  & 0.380 (0.396) & 0.413 (0.313) & 0.571 (0.222) & 0.608 (0.087) & 0.641 (0.028) & 0.627 (0.010) & 0.636 (0.003) & 0.633 (0.000) \\
    \cline{1-10}
    FN rate &  & 0.382 (0.383) & 0.398 (0.302) & 0.411 (0.239) & 0.318 (0.079) & 0.306 (0.033) & 0.286 (0.010) & 0.294 (0.002) & 0.291 (0.000) \\
    \cline{1-10}
    FP rate &  & 0.620 (0.396) & 0.587 (0.313) & 0.429 (0.222) & 0.392 (0.087) & 0.359 (0.028) & 0.373 (0.010) & 0.364 (0.003) & 0.367 (0.000) \\
    \cline{1-10}
    AUC &  & 0.488 (0.071) & 0.517 (0.067) & 0.601 (0.083) & 0.692 (0.023) & 0.716 (0.003) & 0.719 (0.001) & 0.720 (0.001) & 0.720 (0.000) \\
    \cline{1-10}
    SPD (race) &  & 0.015 (0.207) & -0.066 (0.197) & -0.218 (0.265) & -0.268 (0.134) & -0.270 (0.059) & -0.270 (0.027) & -0.284 (0.003) & -0.280 (0.000) \\
    \cline{1-10}
    SPD (sex) &  & 0.018 (0.191) & 0.063 (0.336) & -0.191 (0.264) & -0.210 (0.136) & -0.264 (0.057) & -0.247 (0.026) & -0.251 (0.012) & -0.268 (0.000) \\
    \cline{1-10}
    SPD (race + sex) &  & 0.023 (0.196) & 0.037 (0.324) & -0.234 (0.265) & -0.271 (0.118) & -0.288 (0.048) & -0.278 (0.024) & -0.286 (0.007) & -0.287 (0.000) \\
    \cline{1-10}
    AOD (race) &  & 0.016 (0.203) & -0.065 (0.199) & -0.198 (0.268) & -0.231 (0.137) & -0.229 (0.059) & -0.228 (0.028) & -0.241 (0.004) & -0.238 (0.000) \\
    \cline{1-10}
    AOD (sex) &  & 0.018 (0.197) & 0.063 (0.335) & -0.176 (0.272) & -0.180 (0.142) & -0.230 (0.059) & -0.214 (0.027) & -0.218 (0.013) & -0.236 (0.000) \\
    \cline{1-10}
    AOD (race + sex) &  & 0.025 (0.197) & 0.036 (0.326) & -0.215 (0.274) & -0.233 (0.127) & -0.250 (0.050) & -0.241 (0.027) & -0.250 (0.007) & -0.251 (0.000) \\
    \cline{1-10}
    FN rate balance (race) &  & -0.013 (0.205) & 0.069 (0.193) & 0.200 (0.259) & 0.231 (0.135) & 0.215 (0.067) & 0.215 (0.029) & 0.232 (0.003) & 0.227 (0.000) \\
    \cline{1-10}
    FN rate balance (sex) &  & -0.017 (0.182) & -0.073 (0.341) & 0.166 (0.244) & 0.175 (0.125) & 0.217 (0.056) & 0.195 (0.023) & 0.200 (0.009) & 0.213 (0.000) \\
    \cline{1-10}
    FN rate balance (race + sex) &  & -0.017 (0.186) & -0.045 (0.318) & 0.223 (0.241) & 0.257 (0.103) & 0.258 (0.046) & 0.241 (0.018) & 0.249 (0.005) & 0.249 (0.000) \\
    \cline{1-10}
    FP rate balance (race) &  & 0.019 (0.204) & -0.060 (0.215) & -0.196 (0.281) & -0.231 (0.142) & -0.242 (0.051) & -0.241 (0.027) & -0.251 (0.004) & -0.248 (0.000) \\
    \cline{1-10}
    FP rate balance (sex) &  & 0.018 (0.218) & 0.053 (0.336) & -0.187 (0.308) & -0.184 (0.163) & -0.244 (0.065) & -0.233 (0.032) & -0.235 (0.017) & -0.259 (0.000) \\
    \cline{1-10}
    FP rate balance (race + sex) &  & 0.032 (0.221) & 0.027 (0.346) & -0.208 (0.318) & -0.208 (0.160) & -0.243 (0.059) & -0.241 (0.036) & -0.250 (0.009) & -0.252 (0.000) \\
    \bottomrule
    \end{tabular}}
\end{table}
\end{landscape}

\newpage
\begin{landscape}
\begin{table}
\caption{Full set of simulation results for the COMPAS dataset for $\eta=0.15$ and for various values of the privacy parameter $\varepsilon$.
The other privacy parameter $\delta$ is assumed to be $\delta=10^{-9}$ throughout.
Values in the table are in the format ``mean (standard deviation)'' of the output values from many runs of the evaluations.
At least 30 successful runs were performed in each case, except for $\varepsilon=10^{-2}$, which had 23.
The final column represents a run without the DP synthesis step as a baseline.
}
\label{tab:COMPAS_results_eta15}
\resizebox{1.0\linewidth}{!}{
\begin{tabular}{lrrrrrrrrr}
    \toprule
     & $\varepsilon$ & $10^{-2.0}$ & $10^{-1.5}$ & $10^{-1.0}$ & $10^{-0.5}$ & $10^{0.0}$ & $10^{0.5}$ & $10^{1.0}$ & None \\
    \midrule
    KS Stat &  & 0.531 (0.136) & 0.244 (0.081) & 0.103 (0.029) & 0.050 (0.012) & 0.022 (0.006) & 0.013 (0.003) & 0.012 (0.003) & 0.007 (0.002) \\
    \cline{1-10}
    KS p-val &  & 0.000 (0.000) & 0.000 (0.000) & 0.000 (0.000) & 0.009 (0.029) & 0.380 (0.273) & 0.845 (0.196) & 0.870 (0.146) & 0.997 (0.009) \\
    \cline{1-10}
    Cumulative 1-way TVD &  & 1.345 (0.296) & 0.511 (0.176) & 0.167 (0.042) & 0.066 (0.021) & 0.023 (0.006) & 0.012 (0.003) & 0.011 (0.003) & 0.008 (0.002) \\
    \cline{1-10}
    Cumulative 2-way TVD &  & 5.180 (1.036) & 2.292 (0.705) & 1.010 (0.151) & 0.528 (0.060) & 0.210 (0.034) & 0.111 (0.014) & 0.084 (0.009) & 0.071 (0.006) \\
    \cline{1-10}
    Cumulative 3-way TVD &  & 8.643 (1.535) & 4.341 (1.157) & 2.215 (0.267) & 1.265 (0.145) & 0.539 (0.077) & 0.326 (0.033) & 0.289 (0.019) & 0.182 (0.013) \\
    \cline{1-10}
    COD (race) &  & -0.004 (0.063) & -0.001 (0.043) & -0.031 (0.055) & -0.080 (0.046) & -0.090 (0.020) & -0.100 (0.006) & -0.098 (0.005) & -0.112 (0.004) \\
    \cline{1-10}
    COD (sex) &  & 0.002 (0.019) & -0.004 (0.052) & -0.012 (0.045) & -0.045 (0.056) & -0.087 (0.034) & -0.092 (0.008) & -0.095 (0.008) & -0.095 (0.007) \\
    \cline{1-10}
    COD (race + sex) &  & 0.013 (0.045) & -0.003 (0.060) & -0.027 (0.054) & -0.074 (0.044) & -0.098 (0.023) & -0.091 (0.015) & -0.092 (0.014) & -0.081 (0.010) \\
    \cline{1-10}
    Accuracy &  & 0.475 (0.060) & 0.537 (0.061) & 0.584 (0.061) & 0.648 (0.018) & 0.670 (0.007) & 0.677 (0.004) & 0.677 (0.004) & 0.678 (0.001) \\
    \cline{1-10}
    F1 score &  & 0.333 (0.299) & 0.550 (0.224) & 0.630 (0.086) & 0.684 (0.038) & 0.702 (0.013) & 0.712 (0.007) & 0.712 (0.006) & 0.714 (0.003) \\
    \cline{1-10}
    TP rate &  & 0.385 (0.396) & 0.640 (0.333) & 0.663 (0.170) & 0.693 (0.085) & 0.702 (0.031) & 0.720 (0.016) & 0.722 (0.013) & 0.727 (0.007) \\
    \cline{1-10}
    TN rate &  & 0.587 (0.403) & 0.409 (0.339) & 0.487 (0.196) & 0.593 (0.097) & 0.630 (0.031) & 0.623 (0.013) & 0.620 (0.009) & 0.617 (0.008) \\
    \cline{1-10}
    FN rate &  & 0.615 (0.396) & 0.360 (0.333) & 0.337 (0.170) & 0.307 (0.085) & 0.298 (0.031) & 0.280 (0.016) & 0.278 (0.013) & 0.273 (0.007) \\
    \cline{1-10}
    FP rate &  & 0.413 (0.403) & 0.591 (0.339) & 0.513 (0.196) & 0.407 (0.097) & 0.370 (0.031) & 0.377 (0.013) & 0.380 (0.009) & 0.383 (0.008) \\
    \cline{1-10}
    AUC &  & 0.500 (0.105) & 0.552 (0.082) & 0.604 (0.084) & 0.694 (0.019) & 0.714 (0.003) & 0.718 (0.002) & 0.718 (0.001) & 0.719 (0.001) \\
    \cline{1-10}
    SPD (race) &  & 0.048 (0.127) & -0.067 (0.175) & -0.192 (0.316) & -0.276 (0.114) & -0.218 (0.060) & -0.243 (0.018) & -0.236 (0.016) & -0.245 (0.012) \\
    \cline{1-10}
    SPD (sex) &  & 0.035 (0.155) & -0.029 (0.130) & -0.096 (0.200) & -0.173 (0.075) & -0.221 (0.057) & -0.211 (0.019) & -0.208 (0.015) & -0.205 (0.010) \\
    \cline{1-10}
    SPD (race + sex) &  & 0.060 (0.149) & -0.064 (0.123) & -0.178 (0.269) & -0.241 (0.076) & -0.250 (0.037) & -0.243 (0.019) & -0.243 (0.015) & -0.242 (0.010) \\
    \cline{1-10}
    AOD (race) &  & 0.045 (0.124) & -0.062 (0.174) & -0.172 (0.315) & -0.238 (0.115) & -0.176 (0.060) & -0.199 (0.018) & -0.193 (0.015) & -0.202 (0.013) \\
    \cline{1-10}
    AOD (sex) &  & 0.032 (0.152) & -0.023 (0.130) & -0.079 (0.196) & -0.141 (0.077) & -0.185 (0.057) & -0.174 (0.020) & -0.171 (0.016) & -0.169 (0.010) \\
    \cline{1-10}
    AOD (race + sex) &  & 0.056 (0.147) & -0.058 (0.122) & -0.153 (0.270) & -0.198 (0.078) & -0.208 (0.038) & -0.200 (0.021) & -0.200 (0.016) & -0.200 (0.011) \\
    \cline{1-10}
    FN rate balance (race) &  & -0.046 (0.132) & 0.057 (0.161) & 0.183 (0.311) & 0.246 (0.118) & 0.164 (0.063) & 0.190 (0.020) & 0.183 (0.018) & 0.192 (0.013) \\
    \cline{1-10}
    FN rate balance (sex) &  & -0.032 (0.167) & 0.025 (0.119) & 0.081 (0.191) & 0.143 (0.072) & 0.181 (0.059) & 0.170 (0.021) & 0.166 (0.017) & 0.161 (0.009) \\
    \cline{1-10}
    FN rate balance (race + sex) &  & -0.061 (0.154) & 0.061 (0.114) & 0.184 (0.248) & 0.241 (0.072) & 0.230 (0.039) & 0.223 (0.018) & 0.221 (0.015) & 0.216 (0.008) \\
    \cline{1-10}
    FP rate balance (race) &  & 0.043 (0.119) & -0.067 (0.195) & -0.161 (0.322) & -0.230 (0.115) & -0.188 (0.061) & -0.208 (0.017) & -0.203 (0.013) & -0.211 (0.013) \\
    \cline{1-10}
    FP rate balance (sex) &  & 0.032 (0.141) & -0.022 (0.147) & -0.077 (0.209) & -0.138 (0.085) & -0.188 (0.057) & -0.179 (0.022) & -0.176 (0.017) & -0.177 (0.012) \\
    \cline{1-10}
    FP rate balance (race + sex) &  & 0.052 (0.150) & -0.055 (0.147) & -0.123 (0.305) & -0.155 (0.093) & -0.186 (0.042) & -0.176 (0.028) & -0.180 (0.023) & -0.184 (0.016) \\
    \bottomrule
    \end{tabular}}
\end{table}
\end{landscape}

\newpage
\begin{landscape}
\begin{table}
\caption{Full set of simulation results for the COMPAS dataset for $\eta=0.08$ and for various values of the privacy parameter $\varepsilon$.
The other privacy parameter $\delta$ is assumed to be $\delta=10^{-9}$ throughout.
Values in the table are in the format ``mean (standard deviation)'' of the output values from many runs of the evaluations.
At least 30 successful runs were performed in each case, except for $\varepsilon=10^{-2}$ and $\varepsilon=10^{-1.5}$, which had 25 and 26 successful runs, respectively.
The final column represents a run without the DP synthesis step as a baseline.
}
\label{tab:COMPAS_results_eta08}
\resizebox{1.0\linewidth}{!}{
\begin{tabular}{lrrrrrrrrr}
    \toprule
     & $\varepsilon$ & $10^{-2.0}$ & $10^{-1.5}$ & $10^{-1.0}$ & $10^{-0.5}$ & $10^{0.0}$ & $10^{0.5}$ & $10^{1.0}$ & None \\
    \midrule
    KS Stat &  & 0.540 (0.162) & 0.291 (0.083) & 0.104 (0.033) & 0.047 (0.016) & 0.021 (0.005) & 0.014 (0.005) & 0.014 (0.004) & 0.008 (0.003) \\
    \cline{1-10}
    KS p-val &  & 0.000 (0.000) & 0.000 (0.000) & 0.000 (0.000) & 0.018 (0.039) & 0.408 (0.259) & 0.783 (0.248) & 0.785 (0.243) & 0.974 (0.068) \\
    \cline{1-10}
    Cumulative 1-way TVD &  & 1.310 (0.431) & 0.579 (0.165) & 0.189 (0.046) & 0.059 (0.020) & 0.021 (0.005) & 0.012 (0.004) & 0.009 (0.002) & 0.008 (0.003) \\
    \cline{1-10}
    Cumulative 2-way TVD &  & 5.036 (1.437) & 2.462 (0.692) & 1.071 (0.168) & 0.537 (0.060) & 0.217 (0.025) & 0.135 (0.013) & 0.108 (0.008) & 0.100 (0.010) \\
    \cline{1-10}
    Cumulative 3-way TVD &  & 8.506 (2.026) & 4.511 (1.214) & 2.298 (0.285) & 1.289 (0.149) & 0.572 (0.057) & 0.427 (0.029) & 0.387 (0.018) & 0.301 (0.023) \\
    \cline{1-10}
    COD (race) &  & 0.011 (0.036) & -0.011 (0.032) & -0.015 (0.039) & -0.050 (0.029) & -0.062 (0.011) & -0.066 (0.006) & -0.066 (0.005) & -0.065 (0.006) \\
    \cline{1-10}
    COD (sex) &  & -0.000 (0.024) & -0.013 (0.036) & -0.003 (0.044) & -0.024 (0.031) & -0.045 (0.015) & -0.047 (0.009) & -0.047 (0.009) & -0.049 (0.010) \\
    \cline{1-10}
    COD (race + sex) &  & 0.010 (0.031) & -0.020 (0.105) & -0.013 (0.041) & -0.039 (0.032) & -0.046 (0.017) & -0.041 (0.016) & -0.040 (0.018) & -0.044 (0.013) \\
    \cline{1-10}
    Accuracy &  & 0.518 (0.050) & 0.521 (0.064) & 0.576 (0.060) & 0.653 (0.020) & 0.670 (0.007) & 0.672 (0.005) & 0.672 (0.005) & 0.672 (0.005) \\
    \cline{1-10}
    F1 score &  & 0.483 (0.275) & 0.506 (0.248) & 0.624 (0.087) & 0.694 (0.021) & 0.704 (0.011) & 0.707 (0.008) & 0.707 (0.007) & 0.707 (0.007) \\
    \cline{1-10}
    TP rate &  & 0.577 (0.393) & 0.590 (0.366) & 0.660 (0.168) & 0.711 (0.054) & 0.708 (0.022) & 0.713 (0.016) & 0.713 (0.015) & 0.714 (0.014) \\
    \cline{1-10}
    TN rate &  & 0.444 (0.401) & 0.434 (0.366) & 0.472 (0.228) & 0.581 (0.080) & 0.623 (0.019) & 0.621 (0.009) & 0.621 (0.008) & 0.620 (0.007) \\
    \cline{1-10}
    FN rate &  & 0.423 (0.393) & 0.410 (0.366) & 0.340 (0.168) & 0.289 (0.054) & 0.292 (0.022) & 0.287 (0.016) & 0.287 (0.015) & 0.286 (0.014) \\
    \cline{1-10}
    FP rate &  & 0.556 (0.401) & 0.566 (0.366) & 0.528 (0.228) & 0.419 (0.080) & 0.377 (0.019) & 0.379 (0.009) & 0.379 (0.008) & 0.380 (0.007) \\
    \cline{1-10}
    AUC &  & 0.507 (0.092) & 0.522 (0.100) & 0.589 (0.089) & 0.690 (0.026) & 0.711 (0.002) & 0.713 (0.001) & 0.713 (0.001) & 0.713 (0.002) \\
    \cline{1-10}
    SPD (race) &  & 0.000 (0.153) & -0.042 (0.116) & -0.122 (0.306) & -0.221 (0.089) & -0.180 (0.049) & -0.191 (0.028) & -0.195 (0.029) & -0.191 (0.028) \\
    \cline{1-10}
    SPD (sex) &  & -0.071 (0.191) & -0.046 (0.163) & -0.096 (0.218) & -0.126 (0.096) & -0.169 (0.033) & -0.171 (0.026) & -0.178 (0.023) & -0.178 (0.021) \\
    \cline{1-10}
    SPD (race + sex) &  & -0.069 (0.186) & -0.053 (0.153) & -0.146 (0.265) & -0.163 (0.103) & -0.174 (0.051) & -0.174 (0.047) & -0.187 (0.032) & -0.185 (0.036) \\
    \cline{1-10}
    AOD (race) &  & 0.002 (0.152) & -0.037 (0.117) & -0.104 (0.304) & -0.183 (0.089) & -0.139 (0.048) & -0.149 (0.026) & -0.154 (0.027) & -0.150 (0.026) \\
    \cline{1-10}
    AOD (sex) &  & -0.067 (0.193) & -0.043 (0.168) & -0.078 (0.221) & -0.092 (0.098) & -0.133 (0.032) & -0.135 (0.026) & -0.142 (0.024) & -0.143 (0.021) \\
    \cline{1-10}
    AOD (race + sex) &  & -0.067 (0.186) & -0.047 (0.161) & -0.124 (0.274) & -0.119 (0.110) & -0.133 (0.050) & -0.134 (0.049) & -0.147 (0.032) & -0.145 (0.036) \\
    \cline{1-10}
    FN rate balance (race) &  & -0.010 (0.146) & 0.045 (0.112) & 0.110 (0.295) & 0.183 (0.090) & 0.121 (0.053) & 0.130 (0.032) & 0.135 (0.034) & 0.130 (0.034) \\
    \cline{1-10}
    FN rate balance (sex) &  & 0.077 (0.182) & 0.046 (0.147) & 0.090 (0.211) & 0.097 (0.094) & 0.127 (0.036) & 0.128 (0.026) & 0.135 (0.023) & 0.135 (0.023) \\
    \cline{1-10}
    FN rate balance (race + sex) &  & 0.066 (0.176) & 0.058 (0.133) & 0.154 (0.236) & 0.163 (0.093) & 0.149 (0.057) & 0.148 (0.045) & 0.160 (0.034) & 0.158 (0.038) \\
    \cline{1-10}
    FP rate balance (race) &  & -0.006 (0.162) & -0.030 (0.130) & -0.097 (0.317) & -0.184 (0.093) & -0.156 (0.045) & -0.169 (0.022) & -0.173 (0.021) & -0.170 (0.020) \\
    \cline{1-10}
    FP rate balance (sex) &  & -0.058 (0.206) & -0.041 (0.190) & -0.066 (0.238) & -0.086 (0.106) & -0.139 (0.032) & -0.142 (0.028) & -0.150 (0.026) & -0.150 (0.021) \\
    \cline{1-10}
    FP rate balance (race + sex) &  & -0.069 (0.203) & -0.037 (0.203) & -0.094 (0.321) & -0.074 (0.141) & -0.117 (0.053) & -0.119 (0.056) & -0.134 (0.036) & -0.132 (0.039) \\
    \bottomrule
    \end{tabular}}
\end{table}
\end{landscape}

\newpage
\begin{landscape}
\begin{table}
\caption{Full set of simulation results for the DP-DP-CTGAN + RW experiment on the Adult dataset  with continuous age for various values of the privacy parameter $\varepsilon$ without the fairness data preprocessing.
The other privacy parameter $\delta$ is assumed to be $\delta\approx10^{-7}$ throughout.
Values in the table are in the format ``mean (standard deviation)'' of the output values from many runs of the evaluations.
35 runs were performed in each case.
The final column represents a run that is also without the DP synthesis step as a baseline.
}
\label{tab:Adult_continuous_results_no_fairness}
\resizebox{1.0\linewidth}{!}{
    \begin{tabular}{lrrrrrrrr}
    \toprule
     & $\varepsilon$ & $10^{-0.5}$ & $10^{0.0}$ & $10^{0.5}$ & $10^{1.0}$ & $10^{1.5}$ & $10^{2.0}$ & None \\
    \midrule
    KS Stat &  & 0.799 (0.183) & 0.443 (0.179) & 0.312 (0.089) & 0.259 (0.078) & 0.209 (0.085) & 0.224 (0.087) & 0.000 (0) \\
    \cline{1-9}
    KS p-val &  & 0.000 (0.000) & 0.000 (0.000) & 0.000 (0.000) & 0.000 (0.000) & 0.000 (0.000) & 0.000 (0.000) & 1.000 (0) \\
    \cline{1-9}
    Cumulative 1-way TVD &  & 1.369 (0.357) & 0.774 (0.252) & 0.620 (0.144) & 0.514 (0.136) & 0.400 (0.096) & 0.445 (0.119) & 0.000 (0) \\
    \cline{1-9}
    Cumulative 2-way TVD &  & 3.172 (0.599) & 2.003 (0.508) & 1.602 (0.341) & 1.317 (0.297) & 1.049 (0.211) & 1.174 (0.269) & 0.000 (0) \\
    \cline{1-9}
    Cumulative 3-way TVD &  & 2.497 (0.346) & 1.790 (0.357) & 1.451 (0.287) & 1.209 (0.230) & 0.972 (0.192) & 1.093 (0.229) & 0.000 (0) \\
    \cline{1-9}
    COD (race) &  & 0.133 (0.310) & -0.079 (0.200) & -0.143 (0.109) & -0.143 (0.106) & -0.127 (0.105) & -0.115 (0.105) & -0.099 (0) \\
    \cline{1-9}
    COD (sex) &  & -0.106 (0.164) & -0.210 (0.201) & -0.134 (0.169) & -0.182 (0.125) & -0.183 (0.070) & -0.173 (0.090) & -0.193 (0) \\
    \cline{1-9}
    COD (race+sex) &  & -0.036 (0.185) & -0.193 (0.204) & -0.147 (0.155) & -0.182 (0.120) & -0.179 (0.077) & -0.164 (0.089) & -0.183 (0) \\
    \cline{1-9}
    Accuracy &  & 0.622 (0.201) & 0.728 (0.057) & 0.744 (0.050) & 0.757 (0.045) & 0.771 (0.016) & 0.773 (0.016) & 0.794 (0) \\
    \cline{1-9}
    F1 score &  & 0.121 (0.177) & 0.218 (0.203) & 0.311 (0.178) & 0.367 (0.152) & 0.383 (0.171) & 0.389 (0.108) & 0.463 (0) \\
    \cline{1-9}
    TP rate &  & 0.258 (0.419) & 0.208 (0.233) & 0.288 (0.225) & 0.324 (0.186) & 0.331 (0.175) & 0.314 (0.135) & 0.361 (0) \\
    \cline{1-9}
    TN rate &  & 0.740 (0.400) & 0.899 (0.123) & 0.893 (0.119) & 0.898 (0.102) & 0.915 (0.058) & 0.922 (0.050) & 0.935 (0) \\
    \cline{1-9}
    FN rate &  & 0.742 (0.419) & 0.792 (0.233) & 0.712 (0.225) & 0.676 (0.186) & 0.669 (0.175) & 0.686 (0.135) & 0.639 (0) \\
    \cline{1-9}
    FP rate &  & 0.260 (0.400) & 0.101 (0.123) & 0.107 (0.119) & 0.102 (0.102) & 0.085 (0.058) & 0.078 (0.050) & 0.065 (0) \\
    \cline{1-9}
    AUC &  & 0.485 (0.176) & 0.621 (0.171) & 0.719 (0.077) & 0.746 (0.059) & 0.758 (0.044) & 0.767 (0.032) & 0.800 (0) \\
    \cline{1-9}
    SPD (race) &  & 0.077 (0.244) & 0.006 (0.195) & -0.099 (0.125) & -0.115 (0.132) & -0.111 (0.085) & -0.096 (0.071) & -0.092 (0) \\
    \cline{1-9}
    SPD (sex) &  & -0.014 (0.056) & -0.130 (0.218) & -0.113 (0.157) & -0.160 (0.166) & -0.159 (0.086) & -0.164 (0.105) & -0.196 (0) \\
    \cline{1-9}
    SPD (race+sex) &  & 0.017 (0.096) & -0.092 (0.199) & -0.124 (0.159) & -0.171 (0.179) & -0.167 (0.089) & -0.165 (0.100) & -0.190 (0) \\
    \cline{1-9}
    AOD (race) &  & 0.083 (0.244) & 0.016 (0.233) & -0.103 (0.149) & -0.126 (0.165) & -0.120 (0.123) & -0.100 (0.104) & -0.081 (0) \\
    \cline{1-9}
    AOD (sex) &  & -0.014 (0.041) & -0.136 (0.231) & -0.107 (0.183) & -0.171 (0.189) & -0.174 (0.103) & -0.188 (0.129) & -0.248 (0) \\
    \cline{1-9}
    AOD (race+sex) &  & 0.022 (0.095) & -0.086 (0.207) & -0.116 (0.167) & -0.174 (0.191) & -0.171 (0.101) & -0.173 (0.108) & -0.206 (0) \\
    \cline{1-9}
    FN rate balance (race) &  & -0.090 (0.248) & -0.013 (0.305) & 0.136 (0.204) & 0.177 (0.229) & 0.172 (0.196) & 0.142 (0.169) & 0.111 (0) \\
    \cline{1-9}
    FN rate balance (sex) &  & 0.011 (0.033) & 0.168 (0.271) & 0.144 (0.245) & 0.243 (0.230) & 0.258 (0.156) & 0.280 (0.180) & 0.397 (0) \\
    \cline{1-9}
    FN rate balance (race+sex) &  & -0.030 (0.101) & 0.104 (0.247) & 0.150 (0.203) & 0.236 (0.219) & 0.243 (0.151) & 0.246 (0.142) & 0.310 (0) \\
    \cline{1-9}
    FP rate balance (race) &  & 0.075 (0.245) & 0.019 (0.170) & -0.069 (0.114) & -0.075 (0.117) & -0.068 (0.060) & -0.057 (0.050) & -0.050 (0) \\
    \cline{1-9}
    FP rate balance (sex) &  & -0.016 (0.054) & -0.105 (0.208) & -0.070 (0.137) & -0.100 (0.160) & -0.091 (0.057) & -0.095 (0.085) & -0.099 (0) \\
    \cline{1-9}
    FP rate balance (race+sex) &  & 0.015 (0.093) & -0.069 (0.187) & -0.081 (0.147) & -0.112 (0.177) & -0.100 (0.060) & -0.099 (0.084) & -0.101 (0) \\
    \bottomrule
    \end{tabular}}
\end{table}
\end{landscape}

\newpage
\begin{landscape}
\begin{table}
\caption{Full set of simulation results for the DP-CTGAN + RW experiment on the Adult dataset with continuous age for various values of the privacy parameter $\varepsilon$ with the reweighting fairness data preprocessing.
The other privacy parameter $\delta$ is assumed to be $\delta\approx10^{-7}$ throughout.
Values in the table are in the format ``mean (standard deviation)'' of the output values from many runs of the evaluations.
35 runs were performed in each case.
The final column represents a run that is also without the DP synthesis step as a baseline.
}
\label{tab:Adult_continuous_results_with_fairness}
\resizebox{1.0\linewidth}{!}{
\begin{tabular}{lrrrrrrrr}
    \toprule
     & $\varepsilon$ & $10^{-0.5}$ & $10^{0.0}$ & $10^{0.5}$ & $10^{1.0}$ & $10^{1.5}$ & $10^{2.0}$ & None \\
    \midrule
    KS Stat &  & 0.789 (0.193) & 0.418 (0.156) & 0.303 (0.109) & 0.253 (0.111) & 0.216 (0.072) & 0.214 (0.072) & 0.000 (0.000) \\
    \cline{1-9}
    KS p-val &  & 0.000 (0.000) & 0.000 (0.000) & 0.000 (0.000) & 0.000 (0.000) & 0.000 (0.000) & 0.000 (0.000) & 1.000 (0.000) \\
    \cline{1-9}
    Cumulative 1-way TVD &  & 1.417 (0.410) & 0.832 (0.223) & 0.642 (0.199) & 0.502 (0.162) & 0.427 (0.110) & 0.408 (0.105) & 0.000 (0.000) \\
    \cline{1-9}
    Cumulative 2-way TVD &  & 3.188 (0.639) & 2.094 (0.435) & 1.652 (0.453) & 1.327 (0.362) & 1.101 (0.214) & 1.088 (0.241) & 0.000 (0.000) \\
    \cline{1-9}
    Cumulative 3-way TVD &  & 2.489 (0.369) & 1.832 (0.329) & 1.502 (0.380) & 1.233 (0.306) & 1.016 (0.166) & 1.020 (0.205) & 0.000 (0.000) \\
    \cline{1-9}
    COD (race) &  & 0.097 (0.201) & 0.025 (0.269) & -0.113 (0.158) & -0.134 (0.158) & -0.084 (0.131) & -0.116 (0.094) & -0.099 (0.000) \\
    \cline{1-9}
    COD (sex) &  & -0.064 (0.160) & -0.162 (0.227) & -0.131 (0.135) & -0.218 (0.119) & -0.167 (0.112) & -0.191 (0.081) & -0.193 (0.000) \\
    \cline{1-9}
    COD (race+sex) &  & -0.021 (0.143) & -0.132 (0.246) & -0.144 (0.122) & -0.215 (0.113) & -0.150 (0.112) & -0.183 (0.069) & -0.183 (0.000) \\
    \cline{1-9}
    Accuracy &  & 0.642 (0.194) & 0.685 (0.094) & 0.733 (0.052) & 0.747 (0.045) & 0.756 (0.023) & 0.765 (0.021) & 0.788 (0.000) \\
    \cline{1-9}
    F1 score &  & 0.115 (0.170) & 0.265 (0.209) & 0.272 (0.191) & 0.380 (0.143) & 0.383 (0.153) & 0.433 (0.107) & 0.473 (0.000) \\
    \cline{1-9}
    TP rate &  & 0.230 (0.397) & 0.318 (0.286) & 0.266 (0.244) & 0.349 (0.175) & 0.348 (0.175) & 0.389 (0.144) & 0.386 (0.000) \\
    \cline{1-9}
    TN rate &  & 0.776 (0.386) & 0.805 (0.194) & 0.885 (0.132) & 0.877 (0.093) & 0.889 (0.069) & 0.888 (0.063) & 0.919 (0.000) \\
    \cline{1-9}
    FN rate &  & 0.770 (0.397) & 0.682 (0.286) & 0.734 (0.244) & 0.651 (0.175) & 0.652 (0.175) & 0.611 (0.144) & 0.614 (0.000) \\
    \cline{1-9}
    FP rate &  & 0.224 (0.386) & 0.195 (0.194) & 0.115 (0.132) & 0.123 (0.093) & 0.111 (0.069) & 0.112 (0.063) & 0.081 (0.000) \\
    \cline{1-9}
    AUC &  & 0.503 (0.180) & 0.615 (0.141) & 0.689 (0.089) & 0.736 (0.065) & 0.739 (0.050) & 0.751 (0.034) & 0.790 (0.000) \\
    \cline{1-9}
    SPD (race) &  & 0.016 (0.034) & 0.091 (0.295) & -0.029 (0.140) & -0.064 (0.161) & -0.023 (0.128) & -0.035 (0.113) & -0.035 (0.000) \\
    \cline{1-9}
    SPD (sex) &  & -0.007 (0.129) & -0.103 (0.226) & -0.007 (0.126) & -0.133 (0.121) & -0.109 (0.107) & -0.143 (0.095) & -0.167 (0.000) \\
    \cline{1-9}
    SPD (race+sex) &  & -0.002 (0.100) & -0.047 (0.197) & -0.009 (0.101) & -0.119 (0.096) & -0.087 (0.084) & -0.120 (0.078) & -0.140 (0.000) \\
    \cline{1-9}
    AOD (race) &  & 0.016 (0.031) & 0.115 (0.315) & -0.012 (0.178) & -0.045 (0.198) & 0.012 (0.155) & 0.013 (0.161) & 0.023 (0.000) \\
    \cline{1-9}
    AOD (sex) &  & -0.002 (0.144) & -0.091 (0.231) & 0.020 (0.147) & -0.125 (0.142) & -0.097 (0.138) & -0.138 (0.125) & -0.195 (0.000) \\
    \cline{1-9}
    AOD (race+sex) &  & 0.003 (0.104) & -0.022 (0.197) & 0.016 (0.107) & -0.100 (0.105) & -0.059 (0.098) & -0.088 (0.088) & -0.122 (0.000) \\
    \cline{1-9}
    FN rate balance (race) &  & -0.014 (0.037) & -0.128 (0.362) & 0.012 (0.240) & 0.053 (0.258) & -0.025 (0.203) & -0.040 (0.241) & -0.060 (0.000) \\
    \cline{1-9}
    FN rate balance (sex) &  & -0.004 (0.163) & 0.104 (0.241) & -0.013 (0.184) & 0.173 (0.193) & 0.143 (0.198) & 0.203 (0.177) & 0.314 (0.000) \\
    \cline{1-9}
    FN rate balance (race+sex) &  & -0.008 (0.110) & 0.019 (0.212) & -0.009 (0.127) & 0.132 (0.140) & 0.082 (0.138) & 0.117 (0.123) & 0.181 (0.000) \\
    \cline{1-9}
    FP rate balance (race) &  & 0.018 (0.036) & 0.101 (0.289) & -0.012 (0.127) & -0.037 (0.148) & -0.002 (0.118) & -0.014 (0.086) & -0.014 (0.000) \\
    \cline{1-9}
    FP rate balance (sex) &  & -0.008 (0.129) & -0.078 (0.239) & 0.026 (0.119) & -0.078 (0.103) & -0.051 (0.083) & -0.073 (0.078) & -0.076 (0.000) \\
    \cline{1-9}
    FP rate balance (race+sex) &  & -0.003 (0.102) & -0.026 (0.209) & 0.023 (0.098) & -0.068 (0.082) & -0.036 (0.062) & -0.060 (0.061) & -0.064 (0.000) \\
    \bottomrule
    \end{tabular}}
\end{table}
\end{landscape}

\begin{table*}[!t]
\footnotesize
\caption{Attributes In the German Credit dataset.}
\label{tab:german_credit_features}
\centering
\begin{tabularx}{\linewidth}{lX}
\toprule
Feature & Levels \\
\midrule
Status of existing checking account (qual.) &
A11: $<0$ DM; A12: $0\le\cdots<200$ DM; A13: $\ge200$ DM / salary assignments $\ge$1 year; A14: no checking account. \\
Duration in month (num.) & Numeric (months). \\
Credit history (qual.) &
A30: no credits taken / all credits paid back duly; A31: all credits at this bank paid back duly; A32: existing credits paid back duly till now; A33: delay in paying off in the past; A34: critical account / other credits existing (not at this bank). \\
Purpose (qual.) &
A40: car (new); A41: car (used); A42: furniture/equipment; A43: radio/television; A44: domestic appliances; A45: repairs; A46: education; A47: vacation; A48: retraining; A49: business; A410: others. \\
Credit amount (num.) & Numeric (currency units). \\
Savings account/bonds (qual.) &
A61: $<100$ DM; A62: $100\le\cdots<500$ DM; A63: $500\le\cdots<1000$ DM; A64: $\ge1000$ DM; A65: unknown / no savings account. \\
Present employment since (qual.) &
A71: unemployed; A72: $<1$ year; A73: 1--4 years; A74: 4--7 years; A75: $\ge7$ years. \\
Installment rate of disposable income (num.) & Numeric (percentage). \\
Personal status and sex (qual.) &
A91: male divorced/separated; A92: female divorced/separated/married; A93: male single; A94: male married/widowed; A95: female single. \\
Other debtors/guarantors (qual.) &
A101: none; A102: co-applicant; A103: guarantor. \\
Present residence since (num.) & Numeric (years). \\
Property (qual.) &
A121: real estate; A122: if not A121: building-society savings agreement / life insurance; A123: if not A121/A122: car or other (not in A6); A124: unknown / no property. \\
Age(num.) & Numeric (years). \\
Other installment plans (qual.) &
A141: bank; A142: stores; A143: none. \\
Housing (qual.) &
A151: rent; A152: own; A153: for free. \\
Number of existing credits at this bank (num.) & Numeric (count). \\
Job (qual.) &
A171: unemployed/unskilled—non-resident; A172: unskilled—resident; A173: skilled employee/official; A174: management/self-employed/highly qualified/ officer. \\
Number of people liable for maintenance (num.) & Numeric (count). \\
Telephone (qual.) &
A191: none; A192: yes, registered under customer's name. \\
Foreign worker (qual.) &
A201: yes; A202: no. \\
Target (label) & 1 = good, 0 = bad. \\
\bottomrule
\end{tabularx}
\end{table*}

\end{document}